\documentclass[11pt]{article}

\usepackage[preprint]{acl}

\usepackage{times}
\usepackage{latexsym}

\usepackage[T1]{fontenc}

\usepackage[utf8]{inputenc}

\usepackage{microtype}

\usepackage{inconsolata}

\usepackage{graphicx}

\usepackage{empheq}

\usepackage{xspace}
\usepackage{pifont}
\newcommand{\pheno}{{\fontfamily{lmtt}\selectfont \textbf{The First is The Best}}\xspace}

\newcommand{\ourmethod}{{\fontfamily{lmtt}\selectfont \textbf{RED}}\xspace}

\newcommand{\ffourmethod}{{\fontfamily{lmtt}\selectfont \textbf{\underline{Re}fine First and \underline{D}iscard Subs}}\xspace}

\newcommand{\first}{{\fontfamily{lmtt}\selectfont \textbf{First}}\xspace}

\newcommand{\subs}{{\fontfamily{lmtt}\selectfont \textbf{Subs}}\xspace}

\newcommand{\foe}{{\fontfamily{lmtt}\selectfont \textbf{FoE}}\xspace}

\usepackage[table,xcdraw,dvipsnames]{xcolor}
\usepackage{twemojis}
\usepackage{booktabs}
\usepackage{makecell}
\usepackage{bbding}
\usepackage{multirow}
\usepackage{array}
\newcommand{\darkred}[1]{\textcolor{RedOrange}{\ensuremath{\uparrow}\,#1}}
\newcommand{\darkblue}[1]{\textcolor{BlueGreen}{\ensuremath{\downarrow}\,#1}}
\definecolor{MorandiHeader}{RGB}{219, 226, 236} 
\definecolor{AltRowColor}{RGB}{245,244,242}
\definecolor{DeltaRowColor}{RGB}{236, 244, 221}
\newcommand{\gcell}[1]{\cellcolor{AltRowColor}#1}
\newcommand{\dcell}[1]{\cellcolor{DeltaRowColor}#1}
\definecolor{BaselineTag}{RGB}{238,220,210} 
\definecolor{RegenTag}{RGB}{214,230,242}    

\definecolor{first}{HTML}{C9352B}
\definecolor{sub}{HTML}{299D8F}

\newcommand{\cmark}{\checkmark}  
\newcommand{\xmark}{\ding{55}}    
\usepackage{wrapfig}
\usepackage{enumitem}
\usepackage{amsmath}
\usepackage{amssymb}
\usepackage{mathtools}
\usepackage{amsthm}

\usepackage{times}
\usepackage{latexsym}

\usepackage[T1]{fontenc}

\usepackage[utf8]{inputenc}

\usepackage{microtype}

\usepackage{inconsolata}

\usepackage{graphicx}

\usepackage{xspace}
\usepackage{pifont}
\definecolor{ccr}{RGB}{72, 192, 170}
\usepackage{twemojis}
\usepackage[table,xcdraw,dvipsnames]{xcolor}
\usepackage{booktabs}
\usepackage{makecell}
\usepackage{bbding}
\usepackage{multirow}
\usepackage{array}

\usepackage{fontawesome6}

\usepackage[most]{tcolorbox}
\definecolor{fancyTeal}{HTML}{008080}
\definecolor{softBack}{HTML}{F0F8FF}
\newtcolorbox{cbx}[1][]{
  enhanced,
  breakable,
  title={#1},
  colframe=fancyTeal,
  colback=softBack,
  colbacktitle=fancyTeal,
  fonttitle=\bfseries\sffamily,
  boxrule=0.5mm,
  arc=3mm,
  attach boxed title to top left={
    yshift=-2mm,
    xshift=4mm
  },
  boxed title style={
    boxrule=0mm,
    arc=1.5mm,
    shadow={2mm}{-2mm}{0mm}{black!15}
  },
  drop fuzzy shadow
}

\usepackage{wrapfig}
\usepackage{enumitem}
\usepackage{amsmath}
\usepackage{amssymb}
\usepackage{mathtools}
\usepackage{amsthm}

\usepackage{caption}
\usepackage{subcaption}

\usepackage{algorithm}
\usepackage{algpseudocode}

\usepackage{graphicx}
\usepackage{subcaption}

\lstdefinestyle{PromptStyle}{
    basicstyle=\fontfamily{pcr}\selectfont\small,
    breaklines=true,        
    columns=fullflexible,   
    keepspaces=true,        
    showstringspaces=false, 
    extendedchars=true,     
    frame=none,             
    aboveskip=0pt,          
    belowskip=0pt,
    escapechar=             
}
\newtcolorbox{PromptFrame}[1]{
    enhanced,
    breakable,              
    title={#1},             
    colframe=fancyTeal,
    colback=softBack,
    colbacktitle=fancyTeal,
    fonttitle=\bfseries\sffamily,
    boxrule=0.5mm,
    arc=3mm,
    attach boxed title to top left={
        yshift=-2mm,
        xshift=4mm
    },
    boxed title style={
        boxrule=0mm,
        arc=1.5mm,
        shadow={2mm}{-2mm}{0mm}{black!15}
    },
    drop fuzzy shadow
}

\title{FoE: Forest of Errors Makes the First Solution the Best \\ in Large Reasoning Models}

\author{
{\bfseries Kehan Jiang$^{1*}$ \quad Haonan Dong$^{2*}$ \quad Zhaolu Kang$^{1}$} \\
{\bfseries Zhengzhou Zhu$^{1}$ \quad Guojie Song$^{2\dagger}$} \\
$^{1}$School of Software and Microelectronics, Peking University \\
$^{2}$State Key Laboratory of General Artificial Intelligence, \\
School of Intelligence Science and Technology, Peking University \\
$^{*}$Equal contribution \qquad $^{\dagger}$Corresponding author \\
{\faEnvelope\ jiangkh5521@gmail.com, gjsong@pku.edu.cn}
}

\begin{document}
\maketitle
\begin{abstract}
Recent Large Reasoning Models (LRMs) like DeepSeek-R1 have demonstrated remarkable success in complex reasoning tasks, exhibiting human-like patterns in exploring multiple alternative solutions. Upon closer inspection, however, we uncover a surprising phenomenon: \pheno, where alternative solutions are not merely suboptimal but potentially detrimental. This observation \textit{challenges widely accepted test-time scaling laws}, leading us to hypothesize that \textbf{\textit{errors within the reasoning path scale concurrently with test time}}. Through comprehensive empirical analysis, we characterize errors as a forest-structured Forest of Errors (\foe) and conclude that \textbf{\textit{FoE makes the First the Best}}, which is underpinned by rigorous theoretical analysis. Leveraging these insights, we propose \ourmethod, a self-guided efficient reasoning framework comprising two components: I) \textit{Refining First}, which suppresses \foe growth in the first solution; and II) \textit{Discarding Subs}, which prunes subsequent \foe via dual-consistency. Extensive experiments across five benchmarks and six backbone models demonstrate that \ourmethod outperforms eight competitive baselines, achieving performance gains of up to $19.0\%$ while reducing token consumption by $37.7\%\sim70.4\%$. Moreover, comparative experiments on \foe metrics shed light on how \ourmethod achieves effectiveness.
\end{abstract}

\section{Introduction}
Reasoning capability stands as the cornerstone of human intelligence~\citep{human-reasoning}. Recently, LLMs have achieved significant advancements in reasoning, demonstrating immense potential across mathematical~\citep{cot-math}, scientific~\citep{cot-sci}, and coding tasks~\citep{cot-code}. This progress is largely attributed to the evolution of the Chain-of-Thought (CoT)~\citep{cot} research line, which specifically decomposes complex tasks into step-by-step reasoning processes. Furthermore, the advent of DeepSeek-R1~\citep{deepseekai2025deepseekr1incentivizingreasoningcapability} marks a paradigm shift for Large Reasoning Models (LRMs). Powered by Reinforcement Learning (RL)-based training frameworks~\citep{rl-1,rl-2,yu2025dapo}, R1-like models tend to generate more extensive responses while exhibiting an ``aha-moment'', characterized by self-verification, reflection, and the exploration of alternative methodologies during the reasoning process~\citep{deepseekai2025deepseekr1incentivizingreasoningcapability}.

\begin{figure}[!t]
\centering
\includegraphics[width=1.0\linewidth]{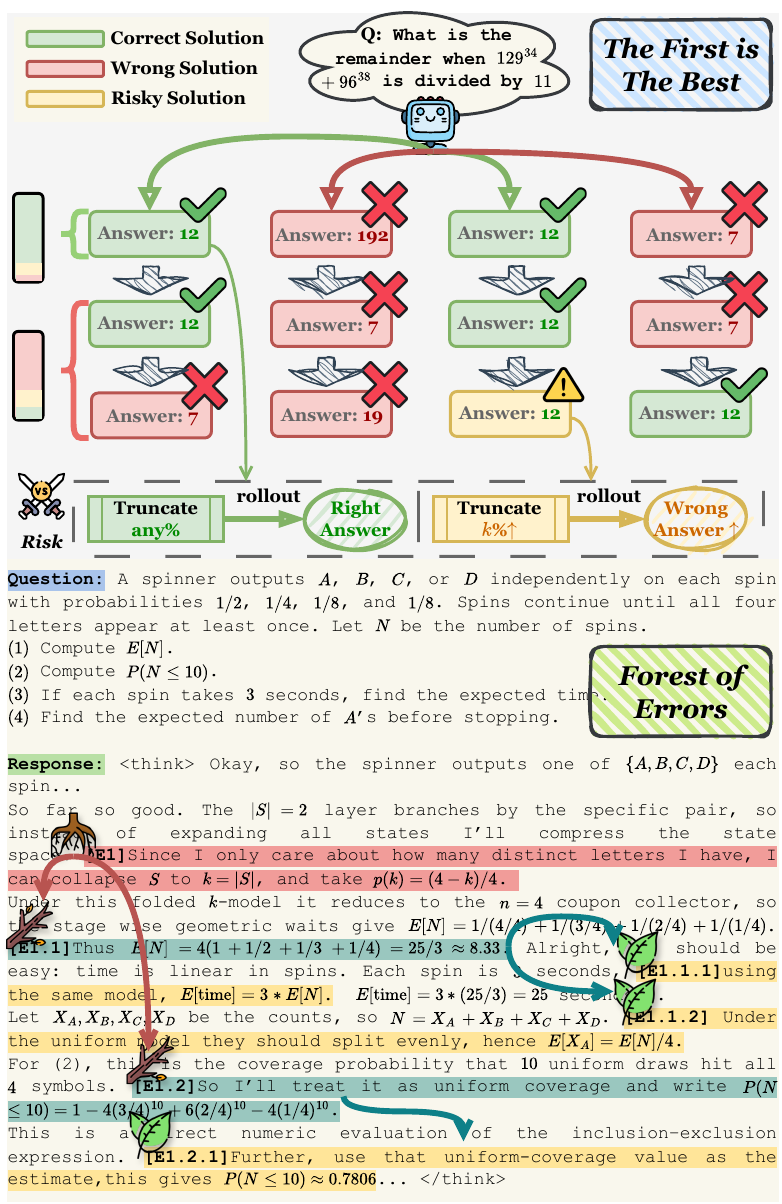} 
\vspace{-2em}
\caption{
    (\textbf{\textit{Upper}}) \pheno. (\textbf{\textit{Lower}}) Forest of Errors.}  
\label{fig:intro}
\vspace{-1.8em}
\end{figure}

\vspace{-0.5em}
\paragraph{Phenomenon \ding{182}.} Despite the success of R1-like models in reasoning tasks achieved through RL-driven self-evolution~\citep{o1,gemini}, we observe that their frequent exploration of multiple potential solutions results in excessively long CoT, leading to a substantial waste of computational resources~\citep{overthinking}. This observation prompts a critical question: \textit{Is the exploration of multiple solutions truly necessary?} Upon closer inspection of the solutions generated during the reasoning process, we uncover a surprising phenomenon: \pheno. We denote the first solution generated by LRM as \first and the subsequent solutions as \subs. Specifically, as illustrated in Figure \ref{fig:intro}(\textbf{\textit{Upper}}) and Table \ref{table:phenomenon}, we find that: \ding{168} \first is optimal in up to $93.7\%$ of cases; \ding{169} \subs fails to rectify an erroneous \first with a $75.4\%\sim82.8\%$ probability; \ding{170} more concerningly, there is a huge probability (up to $21.2\%$) that \subs misleads a potentially correct \first towards an incorrect answer; and \ding{171} even when both \first and \subs are correct, \subs harbors a substantial latent risk of error. To some extent, it \textit{challenges the widely accepted test-time scaling law}~\citep{tts} and motivates us to hypothesize that \textbf{\textit{as test-time scales up, errors within the reasoning path may scale up concurrently}}.

\vspace{-0.5em}
\paragraph{Phenomenon \ding{183}.} To rigorously validate this hypothesis, we conduct an in-depth analysis of errors emerging during the reasoning process. As depicted in Figure \ref{fig:intro} (\textbf{\textit{Lower}}), we observe that errors propagate diffusely, stemming from multiple root causes and ultimately manifesting as a forest-like structure, which we term the Forest of Errors (\foe). Through qualitative and quantitative analyses of \foe, we find that: \ding{168} There exists a strong dependencies between parent and child nodes within \foe, with root error nodes playing a pivotal role in the overall growth of the error structure; \ding{169} The scale of \foe in \first is significantly smaller than that in \subs; \ding{170} The generation of error nodes is closely correlated with entropy and entropy variance, with \subs exhibiting a higher propensity for generating error nodes compared to \first; and \ding{171} The self-reflection mechanisms of LRMs appear ineffective in pruning \foe. Extensive empirical experiments yield a total of 5 observations (\textbf{Obs.}), consistently showing that errors accumulate with reasoning length, validating that \textbf{\textit{FoE makes the First the Best}}.

\vspace{-0.5em}
\paragraph{Practical Method.} Building upon the systematic empirical analysis above, we further provide a rigorous mathematical analysis and proof of \textbf{\textit{FoE makes the First the Best}} through the lens of probabilistic branching process theory. Furthermore, synthesizing insights from Phenomena \ding{182} and \ding{183}, we propose \ourmethod (\ffourmethod), an efficient reasoning method based on self-guidance. Specifically, \ourmethod comprises two components: \textbf{I) \textit{Refining \first}}, inspired by the pivotal role of root nodes in \foe and the strong correlation between error generation and entropy statistics (entropy and its variance), we employ an entropy-based intervention mechanism at positions prone to root errors within \first, thus ensuring a superior one. \textbf{II) \textit{Discarding \subs}}, given our finding that \subs not only fails to improve performance but also carries the risk of misleading the model, we implement a dual-consistency-based early stopping strategy. This prevents inferior subsequent solutions from negatively impacting the final answer. Our contributions are summarized as follows:

\begin{itemize}[leftmargin=*]
\vspace{-0.7em}
\item[\texttwemoji{heart suit}] \textbf{\textit{Phenomenon Discovery.}} We identify two critical phenomena: \pheno and \foe. These findings reveal a pivotal insight: as test-time scales up, errors within the reasoning path scale up concurrently. This leads to the conclusion that FoE renders the first the best.
\vspace{-0.6em}
\item[\texttwemoji{heart suit}] \textbf{\textit{Insightful Analysis.}} Through qualitative and quantitative experiments alongside theoretical derivation, we derive five insightful observations for future reasoning models. Furthermore, we established a \foe-based probabilistic framework to theoretically validate the optimality of the \first (detailed in Appendix \ref{sec:theory}).
\vspace{-0.6em}
\item[\texttwemoji{heart suit}] \textbf{\textit{Practical Method.}} Building upon these empirical foundations, we propose \ourmethod, a self-guided efficient reasoning method designed to optimize \first while pruning \subs. Extensive experiments across four datasets and six backbone models demonstrate that \ourmethod outperforms seven competitive baselines, delivering up to $19.0\%$ accuracy gains and $37.7\%\sim70.4\%$ token reduction. Furthermore, evaluations on \foe metrics confirm that \ourmethod effectively eliminates \foe.
\end{itemize}

\begin{figure*}[!t]
\centering
\includegraphics[width=1.0\linewidth]{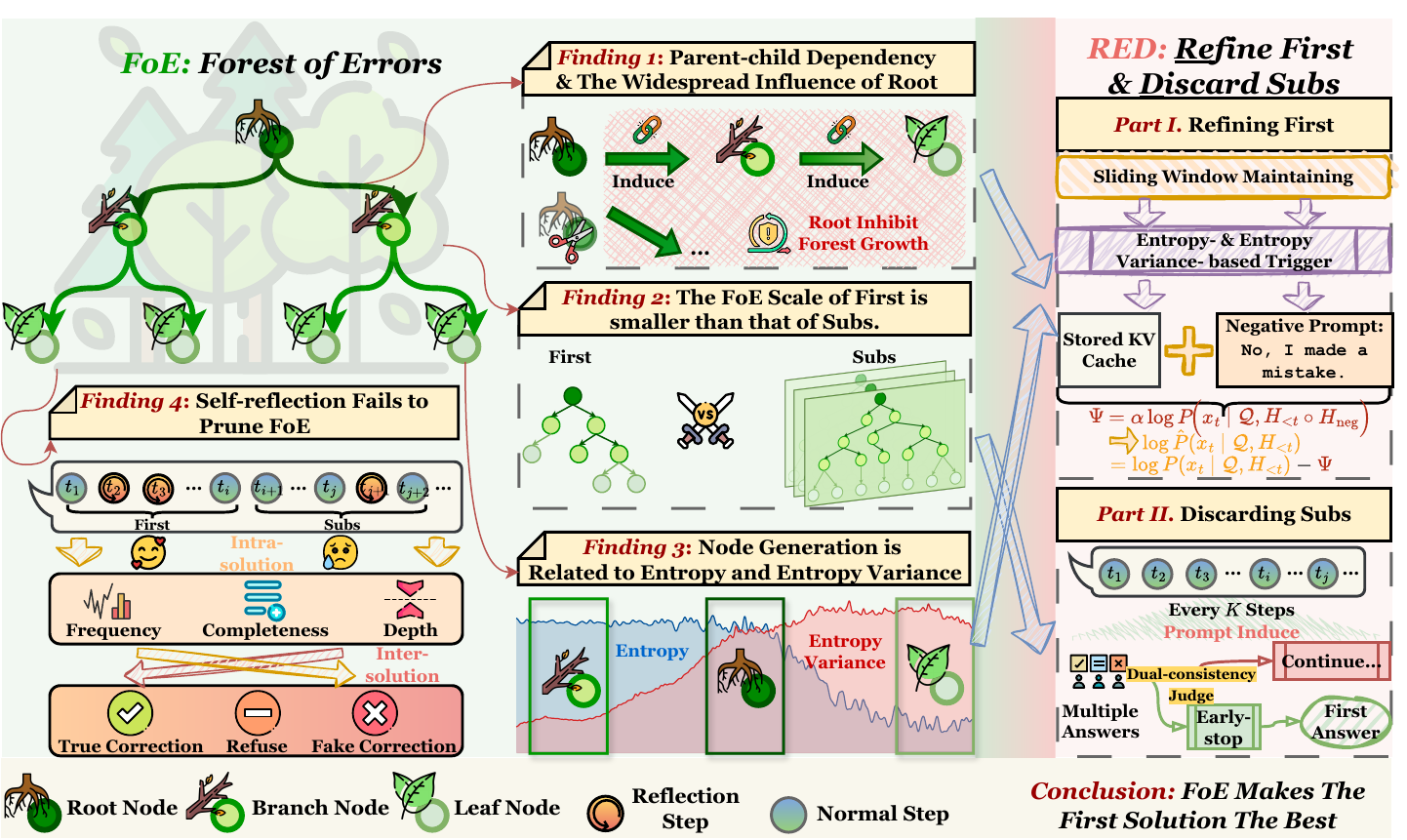} 
\vspace{-2.1em}
\caption{
    (\textbf{\textit{Left}}) Key observations within the \foe. (\textbf{\textit{Right}}) Our proposed \ourmethod framework.}  
\label{fig:over-view}
\vspace{-1em}
\end{figure*}

\vspace{-1em}
\section{\pheno}
\vspace{-0.5em}
In this section, we investigate and identify the \pheno phenomenon across multiple datasets and models, as shown in Table~\ref{table:phenomenon}. We perform a statistical analysis of the correctness relationship between \first and \subs.
\vspace{-0.5em}
\paragraph{Obs.\ding{182}} When the \first is incorrect, an average of 75.4\%-82.8\% of \subs persist in being incorrect. This indicates that \subs struggle to rectify an incorrect \first, failing to provide effective cross-verification while resulting in a waste of computational resources.

\begin{table}[!t]
\centering
\vspace{-0.7em}
\renewcommand\tabcolsep{5.3pt}
\renewcommand\arraystretch{1.1}

\begingroup
\setlength{\abovetopsep}{0pt}
\setlength{\belowrulesep}{0pt}
\setlength{\aboverulesep}{0pt}
\setlength{\belowbottomsep}{0pt}

\resizebox{\linewidth}{!}{%
\begin{tabular}{c|c|c|c|c}
\toprule[1.5pt]
\rowcolor{MorandiHeader}
\textbf{Dataset} & \textbf{Model} &
\textbf{\cmark{}\,$\rightarrow$\,\xmark{}} & 
\textbf{\xmark{}\,$\rightarrow$\,\xmark{}} &
\textbf{\xmark{}\,$\rightarrow$\,\cmark{}} \\ 
\midrule

\multirow{5}{*}{AIME25}
& Qwen-distilled-7b& \textbf{18.3} & 77.4 & \underline{4.3} \\ 
& \gcell{Qwen-distilled-32b} & \gcell{\textbf{16.7}} & \gcell{76.9} & \gcell{\underline{6.4}} \\ 
& Qwen3-8b& \textbf{15.5} & 79.1 & \underline{5.4} \\
& \gcell{Qwen3-32b}& \gcell{\textbf{13.0}} & \gcell{79.9} & \gcell{\underline{7.1}} \\
& Llama-distilled-70b & \textbf{18.8} & 75.4 & \underline{5.8} \\
\hline

\multirow{5}{*}{\shortstack{GPQA-\\Diamond}}
& Qwen-distilled-7b& \textbf{16.0} & 81.1 & \underline{2.9} \\
& \gcell{Qwen-distilled-32b} & \gcell{\textbf{15.3}} & \gcell{81.4} & \gcell{\underline{3.3}} \\
& Qwen3-8b& \textbf{14.4} & 82.7 & \underline{2.9} \\
& \gcell{Qwen3-32b} & \gcell{\textbf{14.1}} & \gcell{82.8} & \gcell{\underline{3.1}} \\
& Llama-distilled-70b & \textbf{21.2} & 77.9 & \underline{0.9} \\
\bottomrule[1.5pt]
\end{tabular}%
}
\endgroup
\vspace{-0.9em}
\caption{Distribution of the Influence of \subs on \first, excluding\textbf{(\cmark{}\,$\rightarrow$\,\cmark{}}) (\%). Notably, 93.7\% of cases favor \first, defined as instances where it is either uniquely correct or more robust than \subs ($\blacktriangleright$ Appendix \ref{sec:rollback}).}
\label{table:phenomenon}
\vspace{-1.8em}
\end{table}

\vspace{-0.7em}
\paragraph{Obs.\ding{183}} The \subs fail to rectify an incorrect first one, with a maximum success rate of merely 2.0\%-7.1\%; in many dataset-model combinations, successful cases are entirely absent.
\vspace{-0.7em}
\paragraph{Obs.\ding{184}} More alarmingly, \subs may mislead the model from a correct first one to an incorrect answer. This probability reaches up to 18.8\%, significantly exceeding the rate of correcting an incorrect first one, implying that the potential risks of generating multiple solutions far outweigh their benefits.
\vspace{-1.8em}
\paragraph{Obs.\ding{185}} Furthermore, even in cases where both \first and \subs appear correct, \subs retains a latent risk of inducing errors. To investigate this stability, we save the KV cache at regular intervals during generation. Upon completion, we rollback all solutions by a range of steps and perform extensive random sampling to regenerate the final answers. We observe that the probability of \first yielding errors after sampling from the interruption point is merely 3.7\% of that of \subs. Detailed experimental results are provided in Appendix \ref{sec:rollback}.

\vspace{-0.7em}
\section{\foe}
\vspace{-0.7em}
Through the case analysis (Figure \ref{fig:intro}), we observe that errors within the reasoning process manifest as a forest-like structure, which we term \foe. To investigate this further, we conduct an in-depth analysis of \foe to validate our hypothesis that as test-time scales up, errors within the reasoning path may scale up concurrently, thereby explaining the optimality of \pheno. To this end, we proceed as follows: \textbf{(i)} we formalize the modeling of \foe and analyze the dependencies between parent and child nodes ($\blacktriangleright$ Section \ref{sec:foe-1}); \textbf{(ii)} we introduce evaluation metrics specific to \foe and assess both \first and \subs ($\blacktriangleright$ Section \ref{sec:foe-2}); \textbf{(iii)} we investigate the genesis of error nodes from an entropy perspective ($\blacktriangleright$ Section \ref{sec:foe-3}); and \textbf{(iv)} we examine whether the self-reflection capabilities of LRMs can prune \foe ($\blacktriangleright$ Section \ref{sec:foe-4}). Ultimately, these empirical analyses consistently support the conclusion that the FoE renders the First the Best.

\vspace{-0.7em}
\subsection{\foe Initialization.}
\label{sec:foe-1}
\vspace{-0.3em}
\paragraph{Forest Modeling.} To facilitate a quantitative analysis of the \foe, we propose a formal modeling approach. Specifically, we assume that all errors within the reasoning trace have been identified using the o1-based annotation method \cite{yang-etal-2025-beyond-first}, which leverages powerful closed-source models for robust error detection.  We first organize all errors into a chronological sequence, denoted as $[e_1, e_2, \dots, e_n]$. For a given error $e_j$, to identify its parent node $e_i$, we define a parent-child association score, $\text{Score}(e_i, e_j)$, which quantifies the likelihood of $e_i$ inducing $e_j$. To implement this, we establish a $1\sim5$ scoring scale based on expert-designed criteria. Leveraging advanced LLMs via few-shot prompting, we evaluate candidate nodes $e_k$ sequentially, ordered by their proximity to $e_j$ (from nearest to farthest). Using a pre-defined threshold $\tau$, if $\text{Score}(e_k, e_j) \geq \tau$, $e_k$ is identified as the parent of $e_j$. If no candidate exceeds the threshold, $e_j$ is designated as a root, instantiating a new Tree of Errors (ToE) within the FoE. The process then advances to the next unprocessed node and iterates. Further Modeling pipeline details are provided in Appendix \ref{sec:modeling}.

\begin{figure}[t]
\centering
\includegraphics[width=1\linewidth]{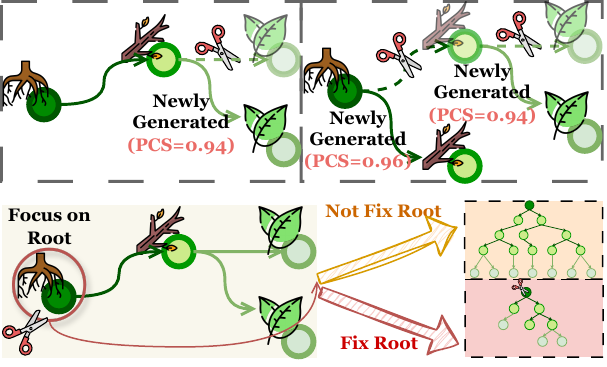} 
\vspace{-2.3em}
\caption{
 Manual correction on distinct error node types. (\textbf{\textit{Upper}}) Impact of rectifying individual \textbf{\underline{G}}randchild, \textbf{\underline{C}}hild, and \textbf{\underline{R}}oot nodes. (\textbf{\textit{Lower}}) Consequences of delayed root correction in a formed tree, demonstrating substantial mitigation of subsequent node proliferation.}  
\label{fig:3.1}
\vspace{-2em}
\end{figure}

\vspace{-0.6em}
\paragraph{Obs.\ding{182} Merely rectifying child nodes results in the parent continuously spawning new offspring, whereas correcting the root node significantly decelerates the error node generation rate.} 

To investigate dependencies, we employ an iterative leaf-to-root correction strategy on grandchild (G), child (C), and root (R) nodes. As shown in Figure \ref{fig:3.1} (\textbf{\textit{Upper}}), correcting only descendants (e.g., G2 or C1) fails to stop error propagation due to unaddressed ancestors. Conversely, correcting the root node even after it has spawned children effectively mitigates the rate of subsequent error generation (\textbf{\textit{Lower}}). Further details are in Appendix \ref{sec:findings-fix}.

\vspace{-0.6em}
\subsection{Evaluation Metrics of \foe}
\label{sec:foe-2}
\vspace{-0.4em}
To quantitatively assess the scale and growth of \foe, we design the static and dynamic metrics. 
\vspace{-0.6em}
\paragraph{Static Metrics.} We employ following static metrics commonly used for forest structures: \textbf{(i) forest size(FS)}, the number of trees within \foe; \textbf{(ii) average nodes per tree(N/T)}, the average error count within a single tree; \textbf{(iii) average depth per tree(D/T)}, the number of layers from the root to the deepest descendant.
\vspace{-0.6em}
\paragraph{Dynamic Metrics.} To characterize the dynamic evolutionary process of the forest in greater detail, we design an additional dynamic metric: \textbf{the Average Error Node Reproduction Rate (Repro)}. Let $V$ denote the set of all error nodes within a given solution, where $|V|$ represents the total number of such nodes. For an arbitrary error node $v \in V$, let $k(v)$ denote the number of direct child error nodes generated by $v$, and let $L_t(v)$ represent the lifespan (i.e., the layer depth) of $v$. The average error node reproduction rate can be expressed as:
\vspace{-0.5em}
\begin{equation}
    \bar{r}
=
\frac{1}{|V|}
\sum_{v \in V}
\frac{k(v)}{\max\!\bigl(L_t(v),\,1\bigr)}
\end{equation}

\begin{table}[t]
\centering
\renewcommand\tabcolsep{5.3pt}
\renewcommand\arraystretch{1.1}

\begingroup
\setlength{\abovetopsep}{0pt}
\setlength{\belowrulesep}{0pt}
\setlength{\aboverulesep}{0pt}
\setlength{\belowbottomsep}{0pt}

\resizebox{\columnwidth}{!}{%
\begin{tabular}{l|ccc|c}
\Xhline{1.2pt}
\rowcolor{MorandiHeader}
\textbf{Dataset} &
\multicolumn{3}{c|}{\textbf{Static Metrics}} &
\textbf{Dynamic Metrics} \\
\rowcolor{MorandiHeader}
& \textbf{FS} & \textbf{N/T} & \textbf{D/T} & \textbf{Repro} \\
\Xhline{1.2pt}

\rowcolor{AltRowColor}
AIME25 &
6.9/\textbf{8.1}\,\darkred{\textbf{\makebox[0pt][r]{$\uparrow$\,}1.2}} &
7.1/\textbf{8.4}\,\darkred{\textbf{\makebox[0pt][r]{$\uparrow$\,}1.3}} &
4.9/\textbf{5.7}\,\darkred{\textbf{\makebox[0pt][r]{$\uparrow$\,}0.8}} &
0.084/\textbf{0.126}\,\darkred{\textbf{\makebox[0pt][r]{$\uparrow$\,}50.0\%}} \\
MATH500 &
3.9/\textbf{5.6}\,\darkred{\textbf{\makebox[0pt][r]{$\uparrow$\,}1.7}} &
5.6/\textbf{7.7}\,\darkred{\textbf{\makebox[0pt][r]{$\uparrow$\,}2.1}} &
3.2/\textbf{4.4}\,\darkred{\textbf{\makebox[0pt][r]{$\uparrow$\,}1.2}} &
0.047/\textbf{0.102}\,\darkred{\textbf{\makebox[0pt][r]{$\uparrow$\,}117.0\%}} \\
\rowcolor{AltRowColor}
GSM8K &
1.1/\textbf{2.1}\,\darkred{\textbf{\makebox[0pt][r]{$\uparrow$\,}1.0}} &
2.1/\textbf{3.7}\,\darkred{\textbf{\makebox[0pt][r]{$\uparrow$\,}1.6}} &
1.5/\textbf{2.0}\,\darkred{\textbf{\makebox[0pt][r]{$\uparrow$\,}0.5}} &
0.009/\textbf{0.027}\,\darkred{\textbf{\makebox[0pt][r]{$\uparrow$\,}200.0\%}} \\
GPQA &
4.2/\textbf{6.1}\,\darkred{\textbf{\makebox[0pt][r]{$\uparrow$\,}1.9}} &
5.1/\textbf{7.6}\,\darkred{\textbf{\makebox[0pt][r]{$\uparrow$\,}2.5}} &
3.4/\textbf{4.9}\,\darkred{\textbf{\makebox[0pt][r]{$\uparrow$\,}1.5}} &
0.052/\textbf{0.107}\,\darkred{\textbf{\makebox[0pt][r]{$\uparrow$\,}105.8\%}} \\

\Xhline{1.2pt}
\end{tabular}%
}
\endgroup
\vspace{-1em}
\caption{Average statistics of error trees across datasets (First Solution / Average Subsequent Solutions).}
\label{tab:metrics}
\vspace{-1.7em}
\end{table}

\vspace{-0.5em}
\paragraph{Evaluation.} To evaluate the static and dynamic metrics of the \foe, we employ Qwen3-8B (thinking) on AIME25, MATH500, GSM8K, and GPQA datasets. The results are presented in Table \ref{tab:metrics}.
\vspace{-0.5em}
\paragraph{Obs.\ding{183} The \foe in \first exhibits a significantly smaller scale and a slower average error node reproduction rate compared to that in \subs.} In terms of \textit{static} metrics, the forest size in \first is substantially smaller than in \subs (6.9 vs. 8.1). Specifically, the forest depth in \subs exceeds that in \first by approximately $16.3\%$. This implies that root errors in \subs propagate over longer durations and exert a broader impact. Additionally, we observe a higher count of average error nodes per tree within \subs (7.1 vs. 8.4), indicating that errors accumulate increasingly as the reasoning process progresses. Regarding \textit{dynamic} metrics, we find that the average error node reproduction rate in \first is $33.3\%$ lower than in \subs. This suggests that error nodes in \first generate offspring at a slower pace, further implying greater controllability over error propagation within \first.

\vspace{-0.5em}
\subsection{The Reason for Node Generation}
\label{sec:foe-3}
\vspace{-0.3em}
Following the above analysis, a natural question arises: \textit{How are error nodes within \foe generated?} 
\vspace{-1.7em}
\paragraph{Insufficiency of individual metrics.} Drawing on prior work linking entropy to uncertainty \cite{farquhar2024detecting}, we analyze node generation via entropy and its variance. However, high entropy may stem from benign alternatives rather than errors, while high variance can simply reflect valid reasoning transitions. Consequently, neither signal alone suffices; instead, it is their joint behavior that effectively characterizes the emergence of errors, particularly structural root nodes.

\begin{figure}[!t]
\vspace{-2em}
\centering
\includegraphics[width=0.9\linewidth]{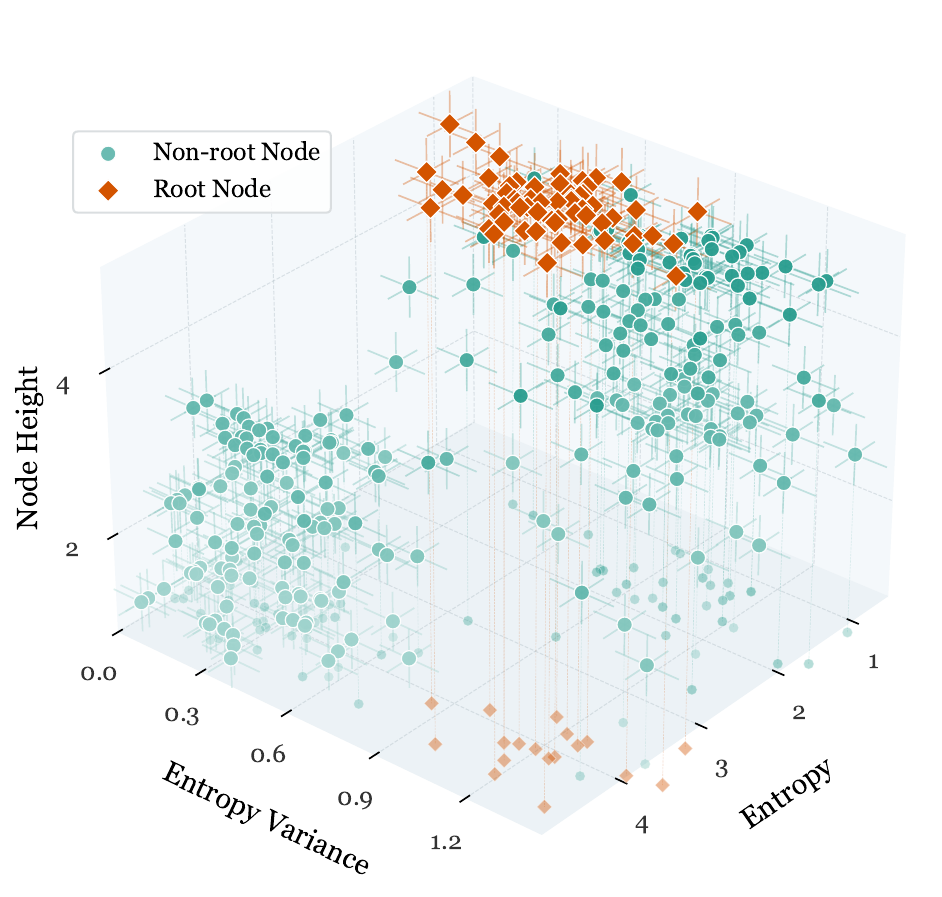} 
\vspace{-1em}
\caption{
   Distribution of various node types with respect to entropy and entropy variance. Experiments were conducted using the Qwen3-8B model on BS-17K-subset.}  
\label{fig:error-node}
\vspace{-1.5em}
\end{figure}

\vspace{-0.5em}
\paragraph{Empirical Substantiate.}
We conduct our experiments on a subset of Bespoke-Stratos-17k~\citep{bespoke_stratos}. Specifically, we select the mathematics and science tasks to construct a subset, denoted as BS-17k-subset. We locate each error node at its first occurrence $t(e)$ and extract the corresponding entropy features $(h_{t(e)}, v_{t(e)})$ using a window length of $L{=}15$, then partition the events into four percentile-based regions: low-low, high-$h$ only, high-$v$ only, and high-high.
For each region, we compute the node-trigger rate (NTR), root-trigger rate (RTR), and average node depth (AND).
\vspace{-0.5em}
\paragraph{Obs.\ding{184} Higher simultaneous levels of entropy and entropy variance correlate with a greater likelihood of root node generation; notably, the probability of root node emergence in \first is lower than in \subs.} Results in Figure~\ref{fig:error-node} show that high-$h$ alone mainly increases the frequency of errors but yields mostly shallow nodes, while high-$v$ alone shifts errors to higher levels without maximizing root-node occurrence; in contrast, the high--high region exhibits the highest RTR, with \first\ consistently achieving a lower RTR than \subs\ under the same condition.
Moreover, this trend is robust to the window choice and remains stable when sweeping $L\in[10,20]$.
Further detailed analysis is provided in Appendix \ref{sec:entropy}.

\vspace{-0.5em}
\subsection{LRMs trys to clear \foe through Reflection}
\label{sec:foe-4}
\begin{figure}[!t]
\centering
\includegraphics[width=1.0\linewidth]{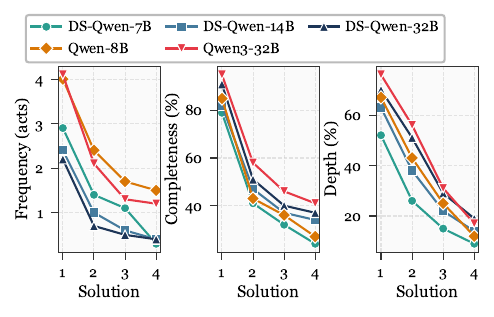} 
\vspace{-2.5em}
\caption{
    Evaluation of intra-solution reflection metrics.}  
\label{fig:reflection}
\vspace{-1.7em}
\end{figure}

\vspace{-0.3em}
One might argue that despite the continuous reproduction of error nodes, models trained via RL possess intrinsic self-reflection capabilities, enabling them to self-prune error nodes. But is this truly the case? Unfortunately, we find that the portion of the FoE cleared through self-reflection is extremely limited. To demonstrate this, we examine two distinct levels: \textit{\underline{intra}}- and \textit{\underline{inter}}-solution.
\vspace{-0.7em}
\paragraph{Intra-solution.} At the intra-solution level, we examine the reflection behaviors inherent to each solution, focusing on three key axes: \textbf{(i) frequency}, denoting the number of reflective instances; \textbf{(ii) completeness}, which assesses whether the reflection is structurally complete, incomplete reflections often involve initiating a reflective thought only to prematurely resume reasoning; and \textbf{(iii) depth}, which gauges the profundity of error detection, shallow reflections typically identify superficial errors while failing to uncover underlying issues. The formal definitions and measurement methods for all axes are detailed in Appendix \ref{sec:reflection}.
\vspace{-0.5em}
\paragraph{Obs.\ding{185} \subs exhibits inferior reflection capabilities and a diminished capacity for error correction compared to \first.} Figure~\ref{fig:reflection} presents large-scale experimental results across multiple backbones on BS-17k-subset.  We observe a significant downward trend in the frequency, completeness, and depth of reflection within the \subs across models of nearly all sizes. For instance, with Qwen-8B-thinking, comparing the first to the last solution reveals a $62.5\%$ reduction in reflection frequency, and a substantial decrease in both completeness ($68.2\%$) and depth ($82.1\%$). This indicates that \subs possesses weaker reflective and error-correction capabilities. Consequently, this suggests that if we aim to leverage self-reflection for error correction in LRMs, we should prioritize incentivizing self-reflection within \first.

\vspace{-0.5em}
\paragraph{Inter-solution.} At the inter-solution level, we investigate whether retained error nodes exert cross-solution influence when self-reflection fails. To demonstrate this, we facilitate model reflection by manually injecting prompts that signal errors within the context. This yields three distinct correction categories: \textbf{(i) True Correction}, where the model successfully identifies and rectifies the error; \textbf{(ii) Refusal to Correct}, involving either explicit denial or passive retention of the error; and \textbf{(iii) Fake Correction}, the most insidious type, characterized by superficial textual edits while subsequent reasoning persists on the original erroneous logic.

\begin{figure}[!t]
\centering
\includegraphics[width=1.0\linewidth]{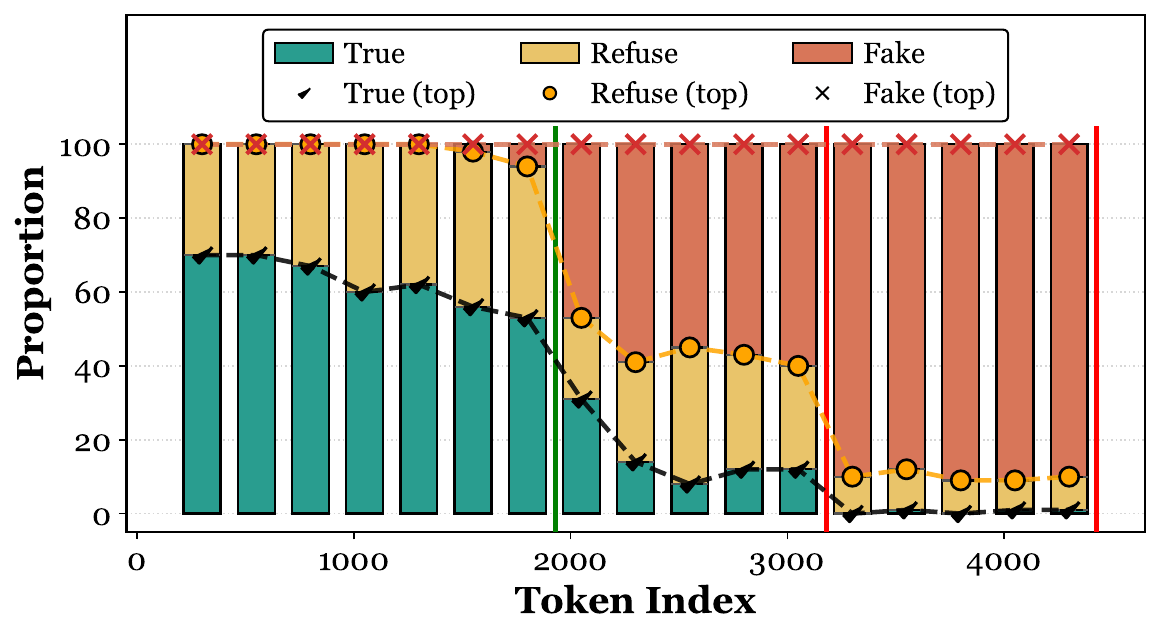} 
\vspace{-2.2em}
\caption{
    Average distribution of correction types (True, Refuse, Fake) on BS-17k-subset + Qwen3-8B. We manually inject error-signaling prompts to probe early-stage errors ($<20\%$). The green and red vertical lines mark the completion of \first and \subs, respectively.}  
\label{fig:error_type}
\vspace{-1.2em}
\end{figure}

\vspace{-0.5em}
\paragraph{Obs.\ding{186} In \subs, fake correction and refusal to correct displace true correction as the dominant behaviors compared to \first.} 
As shown in Figure~\ref{fig:error_type}, we observe that in \first, true correction dominates with a $67.1\%$ share, while fake correction and refusal to correct constitute a minority ($32.4\%$ and $0.5\%$, respectively). Conversely, the scenario is nearly inverted for \subs, fake correction and refusal to correct surge to alarmingly high proportions of $64.2\%$ and $31.1\%$. In the final solution, the behavior is essentially fake correction.

\vspace{-0.7em}
\section{Method}
\vspace{-0.5em}
In this section, we \textbf{(i)} outline the motivation behind our method design, which is grounded in the in-depth analysis of \foe and the extensive observations discussed previously ($\blacktriangleright$ Section \ref{sec:method-1}); and \textbf{(ii)} elaborate on our proposed \ourmethod, a self-guided efficient reasoning method ($\blacktriangleright$ Section \ref{sec:method-2} \& \ref{sec:method-3}).

\vspace{-0.5em}
\subsection{Motivation}
\label{sec:method-1}
\vspace{-0.3em}
\paragraph{Mot.\ding{182} Refining \first.} Our analysis in Section \ref{sec:foe-3} reveals that \first still exhibits instances of simultaneous high entropy and high entropy variance, leading to the genesis of root error nodes, especially in complex tasks. Motivated by the insight from Section \ref{sec:foe-1} that eliminating the root can suppress the growth of the entire \foe, we implement an intervention at segments characterized by high entropy and variance within \first.
\vspace{-0.5em}
\paragraph{Mot.\ding{183} Discarding \subs.} Drawing from our analysis of \foe and the \pheno phenomenon, we realize that the potential risks introduced by \subs far outweigh their possible benefits. Furthermore, generating \subs results in a substantial waste of computational resources. These factors inspire us to discard \subs entirely.

\vspace{-0.7em}
\subsection{Refining \first.}
\label{sec:method-2}
\vspace{-0.3em}
Building on the finding in Section \ref{sec:foe-3} that the co-occurrence of high entropy and high entropy variance precipitates error generation, we intervene at these critical junctures. Specifically, we maintain a sliding window of length $L$ and monitor two key statistics: (1) the entropy variance of tokens within the window; and (2) the average of the maximum Top-$K$ entropy values within the window. We trigger the intervention mechanism when the variance of the current window exceeds the threshold $T$ and the entropy of the next token surpasses the recorded average. Drawing inspiration from classifier-free guidance (CFG)\cite{ho2022classifierfreediffusionguidance}, we append a concise negative prompt (e.g., \texttt{``No, I made a mistake.''}) after the stored KV cache to generate a negative sampling branch. We then extract the logits of the next generated token, which represent the distribution steering the model toward an erroneous trajectory. Finally, we implement the intervention by subtracting these negative logits:
\vspace{-0.5em}
\begin{equation}
\begin{split}
\log \hat{P}(x_t \mid \mathcal{Q}, H_{<t})
= \log P(x_t \mid \mathcal{Q}, H_{<t}) \\
- \alpha \log P\!\bigl(x_t \mid \mathcal{Q},\; H_{<t} \circ H_{\text{neg}}\bigr),
\end{split}
\end{equation}
where $\alpha$ is a tunable scaling coefficient, $\mathcal{Q}$ denotes the input question, and $H$ represents the context. $P(\cdot)$ and $\hat{P}(\cdot)$ denote the pre- and post-intervention prediction probabilities, respectively.

\begin{table*}[!t]
\centering
\renewcommand\tabcolsep{5.3pt}
\renewcommand\arraystretch{1.1}

\resizebox{0.93\linewidth}{!}{
\begin{tabular}{c|l|cc|cc|cc|cc|cc}
\Xhline{1.2pt}
\rowcolor{MorandiHeader}
\textbf{Model} & \textbf{Method} &
\multicolumn{2}{c|}{\textbf{AIME24}} &
\multicolumn{2}{c|}{\textbf{AIME25}} &
\multicolumn{2}{c|}{\textbf{MATH500}} &
\multicolumn{2}{c|}{\textbf{GSM8K}} &
\multicolumn{2}{c}{\textbf{GPQA-Diamond}} \\
\rowcolor{MorandiHeader}
& &
\textbf{Pass@1$\uparrow$} & \textbf{Token$\downarrow$} &
\textbf{Pass@1$\uparrow$} & \textbf{Token$\downarrow$} &
\textbf{Pass@1$\uparrow$} & \textbf{Token$\downarrow$} &
\textbf{Pass@1$\uparrow$} & \textbf{Token$\downarrow$} &
\textbf{Pass@1$\uparrow$} & \textbf{Token$\downarrow$} \\
\Xhline{1.2pt}

\multirow{10}{*}{\rotatebox{90}{Qwen3-8B-thinking}}

& Vanilla &
  70.0 & 11125 &
  61.1 & 12490 &
  95.9 & 4486 &
  94.1 & 1573 &
  58.8 & 6638 \\

& \gcell{DEER} &
  \gcell{71.1} & \gcell{7903} &
  \gcell{58.9} & \gcell{9271} &
  \gcell{95.6} & \gcell{2889} &
  \gcell{94.3} & \gcell{711} &
  \gcell{56.9} & \gcell{\textbf{2926}} \\

& DAST &
  67.8 & 5964 &
  60.0 & \textbf{6055} &
  96.0 & \underline{2158} &
  94.1 & \underline{483} &
  58.4 & \underline{3007} \\

& \gcell{Think or Not} &
  \gcell{68.9} & \gcell{\underline{5387}} &
  \gcell{\underline{62.2}} & \gcell{6674} &
  \gcell{96.3} & \gcell{2408} &
  \gcell{94.2} & \gcell{599} &
  \gcell{59.1} & \gcell{3678} \\

& AlphaOne &
  \underline{73.3} & 8343 &
  \textbf{63.3} & 8607 &
  \underline{96.7} & 4311 &
  94.5 & 1095 &
  \underline{59.8} & 6002 \\

& \gcell{RL + LP} &
  \gcell{71.1} & \gcell{6986} &
  \gcell{58.9} & \gcell{7639} &
  \gcell{96.6} & \gcell{2692} &
  \gcell{94.1} & \gcell{823} &
  \gcell{59.4} & \gcell{3139} \\

& GRPO &
  72.2 & 10931 &
  \underline{62.2} & 11098 &
  \underline{96.7} & 4225 &
  94.4 & 1373 &
  59.1 & 7472 \\

& \gcell{S-GRPO} &
  \gcell{\underline{73.3}} & \gcell{6771} &
  \gcell{61.1} & \gcell{7331} &
  \gcell{96.5} & \gcell{2576} &
  \gcell{\underline{94.6}} & \gcell{775} &
  \gcell{59.6} & \gcell{3046} \\

& \ourmethod &
  \textbf{75.6} & \textbf{5209} &
  \textbf{63.3} & \underline{6203} &
  \textbf{97.0} & \textbf{2029} &
  \textbf{95.0} & \textbf{465} &
  \textbf{60.1} & 3309 \\

& \dcell{$\Delta$} &
  \dcell{\darkred{\textbf{5.6}}} & \dcell{\darkblue{\textbf{53.1\%}}} &
  \dcell{\darkred{\textbf{2.2}}} & \dcell{\darkblue{\textbf{50.3\%}}} &
  \dcell{\darkred{\textbf{1.1}}} & \dcell{\darkblue{\textbf{54.8\%}}} &
  \dcell{\darkred{\textbf{0.9}}} & \dcell{\darkblue{\textbf{70.4\%}}} &
  \dcell{\darkred{\textbf{1.3}}} & \dcell{\darkblue{\textbf{50.2\%}}} \\

\hline

\multirow{10}{*}{\rotatebox{90}{Qwen3-32B-thinking}}
& Vanilla &
  77.8 & 10677 &
  68.9 & 11589 &
  96.8 & 4318 &
  94.3 & 1435 &
  65.3 & 5475 \\
& \gcell{DEER} &
  \gcell{74.4} & \gcell{7894} &
  \gcell{66.7} & \gcell{8417} &
  \gcell{96.4} & \gcell{2733} &
  \gcell{94.1} & \gcell{792} &
  \gcell{64.8} & \gcell{\underline{2247}} \\
& Think or Not &
  76.7 & 6173 &
  65.6 & 6772 &
  97.1 & \underline{2179} &
  94.5 & 641 &
  64.5 & \textbf{1713} \\
& \gcell{AlphaOne} &
  \gcell{\underline{78.9}} & \gcell{8007} &
  \gcell{\underline{71.1}} & \gcell{8569} &
  \gcell{\underline{97.8}} & \gcell{3170} &
  \gcell{94.4} & \gcell{1090} &
  \gcell{\textbf{66.8}} & \gcell{5591} \\
& DAST &
  77.8 & \underline{5981} &
  68.9 & 6504 &
  96.8 & 2417 &
  94.2 & \underline{609} &
  65.7 & 2918 \\
& \gcell{RL + LP} &
  \gcell{76.7} & \gcell{6238} &
  \gcell{66.7} & \gcell{6854} &
  \gcell{97.2} & \gcell{2701} &
  \gcell{\textbf{94.7}} & \gcell{822} &
  \gcell{66.0} & \gcell{4362} \\
& GRPO &
  \underline{78.9} & 11934 &
  70.0 & 12487 &
  97.1 & 4641 &
  94.5 & 1558 &
  66.2 & 6151 \\
& \gcell{S-GRPO} &
  \gcell{77.8} & \gcell{6040} &
  \gcell{\underline{71.1}} & \gcell{\underline{6389}} &
  \gcell{97.3} & \gcell{2566} &
  \gcell{\underline{94.6}} & \gcell{809} &
  \gcell{\underline{66.5}} & \gcell{\underline{4151}} \\
& \ourmethod &
  \textbf{80.0} & \textbf{5793} &
  \textbf{72.2} & \textbf{5908} &
  \textbf{97.9} & \textbf{1995} &
  \underline{94.6} & \textbf{443} &
  \textbf{66.8} & 2477 \\
& \dcell{$\Delta$} &
  \dcell{\darkred{\textbf{2.2}}}  & \dcell{\darkblue{\textbf{45.7\%}}} &
  \dcell{\darkred{\textbf{3.3}}}  & \dcell{\darkblue{\textbf{49.0\%}}} &
  \dcell{\darkred{\textbf{1.1}}}  & \dcell{\darkblue{\textbf{53.8\%}}} &
  \dcell{\darkred{\textbf{0.3}}}  & \dcell{\darkblue{\textbf{69.1\%}}} &
  \dcell{\darkred{\textbf{1.5}}}  & \dcell{\darkblue{\textbf{54.8\%}}} \\
\hline

\multirow{10}{*}{\rotatebox{90}{DpSk-R1-Distill-Qwen-7B}}
& Vanilla &
  54.4 & 10438 &
  43.3 & 11454 &
  91.8 & 2887 &
  92.4 & 442 &
  49.2 & 8016 \\
& \gcell{DEER} &
  \gcell{53.3} & \gcell{7197} &
  \gcell{42.2} & \gcell{8261} &
  \gcell{92.4} & \gcell{1494} &
  \gcell{89.7} & \gcell{297} &
  \gcell{47.3} & \gcell{4423} \\
& Think or Not &
  52.2 & \textbf{4341} &
  38.9 & \textbf{4760} &
  92.0 & \textbf{1103} &
  92.9 & \textbf{264} &
  47.0 & 3390 \\
& \gcell{AlphaOne} &
  \gcell{55.6} & \gcell{8224} &
  \gcell{44.4} & \gcell{8921} &
  \gcell{92.5} & \gcell{3791} &
  \gcell{93.4} & \gcell{459} &
  \gcell{\underline{50.5}} & \gcell{8591} \\
& DAST &
  55.6 & 7258 &
  41.1 & 7904 &
  91.6 & 1330 &
  91.8 & 301 &
  48.8 & 3635 \\
& \gcell{RL + LP} &
  \gcell{52.2} & \gcell{5693} &
  \gcell{43.3} & \gcell{6255} &
  \gcell{93.4} & \gcell{\underline{1322}} &
  \gcell{92.5} & \gcell{291} &
  \gcell{\underline{50.3}} & \gcell{\underline{3209}} \\
& GRPO &
  \underline{56.7} & 11673 &
  \underline{45.6} & 12006 &
  \underline{93.9} & 2873 &
  92.1 & 275 &
  49.8 & 8890 \\
& \gcell{S-GRPO} &
  \gcell{54.4} & \gcell{5094} &
  \gcell{44.4} & \gcell{5794} &
  \gcell{93.2} & \gcell{1204} &
  \gcell{\underline{93.7}} & \gcell{\underline{297}} &
  \gcell{49.5} & \gcell{\textbf{3107}} \\
& \ourmethod &
  \textbf{57.8} & \textbf{4293} &
  \textbf{47.8} & \underline{5690} &
  \textbf{94.1} & \underline{1187} &
  \textbf{94.1} & \underline{271} &
  \textbf{51.2} & \underline{4109} \\
& \dcell{$\Delta$} &
  \dcell{\darkred{\textbf{3.4}}}  & \dcell{\darkblue{\textbf{58.9\%}}} &
  \dcell{\darkred{\textbf{4.5}}}  & \dcell{\darkblue{\textbf{50.3\%}}} &
  \dcell{\darkred{\textbf{2.3}}}  & \dcell{\darkblue{\textbf{58.9\%}}} &
  \dcell{\darkred{\textbf{1.7}}}  & \dcell{\darkblue{\textbf{38.7\%}}} &
  \dcell{\darkred{\textbf{2.0}}}  & \dcell{\darkblue{\textbf{48.7\%}}} \\
\hline

\multirow{10}{*}{\rotatebox{90}{DpSk-R1-Distill-Qwen-32B}}
& Vanilla &
  70.0 & 7873 &
  58.9 & 8906 &
  93.3 & 2337 &
  93.9 & 438 &
  60.8 & 6027 \\
& \gcell{DEER} &
  \gcell{68.9} & \gcell{6461} &
  \gcell{57.8} & \gcell{8008} &
  \gcell{91.8} & \gcell{1697} &
  \gcell{92.5} & \gcell{290} &
  \gcell{58.9} & \gcell{4611} \\
& Think or Not &
  67.8 & \underline{3993} &
  58.9 & \textbf{4711} &
  92.9 & 1410 &
  93.3 & 247 &
  60.4 & 3706 \\
& \gcell{AlphaOne} &
  \gcell{72.2} & \gcell{8210} &
  \gcell{61.1} & \gcell{8994} &
  \gcell{\underline{94.4}} & \gcell{3037} &
  \gcell{94.1} & \gcell{433} &
  \gcell{\underline{61.3}} & \gcell{6771} \\
& DAST &
  70.0 & 5802 &
  54.4 & 6647 &
  93.5 & 1421 &
  93.9 & 266 &
  60.3 & 4048 \\
& \gcell{RL + LP} &
  \gcell{\underline{71.1}} & \gcell{5492} &
  \gcell{\underline{62.2}} & \gcell{6124} &
  \gcell{93.1} & \gcell{\underline{1379}} &
  \gcell{\textbf{94.7}} & \gcell{\underline{239}} &
  \gcell{60.4} & \gcell{3596} \\
& GRPO &
  \underline{73.3} & 8389 &
  \underline{62.2} & 8990 &
  93.4 & 3012 &
  94.4 & 420 &
  60.4 & 7123 \\
& \gcell{S-GRPO} &
  \gcell{70.0} & \gcell{4906} &
  \gcell{60.0} & \gcell{5334} &
  \gcell{94.2} & \gcell{1556} &
  \gcell{\textbf{94.7}} & \gcell{269} &
  \gcell{\underline{61.3}} & \gcell{\textbf{3119}} \\
& \ourmethod &
  \textbf{74.4} & \textbf{3898} &
  \textbf{63.3} & \underline{5018} &
  \textbf{95.2} & \textbf{1347} &
  \underline{94.6} & \textbf{209} &
  \textbf{61.8} & \underline{3497} \\
& \dcell{$\Delta$} &
  \dcell{\darkred{\textbf{4.4}}}  & \dcell{\darkblue{\textbf{50.5\%}}} &
  \dcell{\darkred{\textbf{4.4}}}  & \dcell{\darkblue{\textbf{43.7\%}}} &
  \dcell{\darkred{\textbf{1.9}}}  & \dcell{\darkblue{\textbf{42.4\%}}} &
  \dcell{\darkred{\textbf{0.7}}}  & \dcell{\darkblue{\textbf{52.3\%}}} &
  \dcell{\darkred{\textbf{1.0}}}  & \dcell{\darkblue{\textbf{42.0\%}}} \\
\hline

\multirow{10}{*}{\rotatebox{90}{DeepSeek-R1-Distill-Llama-8B}}
& Vanilla &
  45.6 & 10798.9 &
  28.9 & 11548.2 &
  86.2 & 3635 &
  92.3 & 606 &
  46.3 & 8341 \\
& \gcell{DEER} &
  \gcell{\underline{46.7}} & \gcell{8001} &
  \gcell{\underline{31.1}} & \gcell{8936} &
  \gcell{86.5} & \gcell{2171} &
  \gcell{89.6} & \gcell{394} &
  \gcell{45.5} & \gcell{4152} \\
& Think or Not &
  44.4 & 6761 &
  30.0 & 7158 &
  87.4 & \textbf{1954} &
  92.5 & \underline{257} &
  46.8 & 3729 \\
& \gcell{AlphaOne} &
  \gcell{\textbf{47.8}} & \gcell{8339} &
  \gcell{\textbf{34.4}} & \gcell{9005} &
  \gcell{89.1} & \gcell{3804} &
  \gcell{\underline{93.1}} & \gcell{598} &
  \gcell{\underline{47.6}} & \gcell{8569} \\
& DAST &
  45.6 & 8246 &
  \underline{32.2} & 8438 &
  87.0 & 2458 &
  91.9 & 388 &
  46.1 & 4410 \\
& \gcell{RL + LP} &
  \gcell{45.6} & \gcell{5333} &
  \gcell{30.0} & \gcell{5897} &
  \gcell{89.4} & \gcell{2290} &
  \gcell{92.3} & \gcell{446} &
  \gcell{45.3} & \gcell{\textbf{3299}} \\
& GRPO &
  44.4 & 11312 &
  30.0 & 11987 &
  \underline{89.6} & 3309 &
  92.9 & 571 &
  46.6 & 8783 \\
& \gcell{S-GRPO} &
  \gcell{45.6} & \gcell{\underline{4809}} &
  \gcell{31.1} & \gcell{\underline{5426}} &
  \gcell{89.1} & \gcell{2195} &
  \gcell{\textbf{93.2}} & \gcell{432} &
  \gcell{47.0} & \gcell{3624} \\
& \ourmethod &
  \textbf{47.8} & \textbf{4507} &
  \textbf{34.4} & \textbf{5039} &
  \textbf{90.1} & \underline{2009} &
  \underline{93.1} & \underline{283} &
  \textbf{47.8} & \underline{3593} \\
& \dcell{$\Delta$} &
  \dcell{\darkred{\textbf{2.2}}}  & \dcell{\darkblue{\textbf{58.3\%}}} &
  \dcell{\darkred{\textbf{5.5}}}  & \dcell{\darkblue{\textbf{56.4\%}}} &
  \dcell{\darkred{\textbf{3.9}}}  & \dcell{\darkblue{\textbf{44.7\%}}} &
  \dcell{\darkred{\textbf{0.8}}}  & \dcell{\darkblue{\textbf{53.3\%}}} &
  \dcell{\darkred{\textbf{1.5}}}  & \dcell{\darkblue{\textbf{56.9\%}}} \\
\hline

\multirow{10}{*}{\rotatebox{90}{DeepSeek-R1-Distill-Llama-70B}}
& Vanilla &
  68.9 & 7766 &
  47.8 & 8909 &
  94.1 & 2433 &
  94.0 & 432 &
  64.5 & 5881 \\
& \gcell{DEER} &
  \gcell{70.0} & \gcell{6829} &
  \gcell{48.9} & \gcell{7478} &
  \gcell{92.3} & \gcell{1817} &
  \gcell{93.3} & \gcell{277} &
  \gcell{63.1} & \gcell{4502} \\
& Think or Not &
  71.1 & \underline{4005} &
  46.7 & \underline{4414} &
  94.4 & 1360 &
  93.7 & \underline{241} &
  64.6 & \underline{3544} \\
& \gcell{AlphaOne} &
  \gcell{\underline{72.2}} & \gcell{7873} &
  \gcell{\underline{50.0}} & \gcell{8689} &
  \gcell{\underline{95.3}} & \gcell{2009} &
  \gcell{94.2} & \gcell{427} &
  \gcell{65.3} & \gcell{4322} \\
& DAST &
  67.8 & 5115 &
  45.6 & 5932 &
  94.2 & 1563 &
  93.9 & \textbf{239} &
  63.5 & 4026 \\
& \gcell{RL + LP} &
  \gcell{66.7} & \gcell{5304} &
  \gcell{45.6} & \gcell{5967} &
  \gcell{93.8} & \gcell{\underline{1149}} &
  \gcell{94.4} & \gcell{255} &
  \gcell{65.0} & \gcell{4101} \\
& GRPO &
  \underline{72.2} & 8109 &
  \textbf{51.1} & 8715 &
  94.5 & 3167 &
  \underline{94.7} & 473 &
  64.8 & 6274 \\
& \gcell{S-GRPO} &
  \gcell{70.0} & \gcell{5002} &
  \gcell{48.9} & \gcell{5896} &
  \gcell{95.1} & \gcell{1252} &
  \gcell{94.5} & \gcell{249} &
  \gcell{\underline{65.2}} & \gcell{4001} \\
& \ourmethod &
  \textbf{73.3} & \textbf{3974} &
  \underline{50.0} & \textbf{4303} &
  \textbf{96.2} & \textbf{932} &
  \textbf{94.9} & 269 &
  \textbf{66.2} & \underline{3563} \\
& \dcell{$\Delta$} &
  \dcell{\darkred{\textbf{4.4}}}  & \dcell{\darkblue{\textbf{48.8\%}}} &
  \dcell{\darkred{\textbf{2.2}}}  & \dcell{\darkblue{\textbf{51.7\%}}} &
  \dcell{\darkred{\textbf{2.1}}}  & \dcell{\darkblue{\textbf{61.7\%}}} &
  \dcell{\darkred{\textbf{0.9}}}  & \dcell{\darkblue{\textbf{37.7\%}}} &
  \dcell{\darkred{\textbf{1.7}}}  & \dcell{\darkblue{\textbf{39.4\%}}} \\
\hline

\Xhline{1.2pt}
\end{tabular}
}
\vspace{-0.7em}
\caption{Main results. Best results are highlighted in \textbf{bold}, with runners-up \underline{underlined}.}
\label{tab:consolidated_results_wide}
\vspace{-1.0em}
\end{table*}

\vspace{-0.5em}
\subsection{Discarding \subs}
\label{sec:method-3}
\vspace{-0.3em}
Our method leverages two empirical observations under high-temperature sampling: (1) \textbf{\textit{Convergence}}: induced answers evolve from early-stage diversity to a single dominant mode as reasoning progresses; (2) \textbf{\textit{Robustness}}: unlike brittle intermediate states, stabilized induced answers remain consistent across different probe prompts. Further details are provided in Appendix \ref{sec:early-stop}.
\vspace{-0.7em}
\paragraph{Dual-consistency Mechanism.} Based on this, we trigger a periodic probe every $K$ steps. We pause generation and concurrently inject $M$ distinct prompts to induce interim answers (each prompt with $N$ parallel samples). To facilitate reliable extraction, we employ vLLM's guided decoding to constrain generation to a predefined set of valid answer tokens (e.g., option letters, numerical digits, or symbols). We trigger an early exit only if a \textit{dual-consistency} condition is met: (i) \textbf{Internal Consistency}, where the dominant answer within each prompt template must appear with a frequency $\ge P\%$; and (ii) \textbf{Cross-Prompt Agreement}, where the dominant answers across all $M$ templates must be identical, ensuring the emerging answer is a robust solution rather than a prompt-sensitive artifact.

\vspace{-0.7em}
\section{Experiments}
\vspace{-0.5em}
\subsection{Experimental Setup}
\vspace{-0.3em}
\paragraph{Backbones.} We conduct experiments using representative open-source LRMs with diverse architectures from different families. \textbf{I) \textit{Qwen family}}, Qwen3-thinking series (8B \& 32B)~\citep{yang2025qwen3technicalreport}, DeepSeek-R1-Distill-Qwen series (7B \& 32B); \textbf{II) \textit{Llama family}}, DeepSeek-R1-Distill-Llama series (8B \& 70B).
\vspace{-0.5em}
\paragraph{Baselines.} 

We compare \ourmethod against a comprehensive set of baselines categorized into three groups: \textbf{I) Vanilla Model}, the original backbone LRM; \textbf{II) Training-free methods}, including DEER~\citep{yang2025dynamicearlyexitreasoning}, Think or Not~\citep{yong2025think}, and AlphaOne~\citep{zhang-etal-2025-alphaone}; \textbf{III) RL-based strategies}, including DAST~\citep{shen-etal-2025-dast}, RL + Length Penalty~\citep{arora2025training}, GRPO~\citep{deepseekai2025deepseekr1incentivizingreasoningcapability}, and S-GRPO~\citep{dai2025sgrpo}. Detailed settings of baselines are provided in Appendix \ref{sec:baseline}.
\vspace{-0.5em}
\paragraph{Benchmarks.} We conduct extensive evaluations of \ourmethod on five benchmarks spanning two complex reasoning domains: 
\textbf{I) Mathematical Reasoning}, including GSM8K~\cite{cobbe2021gsm8k}, MATH500~\cite{lightman2023letsverifystepstep}, AIME 2024, and AIME 2025; 
and \textbf{II) Scientific Reasoning}, specifically GPQA-Diamond~\cite{rein2023gpqagraduatelevelgoogleproofqa}.
\vspace{-0.5em}
\paragraph{Implementation details.} 
\label{sec:implementation}
We use a sampling temperature of $0.6$ and top-$p$ of $0.95$, reporting Pass@1 averaged over three runs.
For refining \first, we set the sliding-window length $L{=}15$, threshold $T{=}2.4$, and top-$K{=}3$.
For discarding \subs, we probe every 2 steps with $M{=}4$ distinct prompt templates, each with $N{=}12$ parallel samples and a consistency rate of $60\%$.

\definecolor{HeaderLightGreen}{RGB}{232,245,236}
\definecolor{MetricGreen}{RGB}{56,153,92}
\definecolor{MetricGreenDark}{RGB}{24,115,62}

\newcommand{\mg}[1]{\textcolor{MetricGreen}{#1}}
\newcommand{\mdg}[1]{\textcolor{MetricGreenDark}{#1}}

\begin{table}[t]
\centering
\renewcommand\tabcolsep{5pt}
\renewcommand\arraystretch{1.1}

\resizebox{\linewidth}{!}{
\begin{tabular}{c|l|ccccc}
\Xhline{1.2pt}
\rowcolor{HeaderLightGreen}
\textbf{Model} & \textbf{Method} &
\multicolumn{5}{c}{\textbf{AIME25}} \\
\rowcolor{HeaderLightGreen}
& &
\textbf{Pass@1 $\uparrow$} &
\textbf{\mg{FS$\downarrow$} } & \textbf{\mg{N/T$\downarrow$}} &
\textbf{\mg{D/T$\downarrow$}} & \textbf{\mdg{Repro$\downarrow$}} \\
\Xhline{1.2pt}

\multirow{6}{*}{
  \rotatebox{90}{
    \fontsize{9.8pt}{11pt}\selectfont
    Qwen3-32B-thinking
  }
}
& \gcell{Vanilla} &
  \gcell{68.9} & \gcell{\mg{6.8}} & \gcell{\mg{7.4}} & \gcell{\mg{5.3}} & \gcell{\mdg{0.081}} \\
& DEER &
  66.7 & \mg{6.3} & \mg{7.2} & \mg{5.4} & \mdg{0.083} \\
& \gcell{RL + LP} &
  \gcell{66.7} & \gcell{\mg{5.9}} & \gcell{\mg{6.9}} & \gcell{\mg{4.7}} & \gcell{\mdg{0.071}} \\
& S-GRRO &
  71.1 & \mg{6.1} & \mg{7.0} & \mg{5.1} & \mdg{0.074} \\
& \gcell{\ourmethod} &
  \gcell{\textbf{72.2}} & \gcell{\textbf{\mg{3.1}}} & \gcell{\textbf{\mg{4.2}}} & \gcell{\textbf{\mg{3.1}}} & \gcell{\textbf{\mdg{0.026}}} \\
& \dcell{$\Delta$} &
  \dcell{\darkred{\textbf{+3.3}}} &
  \dcell{\darkblue{\textbf{\makebox[0pt][r]{$\downarrow$\,}54.4\%}}} &
  \dcell{\darkblue{\textbf{\makebox[0pt][r]{$\downarrow$\,}43.2\%}}} &
  \dcell{\darkblue{\textbf{\makebox[0pt][r]{$\downarrow$\,}41.5\%}}} &
  \dcell{\darkblue{\textbf{\makebox[0pt][r]{$\downarrow$\,}67.9\%}}} \\
\hline

\multirow{6}{*}{
  \rotatebox{90}{
    \fontsize{9.8pt}{11pt}\selectfont
    R1-Qwen-32B
  }
}
& \gcell{Vanilla} &
  \gcell{58.9} & \gcell{\mg{7.0}} & \gcell{\mg{7.8}} & \gcell{\mg{5.8}} & \gcell{\mdg{0.095}} \\
& DEER &
  57.8 & \mg{6.5} & \mg{7.6} & \mg{5.9} & \mdg{0.097} \\
& \gcell{RL + LP} &
  \gcell{62.2} & \gcell{\mg{6.1}} & \gcell{\mg{7.3}} & \gcell{\mg{5.2}} & \gcell{\mdg{0.085}} \\
& S-GRRO &
  60.0 & \mg{6.3} & \mg{7.4} & \mg{5.6} & \mdg{0.088} \\
& \gcell{\ourmethod} &
  \gcell{\textbf{63.3}} & \gcell{\textbf{\mg{3.2}}} & \gcell{\textbf{\mg{4.6}}} & \gcell{\textbf{\mg{3.6}}} & \gcell{\textbf{\mdg{0.040}}} \\
& \dcell{$\Delta$} &
  \dcell{\darkred{\textbf{+4.4}}} &
  \dcell{\darkblue{\textbf{\makebox[0pt][r]{$\downarrow$\,}54.3\%}}} &
  \dcell{\darkblue{\textbf{\makebox[0pt][r]{$\downarrow$\,}41.0\%}}} &
  \dcell{\darkblue{\textbf{\makebox[0pt][r]{$\downarrow$\,}37.9\%}}} &
  \dcell{\darkblue{\textbf{\makebox[0pt][r]{$\downarrow$\,}57.9\%}}} \\
\hline

\multirow{6}{*}{
  \rotatebox{90}{
    \fontsize{9.8pt}{11pt}\selectfont
    R1-Llama-70B
  }
}
& \gcell{Vanilla} &
  \gcell{47.8} & \gcell{\mg{8.6}} & \gcell{\mg{8.3}} & \gcell{\mg{6.4}} & \gcell{\mdg{0.125}} \\
& DEER &
  48.9 & \mg{7.9} & \mg{7.8} & \mg{6.5} & \mdg{0.128} \\
& \gcell{RL + LP} &
  \gcell{45.6} & \gcell{\mg{7.6}} & \gcell{\mg{7.5}} & \gcell{\mg{5.8}} & \gcell{\mdg{0.110}} \\
& S-GRRO &
  48.9 & \mg{7.8} & \mg{7.6} & \mg{6.1} & \mdg{0.115} \\
& \gcell{\ourmethod} &
  \gcell{\textbf{50.0}} & \gcell{\textbf{\mg{3.9}}} & \gcell{\textbf{\mg{4.7}}} & \gcell{\textbf{\mg{3.7}}} & \gcell{\textbf{\mdg{0.040}}} \\
& \dcell{$\Delta$} &
  \dcell{\darkred{\textbf{+2.2}}} &
  \dcell{\darkblue{\textbf{\makebox[0pt][r]{$\downarrow$\,}54.7\%}}} &
  \dcell{\darkblue{\textbf{\makebox[0pt][r]{$\downarrow$\,}43.4\%}}} &
  \dcell{\darkblue{\textbf{\makebox[0pt][r]{$\downarrow$\,}42.2\%}}} &
  \dcell{\darkblue{\textbf{\makebox[0pt][r]{$\downarrow$\,}68.0\%}}} \\
\hline

\Xhline{1.2pt}
\end{tabular}
}
\vspace{-0.8em}
\caption{Results on AIME25 with FoE-related metrics. Additional results are provided in Appendix \ref{sec:FoE-metric}.}
\label{tab:foe_aime25_metric}
\vspace{-1.8em}
\end{table}

\vspace{-0.5em}
\subsection{Main Results}
\vspace{-0.3em}
\paragraph{Obs.\ding{182} \ourmethod strikes the best balance between performance and efficiency across almost all scales.} Compared to eight baselines on five benchmarks (Table \ref{tab:consolidated_results_wide}),
\ourmethod generally improves performance by $0.3\sim5.6$ ($3.2\%\sim19.0\%$) over the Vanilla model while slashing token consumption by $37.7\%\sim70.4\%$. Notably, on DeepSeek-R1-Distill-Llama-8B+AIME25, \ourmethod outperforms all baselines with a $5.5$ score increase and $56.4\%$ token reduction. It even surpasses the strong S-GRPO baseline by $3.3$ in score and $7.1\%$ in efficiency. Furthermore, scalability tests on the DeepSeek-R1-Distill-Llama series reveal that while DAST's performance gain degrades from $+3.3$ (8B) to $-2.2$ (70B) relative to Vanilla, \ourmethod remains robust, achieving gains of $+5.5$ and $+2.2$, respectively.
\vspace{-0.5em}
\paragraph{Obs.\ding{183} \ourmethod achieves substantial pruning of the \foe.} Table \ref{tab:foe_aime25_metric} shows that our approach reduces all \foe metrics by $41.0\%\sim68.0\%$, setting a new standard against baselines. Scrutinizing the results reveals a sharp contrast: even the competitive S-GRPO struggles with \foe mitigation. On DeepSeek-R1-Distill-Qwen-32B, S-GRPO lowers the static D/T metric by a negligible $3.4\%$ and the dynamic Repro metric by $7.4\%$. In comparison, \ourmethod dramatically reduces by $37.9\%$ and $57.9\%$, respectively. By simultaneously shrinking forest size and inhibiting node reproduction, \ourmethod validates the effectiveness of our proposed mechanism.

\vspace{-0.5em}
\subsection{Framework Analysis}
\vspace{-0.3em}
Ablation studies, sensitivity analysis, and Cons@k are detailed in Appendices \ref{sec:ably}, \ref{sec:hyperparam}, and \ref{sec:cons}.

\vspace{-0.7em}
\section{Conclusion}
\vspace{-0.5em}
In this work, we uncover the counter-intuitive \pheno in LRMs, significantly challenging the test-time scaling laws. Through a rigorous analysis, we attribute it to the Forest of Errors (\foe), revealing that reasoning errors scale concurrently with test time, making the first the best. Motivated by these, we introduce \ourmethod, a framework that synergizes \textit{Refining First} to inhibit error growth and \textit{Discarding Subs} to eliminate redundant, error-prone computations. We believe our findings offer valuable insights for further research.

\section*{Limitations}
A potential limitation of \ourmethod is the introduction of additional latency due to extra operations within the decoding process. However, we evaluated this additional latency through a rigorous latency stress test ($\blacktriangleright$ Appendix \ref{sec:stime}). Specifically, we isolated the operational overhead by disabling the early-exit mechanism (i.e., continuing generation even if exit criteria are satisfied) while fully maintaining the periodic injection of probe prompts and conducting inference intervention based on entropy and entropy variance. This worst-case profiling reveals that the additional latency overhead averages only 4.6\%. Since this additional latency is minimal, the time saved by the early-exit mechanism significantly outweighs it. As a result, \ourmethod achieves a net speedup compared to the baseline.

\bibliography{main}

\appendix

\section{Related Work}
\paragraph{Efficient LLM Reasoning.} According to the efficient-reasoning survey of \citet{eff-survey}, existing approaches can be organized into four families: (1) Model-based Efficient Reasoning, (2) Reasoning Output-based Efficient Reasoning, (3) Input Prompts-based Efficient Reasoning, and (4) Efficient Data and Models. \ding{182} \textbf{Model-based} methods internalize conciseness by training: a major thread augments RL with explicit length-aware objectives to curb overlong chains while preserving accuracy (e.g., O1-Pruner~\citep{o1-pruner}, Kimi~k1.5~\citep{kimi}, L1~\citep{l1}, DAST~\citep{shen-etal-2025-dast}, while another line performs SFT~\citep{aurora} on variable-length CoT data, either compressing long traces post hoc (e.g., C3oT~\citep{c3ot}, TokenSkip~\citep{tokenskip}) or eliciting shorter reasoning during generation (e.g., Learn-to-Skip~\citep{learntoskip}, Self-Training~\citep{selftraining}, CoT-Valve~\citep{cotvalve}). \ding{183} \textbf{Reasoning Output-based} methods reduce explicit tokens by altering the reasoning form at inference: latent reasoning compresses intermediate steps into hidden/continuous representations (e.g., Coconut~\citep{coconut}, CCOT~\citep{ccot}, SoftCoT~\citep{softcot}), and dynamic inference allocates computation adaptively via reward-, confidence-, or consistency-guided control (e.g., RSD~\citep{rsd}, Dynasor~\citep{dynasor}, ST-BoN~\citep{st-bon}, LightThinker~\citep{lightthinker}). \ding{184} \textbf{Input Prompts-based} approaches improve efficiency without changing core weights by (i) directly prompting for brevity or budget adherence (e.g., Token-Budget~\citep{token-budget}, Chain-of-Draft~\citep{cod}) and (ii) routing queries to different reasoning modes/models based on prompt attributes such as difficulty (e.g., Sketch-of-Thought~\citep{sot}, RouteLLM~\citep{routellm}, ThinkSwitcher~\citep{thinkswitcher}). \ding{185} Finally, \textbf{Efficient Data and Models} target deployment-oriented efficiency through data-efficient reasoning supervision (e.g., LIMO~\citep{limo}, s1~\citep{s1}, S2R~\citep{s2r}). In contrast to these works, we first identify two critical phenomena: \pheno and \foe. Following a series of in-depth analyses, we introduce \ourmethod, a method grounded in these insights and underpinned by extensive empirical observations.

\section{Additional Experimental Results}

\subsection{Additional Results of FoE-related Metrics}
\label{sec:FoE-metric}

\paragraph{Mathematical Task.} On MATH500 (Table \ref{tab:foe_math500_full}), \ourmethod continues to substantially prune the \foe across all three backbones, while maintaining (and slightly improving) accuracy: Pass@1 increases by $+1.1\sim+2.1$ compared to Vanilla, and all \foe metrics (FS, N/T, D/T, Repro) are reduced by $37.1\%\sim68.0\%$. Importantly, the advantage is not merely from “shorter” traces, but from directly suppressing both the \emph{static} forest structure and the \emph{dynamic} reproduction process. This is evident when contrasting against the strong RL baseline S-GRPO: on DeepSeek-R1-Distill-Qwen-32B, S-GRPO only yields a negligible D/T reduction of $2.9\%$ and a Repro reduction of $7.0\%$, whereas \ourmethod reduces D/T and Repro by $37.1\%$ and $57.9\%$, respectively. The same pattern holds for Qwen3-32B-Thinking and R1-Distill-Llama-70B, indicating that competitive baselines may reach similar Pass@1 but still struggle to mitigate the underlying \foe growth.

\paragraph{Scientific Task.} On GPQA-Diamond (Table \ref{tab:foe_gpqa_full}), the same phenomenon persists under a clear domain shift from math to scientific reasoning: \ourmethod consistently improves Pass@1 by $+1.0\sim+1.7$ and simultaneously reduces all \foe metrics by $38.6\%\sim68.1\%$ relative to Vanilla. Again, S-GRPO exhibits limited \foe mitigation—on DeepSeek-R1-Distill-Qwen-32B, it reduces D/T by only $4.5\%$ and Repro by $7.0\%$, whereas \ourmethod achieves much larger reductions of $38.6\%$ and $57.7\%$, respectively. Together with the AIME25 evidence (Table \ref{tab:foe_aime25_full}), these consistent gains across datasets strengthen the core conclusion in the main text: naively extending test-time exploration does not reliably “fix” earlier mistakes, but instead tends to expand a \foe through both wider/deeper error structures and faster error reproduction. By simultaneously shrinking the forest (FS, N/T, D/T) and inhibiting node reproduction (Repro), \ourmethod offers a direct mechanism-level explanation for why the First solution often remains the best under test-time scaling.

\begin{table}[t]
\centering
\renewcommand\tabcolsep{5pt}
\renewcommand\arraystretch{1.1}

\resizebox{\linewidth}{!}{
\begin{tabular}{c|l|ccccc}
\Xhline{1.2pt}
\rowcolor{HeaderLightGreen}
\textbf{Model} & \textbf{Method} &
\multicolumn{5}{c}{\textbf{AIME25}} \\
\rowcolor{HeaderLightGreen}
& &
\textbf{Pass@1$\uparrow$} &
\textbf{\mg{FS$\downarrow$}} & \textbf{\mg{N/T$\downarrow$}} &
\textbf{\mg{D/T$\downarrow$}} & \textbf{\mdg{Repro$\downarrow$}} \\
\Xhline{1.2pt}

\multirow{6}{*}{
  \rotatebox{90}{
    \fontsize{9.8pt}{11pt}\selectfont
    Qwen3-32B-Thinking
  }
}
& \gcell{Vanilla} &
  \gcell{68.9} & \gcell{\mg{6.8}} & \gcell{\mg{7.4}} & \gcell{\mg{5.3}} & \gcell{\mdg{0.081}} \\
& DEER &
  66.7 & \mg{6.3} & \mg{7.2} & \mg{5.4} & \mdg{0.083} \\
& \gcell{RL + LP} &
  \gcell{66.7} & \gcell{\mg{5.9}} & \gcell{\mg{6.9}} & \gcell{\mg{4.7}} & \gcell{\mdg{0.071}} \\
& S-GRRO &
  71.1 & \mg{6.1} & \mg{7.0} & \mg{5.1} & \mdg{0.074} \\
& \gcell{\ourmethod} &
  \gcell{\textbf{72.2}} & \gcell{\textbf{\mg{3.1}}} & \gcell{\textbf{\mg{4.2}}} & \gcell{\textbf{\mg{3.1}}} & \gcell{\textbf{\mdg{0.026}}} \\
& \dcell{$\Delta$} &
  \dcell{\darkred{\textbf{+3.3}}} &
  \dcell{\darkblue{\textbf{\makebox[0pt][r]{$\downarrow$\,}54.4\%}}} &
  \dcell{\darkblue{\textbf{\makebox[0pt][r]{$\downarrow$\,}43.2\%}}} &
  \dcell{\darkblue{\textbf{\makebox[0pt][r]{$\downarrow$\,}41.5\%}}} &
  \dcell{\darkblue{\textbf{\makebox[0pt][r]{$\downarrow$\,}67.9\%}}} \\
\hline

\multirow{6}{*}{
  \rotatebox{90}{
    \fontsize{9.8pt}{11pt}\selectfont
    R1-Qwen-32B
  }
}
& \gcell{Vanilla} &
  \gcell{58.9} & \gcell{\mg{7.0}} & \gcell{\mg{7.8}} & \gcell{\mg{5.8}} & \gcell{\mdg{0.095}} \\
& DEER &
  57.8 & \mg{6.5} & \mg{7.6} & \mg{5.9} & \mdg{0.097} \\
& \gcell{RL + LP} &
  \gcell{62.2} & \gcell{\mg{6.1}} & \gcell{\mg{7.3}} & \gcell{\mg{5.2}} & \gcell{\mdg{0.085}} \\
& S-GRRO &
  60.0 & \mg{6.3} & \mg{7.4} & \mg{5.6} & \mdg{0.088} \\
& \gcell{\ourmethod} &
  \gcell{\textbf{63.3}} & \gcell{\textbf{\mg{3.2}}} & \gcell{\textbf{\mg{4.6}}} & \gcell{\textbf{\mg{3.6}}} & \gcell{\textbf{\mdg{0.040}}} \\
& \dcell{$\Delta$} &
  \dcell{\darkred{\textbf{+4.4}}} &
  \dcell{\darkblue{\textbf{\makebox[0pt][r]{$\downarrow$\,}54.3\%}}} &
  \dcell{\darkblue{\textbf{\makebox[0pt][r]{$\downarrow$\,}41.0\%}}} &
  \dcell{\darkblue{\textbf{\makebox[0pt][r]{$\downarrow$\,}37.9\%}}} &
  \dcell{\darkblue{\textbf{\makebox[0pt][r]{$\downarrow$\,}57.9\%}}} \\
\hline

\multirow{6}{*}{
  \rotatebox{90}{
    \fontsize{9.8pt}{11pt}\selectfont
    R1-Llama-70B
  }
}
& \gcell{Vanilla} &
  \gcell{47.8} & \gcell{\mg{8.6}} & \gcell{\mg{8.3}} & \gcell{\mg{6.4}} & \gcell{\mdg{0.125}} \\
& DEER &
  48.9 & \mg{7.9} & \mg{7.8} & \mg{6.5} & \mdg{0.128} \\
& \gcell{RL + LP} &
  \gcell{45.6} & \gcell{\mg{7.6}} & \gcell{\mg{7.5}} & \gcell{\mg{5.8}} & \gcell{\mdg{0.110}} \\
& S-GRRO &
  48.9 & \mg{7.8} & \mg{7.6} & \mg{6.1} & \mdg{0.115} \\
& \gcell{\ourmethod} &
  \gcell{\textbf{50.0}} & \gcell{\textbf{\mg{3.9}}} & \gcell{\textbf{\mg{4.7}}} & \gcell{\textbf{\mg{3.7}}} & \gcell{\textbf{\mdg{0.040}}} \\
& \dcell{$\Delta$} &
  \dcell{\darkred{\textbf{+2.2}}} &
  \dcell{\darkblue{\textbf{\makebox[0pt][r]{$\downarrow$\,}54.7\%}}} &
  \dcell{\darkblue{\textbf{\makebox[0pt][r]{$\downarrow$\,}43.4\%}}} &
  \dcell{\darkblue{\textbf{\makebox[0pt][r]{$\downarrow$\,}42.2\%}}} &
  \dcell{\darkblue{\textbf{\makebox[0pt][r]{$\downarrow$\,}68.0\%}}} \\
\hline

\Xhline{1.2pt}
\end{tabular}
}
\caption{Results on AIME25 with FoE-related metrics. }
\label{tab:foe_aime25_full}
\end{table}

\begin{table}[ht]
\centering
\renewcommand\tabcolsep{5pt}
\renewcommand\arraystretch{1.1}

\resizebox{\linewidth}{!}{
\begin{tabular}{c|l|ccccc}
\Xhline{1.2pt}
\rowcolor{HeaderLightGreen}
\textbf{Model} & \textbf{Method} &
\multicolumn{5}{c}{\textbf{MATH500}} \\

\rowcolor{HeaderLightGreen}
& &
\textbf{Pass@1} &
\textbf{\mg{FS$\downarrow$}} & \textbf{\mg{N/T$\downarrow$}} &
\textbf{\mg{D/T$\downarrow$}} & \textbf{\mdg{Repro$\downarrow$}} \\
\Xhline{1.2pt}

\multirow{6}{*}{
  \rotatebox{90}{
    \fontsize{9.8pt}{11pt}\selectfont
    Qwen3-32B-Thinking
  }
}
& \gcell{Vanilla} &
  \gcell{96.8} & \gcell{\mg{4.1}} & \gcell{\mg{4.4}} & \gcell{\mg{3.2}} & \gcell{\mdg{0.049}} \\
& DEER &
  96.4 & \mg{3.8} & \mg{4.3} & \mg{3.2} & \mdg{0.050} \\
& \gcell{RL + LP} &
  \gcell{97.2} & \gcell{\mg{3.5}} & \gcell{\mg{4.1}} & \gcell{\mg{2.8}} & \gcell{\mdg{0.043}} \\
& S-GRRO &
  97.3 & \mg{3.7} & \mg{4.2} & \mg{3.1} & \mdg{0.044} \\
& \gcell{\ourmethod} &
  \gcell{\textbf{97.9}} & \gcell{\textbf{\mg{1.9}}} & \gcell{\textbf{\mg{2.5}}} & \gcell{\textbf{\mg{1.9}}} & \gcell{\textbf{\mdg{0.016}}} \\
& \dcell{$\Delta$} &
  \dcell{\darkred{\textbf{+1.1}}} &
  \dcell{\darkblue{\textbf{\makebox[0pt][r]{$\downarrow$\,}54.4\%}}} &
  \dcell{\darkblue{\textbf{\makebox[0pt][r]{$\downarrow$\,}43.2\%}}} &
  \dcell{\darkblue{\textbf{\makebox[0pt][r]{$\downarrow$\,}41.5\%}}} &
  \dcell{\darkblue{\textbf{\makebox[0pt][r]{$\downarrow$\,}67.9\%}}} \\
\hline

\multirow{6}{*}{
  \rotatebox{90}{
    \fontsize{9.8pt}{11pt}\selectfont
    R1-Qwen-32B
  }
}
& \gcell{Vanilla} &
  \gcell{93.3} & \gcell{\mg{4.2}} & \gcell{\mg{4.7}} & \gcell{\mg{3.5}} & \gcell{\mdg{0.057}} \\
& DEER &
  91.8 & \mg{3.9} & \mg{4.6} & \mg{3.5} & \mdg{0.058} \\
& \gcell{RL + LP} &
  \gcell{93.1} & \gcell{\mg{3.7}} & \gcell{\mg{4.4}} & \gcell{\mg{3.1}} & \gcell{\mdg{0.051}} \\
& S-GRRO &
  94.2 & \mg{3.8} & \mg{4.4} & \mg{3.4} & \mdg{0.053} \\
& \gcell{\ourmethod} &
  \gcell{\textbf{95.2}} & \gcell{\textbf{\mg{1.9}}} & \gcell{\textbf{\mg{2.8}}} & \gcell{\textbf{\mg{2.2}}} & \gcell{\textbf{\mdg{0.024}}} \\
& \dcell{$\Delta$} &
  \dcell{\darkred{\textbf{+1.9}}} &
  \dcell{\darkblue{\textbf{\makebox[0pt][r]{$\downarrow$\,}54.3\%}}} &
  \dcell{\darkblue{\textbf{\makebox[0pt][r]{$\downarrow$\,}41.0\%}}} &
  \dcell{\darkblue{\textbf{\makebox[0pt][r]{$\downarrow$\,}37.9\%}}} &
  \dcell{\darkblue{\textbf{\makebox[0pt][r]{$\downarrow$\,}57.9\%}}} \\
\hline

\multirow{6}{*}{
  \rotatebox{90}{
    \fontsize{9.8pt}{11pt}\selectfont
    R1-Llama-70B
  }
}
& \gcell{Vanilla} &
  \gcell{94.1} & \gcell{\mg{5.2}} & \gcell{\mg{5.0}} & \gcell{\mg{3.8}} & \gcell{\mdg{0.075}} \\
& DEER &
  92.3 & \mg{4.7} & \mg{4.7} & \mg{3.9} & \mdg{0.077} \\
& \gcell{RL + LP} &
  \gcell{93.8} & \gcell{\mg{4.6}} & \gcell{\mg{4.5}} & \gcell{\mg{3.5}} & \gcell{\mdg{0.066}} \\
& S-GRRO &
  95.1 & \mg{4.7} & \mg{4.6} & \mg{3.7} & \mdg{0.069} \\
& \gcell{\ourmethod} &
  \gcell{\textbf{96.2}} & \gcell{\textbf{\mg{2.3}}} & \gcell{\textbf{\mg{2.8}}} & \gcell{\textbf{\mg{2.2}}} & \gcell{\textbf{\mdg{0.024}}} \\
& \dcell{$\Delta$} &
  \dcell{\darkred{\textbf{+2.1}}} &
  \dcell{\darkblue{\textbf{\makebox[0pt][r]{$\downarrow$\,}54.7\%}}} &
  \dcell{\darkblue{\textbf{\makebox[0pt][r]{$\downarrow$\,}43.4\%}}} &
  \dcell{\darkblue{\textbf{\makebox[0pt][r]{$\downarrow$\,}42.2\%}}} &
  \dcell{\darkblue{\textbf{\makebox[0pt][r]{$\downarrow$\,}68.0\%}}} \\
\hline

\Xhline{1.2pt}
\end{tabular}
}
\caption{Results on MATH500 with FoE-related metrics.}
\label{tab:foe_math500_full}
\end{table}

\begin{table}[ht]
\centering
\renewcommand\tabcolsep{5pt}
\renewcommand\arraystretch{1.1}

\resizebox{\linewidth}{!}{
\begin{tabular}{c|l|ccccc}
\Xhline{1.2pt}
\rowcolor{HeaderLightGreen}
\textbf{Model} & \textbf{Method} &
\multicolumn{5}{c}{\textbf{GPQA-Diamond}} \\

\rowcolor{HeaderLightGreen}
& &
\textbf{Pass@1$\uparrow$} &
\textbf{\mg{FS$\downarrow$}} & \textbf{\mg{N/T$\downarrow$}} &
\textbf{\mg{D/T$\downarrow$}} & \textbf{\mdg{Repro$\downarrow$}} \\
\Xhline{1.2pt}

\multirow{6}{*}{
  \rotatebox{90}{
    \fontsize{9.8pt}{11pt}\selectfont
    Qwen3-32B-Thinking
  }
}
& \gcell{Vanilla} &
  \gcell{65.3} & \gcell{\mg{5.1}} & \gcell{\mg{5.6}} & \gcell{\mg{4.0}} & \gcell{\mdg{0.061}} \\
& DEER &
  64.8 & \mg{4.7} & \mg{5.4} & \mg{4.1} & \mdg{0.062} \\
& \gcell{RL + LP} &
  \gcell{66.0} & \gcell{\mg{4.4}} & \gcell{\mg{5.2}} & \gcell{\mg{3.5}} & \gcell{\mdg{0.053}} \\
& S-GRRO &
  66.5 & \mg{4.6} & \mg{5.3} & \mg{3.8} & \mdg{0.056} \\
& \gcell{\ourmethod} &
  \gcell{\textbf{66.8}} & \gcell{\textbf{\mg{2.3}}} & \gcell{\textbf{\mg{3.2}}} & \gcell{\textbf{\mg{2.3}}} & \gcell{\textbf{\mdg{0.020}}} \\
& \dcell{$\Delta$} &
  \dcell{\darkred{\textbf{+1.5}}} &
  \dcell{\darkblue{\textbf{\makebox[0pt][r]{$\downarrow$\,}54.4\%}}} &
  \dcell{\darkblue{\textbf{\makebox[0pt][r]{$\downarrow$\,}43.2\%}}} &
  \dcell{\darkblue{\textbf{\makebox[0pt][r]{$\downarrow$\,}41.5\%}}} &
  \dcell{\darkblue{\textbf{\makebox[0pt][r]{$\downarrow$\,}67.9\%}}} \\
\hline

\multirow{6}{*}{
  \rotatebox{90}{
    \fontsize{9.8pt}{11pt}\selectfont
    R1-Qwen-32B
  }
}
& \gcell{Vanilla} &
  \gcell{60.8} & \gcell{\mg{5.3}} & \gcell{\mg{5.9}} & \gcell{\mg{4.4}} & \gcell{\mdg{0.071}} \\
& DEER &
  58.9 & \mg{4.9} & \mg{5.7} & \mg{4.4} & \mdg{0.073} \\
& \gcell{RL + LP} &
  \gcell{60.4} & \gcell{\mg{4.6}} & \gcell{\mg{5.5}} & \gcell{\mg{3.9}} & \gcell{\mdg{0.064}} \\
& S-GRRO &
  61.3 & \mg{4.7} & \mg{5.6} & \mg{4.2} & \mdg{0.066} \\
& \gcell{\ourmethod} &
  \gcell{\textbf{61.8}} & \gcell{\textbf{\mg{2.4}}} & \gcell{\textbf{\mg{3.5}}} & \gcell{\textbf{\mg{2.7}}} & \gcell{\textbf{\mdg{0.030}}} \\
& \dcell{$\Delta$} &
  \dcell{\darkred{\textbf{+1.0}}} &
  \dcell{\darkblue{\textbf{\makebox[0pt][r]{$\downarrow$\,}54.3\%}}} &
  \dcell{\darkblue{\textbf{\makebox[0pt][r]{$\downarrow$\,}41.0\%}}} &
  \dcell{\darkblue{\textbf{\makebox[0pt][r]{$\downarrow$\,}37.9\%}}} &
  \dcell{\darkblue{\textbf{\makebox[0pt][r]{$\downarrow$\,}57.9\%}}} \\
\hline

\multirow{6}{*}{
  \rotatebox{90}{
    \fontsize{9.8pt}{11pt}\selectfont
    R1-Llama-70B
  }
}
& \gcell{Vanilla} &
  \gcell{64.5} & \gcell{\mg{6.5}} & \gcell{\mg{6.2}} & \gcell{\mg{4.8}} & \gcell{\mdg{0.094}} \\
& DEER &
  63.1 & \mg{5.9} & \mg{5.9} & \mg{4.9} & \mdg{0.096} \\
& \gcell{RL + LP} &
  \gcell{65.0} & \gcell{\mg{5.7}} & \gcell{\mg{5.6}} & \gcell{\mg{4.4}} & \gcell{\mdg{0.083}} \\
& S-GRRO &
  65.2 & \mg{5.9} & \mg{5.7} & \mg{4.6} & \mdg{0.086} \\
& \gcell{\ourmethod} &
  \gcell{\textbf{66.2}} & \gcell{\textbf{\mg{2.9}}} & \gcell{\textbf{\mg{3.5}}} & \gcell{\textbf{\mg{2.8}}} & \gcell{\textbf{\mdg{0.030}}} \\
& \dcell{$\Delta$} &
  \dcell{\darkred{\textbf{+1.7}}} &
  \dcell{\darkblue{\textbf{\makebox[0pt][r]{$\downarrow$\,}54.7\%}}} &
  \dcell{\darkblue{\textbf{\makebox[0pt][r]{$\downarrow$\,}43.4\%}}} &
  \dcell{\darkblue{\textbf{\makebox[0pt][r]{$\downarrow$\,}42.2\%}}} &
  \dcell{\darkblue{\textbf{\makebox[0pt][r]{$\downarrow$\,}68.0\%}}} \\
\hline

\Xhline{1.2pt}
\end{tabular}
}
\caption{Results on GPQA-Diamond with FoE-related metrics.}
\label{tab:foe_gpqa_full}
\end{table}

\subsection{Self-Consistency Experiment}
\label{sec:cons}
\begin{figure}[t]
  \label{fig:cons}
  \centering
  \begin{subfigure}{\linewidth}
    \centering
    \includegraphics[width=\linewidth]{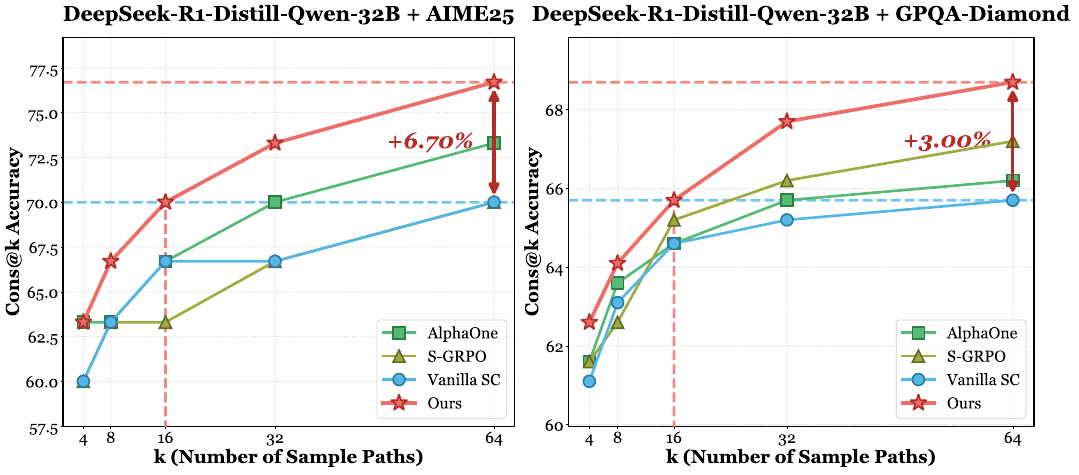}
    \vspace{-1.5em} 
    \caption{Test-time scalability on Qwen model}
    \label{fig:case-qwen}
  \end{subfigure}
  
  \vspace{0.1em} 
  
  \begin{subfigure}{\linewidth}
    \centering
    \includegraphics[width=\linewidth]{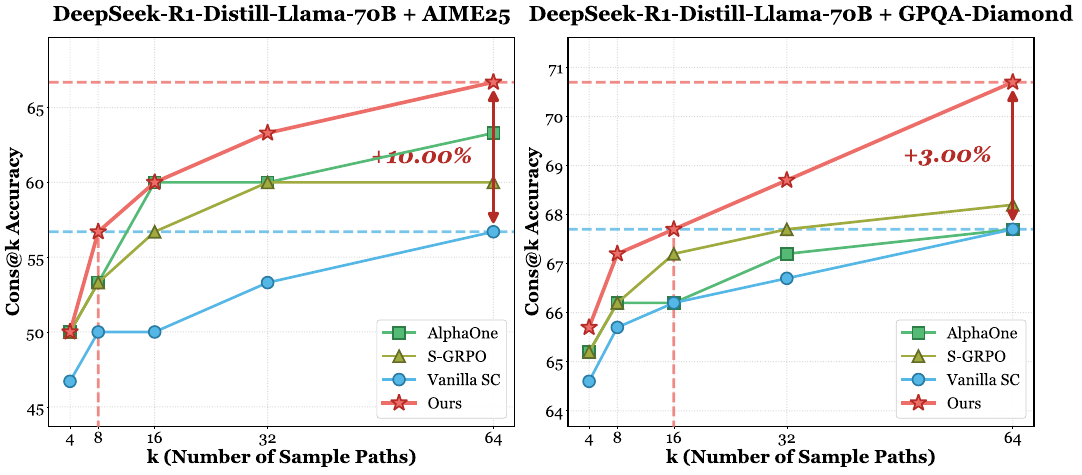}
    \vspace{-1.5em} 
    \caption{Test-time scalability on Llama model}
    \label{fig:case-llama}
  \end{subfigure}
  
  \caption{Test-time scalability under self-consistency.}
  \label{fig:scalability_comparison}
\end{figure}

Figure \ref{fig:case-qwen} and \ref{fig:case-llama} illustrate that \textbf{RED} achieves superior Cons@k performance across nearly all $k$ values compared to Vanilla SC, AlphaOne, and S-GRPO. 
On the challenging AIME25 benchmark with the DeepSeek-R1-Distill-Llama-70B backbone, \textbf{RED} demonstrates substantial improvements, outperforming the vanilla baseline by a remarkable margin of \textbf{10.0\%} at $k=64$ ($56.7\% \to 66.7\%$). 
Notably, \textbf{RED} exhibits exceptional sample efficiency. As shown in Figure~\ref{fig:case-llama}, on the Llama-70B model, our method at $k=8$ achieves performance ($56.7\%$) comparable to the vanilla method at $k=64$, indicating an $\sim 8\times$ reduction in computational cost to reach the same accuracy. 
Similarly, on the DeepSeek-R1-Distill-Qwen-32B model, \textbf{RED} consistently surpasses the baselines. On GPQA-Diamond, \textbf{RED} at $k=16$ matches the performance of the vanilla baseline at $k=64$ ($65.7\%$), effectively accelerating inference by $4\times$.
At $k=64$, \textbf{RED} achieves consistent gains ranging from $3.0\%$ to $10.0\%$ across all datasets and models, underscoring its robustness in scaling test-time computation.

\subsection{Ablation Study}
\label{sec:ably}

Table~\ref{tab:ably} validates the necessity of combining \textit{Refining First} and \textit{Discarding Subs}.
\ding{182} Removing \textit{Discarding Subs} consistently hurts Pass@1 compared to RED, even though \textit{Refining First} is still applied.
This is because subsequent solutions can contaminate the improved first trajectory: later rollouts may reuse or amplify wrong artifacts and eventually override an otherwise high-quality first solution.
In addition, keeping subs naturally increases the generation length, leading to substantially more tokens. \ding{183} Removing \textit{Refining First} while still \textit{Discarding Subs} yields slightly higher Pass@1 than the vanilla setting on most benchmarks.
This supports our hypothesis that a major failure mode comes from the interference introduced by subs; early stopping reduces such interference and stabilizes the final decision.
However, this variant remains below RED because it lacks targeted interventions on error-prone positions in the first solution, so the model may require more steps to reach a stable and correct answer, which also results in higher token usage than RED. Overall, RED achieves the best trade-off: \textit{Refining First} improves the quality of the first solution by reducing root errors, and \textit{Discarding Subs} prevents later solutions from perturbing that improved trajectory, jointly improving Pass@1 while reducing tokens.

\begin{table*}[htbp]
\centering
\renewcommand\tabcolsep{5.3pt}
\renewcommand\arraystretch{1.1}

\resizebox{\linewidth}{!}{
\begin{tabular}{c|l|cc|cc|cc|cc|cc}
\Xhline{1.2pt}
\rowcolor{MorandiHeader}
\textbf{Model} & \textbf{Method} &
\multicolumn{2}{c|}{\textbf{AIME24}} &
\multicolumn{2}{c|}{\textbf{AIME25}} &
\multicolumn{2}{c|}{\textbf{MATH500}} &
\multicolumn{2}{c|}{\textbf{GSM8K}} &
\multicolumn{2}{c}{\textbf{GPQA-Diamond}} \\
\rowcolor{MorandiHeader}
& &
\textbf{Pass@1 $\uparrow$} & \textbf{Token $\downarrow$} &
\textbf{Pass@1 $\uparrow$} & \textbf{Token $\downarrow$} &
\textbf{Pass@1 $\uparrow$} & \textbf{Token $\downarrow$} &
\textbf{Pass@1 $\uparrow$} & \textbf{Token $\downarrow$} &
\textbf{Pass@1 $\uparrow$} & \textbf{Token $\downarrow$} \\
\Xhline{1.2pt}
                                            
\multirow{4}{*}{\rotatebox{90}{\fontsize{8.6pt}{9.4pt}\selectfont DeepSeek-R1}}
& \gcell{\textbf{\textit{w/o} All (Vanilla)}} &
  \gcell{76.7} & \gcell{11638} &
  \gcell{67.8} & \gcell{12625} &
  \gcell{97.2} & \gcell{4716} &
  \gcell{95.1} & \gcell{1325} &
  \gcell{70.9} & \gcell{7544} \\
& \textbf{\textit{w/o} Refining First} &
  76.9 & 6739 &
  67.9 & 7067 &
  97.2 & 2769 &
  95.0 & 519 &
  71.0 & 4179 \\
& \gcell{\textbf{\textit{w/o} Discarding Subs}} &
  \gcell{77.9} & \gcell{9831} &
  \gcell{69.3} & \gcell{10469} &
  \gcell{97.6} & \gcell{3887} &
  \gcell{94.8} & \gcell{947} &
  \gcell{71.8} & \gcell{5947} \\
& \textbf{RED (ours)} &
  78.9 & 6148 &
  70.0 & 6404 &
  98.1 & 2408 &
  95.1 & 447 &
  72.4 & 3704 \\
\hline

\multirow{4}{*}{\rotatebox{90}{\fontsize{8.6pt}{9.4pt}\selectfont \shortstack[c]{Qwen3-8B\\Thinking}}}
& \gcell{\textbf{\textit{w/o} All (Vanilla)}} &
  \gcell{70.0} & \gcell{11125} &
  \gcell{61.1} & \gcell{12490} &
  \gcell{95.9} & \gcell{4486} &
  \gcell{94.1} & \gcell{1573} &
  \gcell{58.8} & \gcell{6638} \\
& \textbf{\textit{w/o} Refining First} &
  70.1 & 6907 &
  61.3 & 6815 &
  96.0 & 2529 &
  93.9 & 567 &
  58.8 & 3639 \\
& \gcell{\textbf{\textit{w/o} Discarding Subs}} &
  \gcell{73.8} & \gcell{9127} &
  \gcell{62.5} & \gcell{10612} &
  \gcell{96.5} & \gcell{3611} &
  \gcell{94.4} & \gcell{1019} &
  \gcell{59.6} & \gcell{5693} \\
& \textbf{RED (ours)} &
  75.6 & 5583 &
  63.3 & 6203 &
  97.0 & 2301 &
  95.0 & 465 &
  60.1 & 3309 \\
\hline

\multirow{4}{*}{\rotatebox{90}{\fontsize{8.6pt}{9.4pt}\selectfont \shortstack[c]{Qwen3-32B\\Thinking}}}
& \gcell{\textbf{\textit{w/o} All (Vanilla)}} &
  \gcell{77.8} & \gcell{10677} &
  \gcell{68.9} & \gcell{11589} &
  \gcell{96.8} & \gcell{4318} &
  \gcell{94.3} & \gcell{1435} &
  \gcell{65.3} & \gcell{5475} \\
& \textbf{\textit{w/o} Refining First} &
  78.0 & 6354 &
  69.0 & 6729 &
  97.0 & 2361 &
  94.2 & 517 &
  65.4 & 2971 \\
& \gcell{\textbf{\textit{w/o} Discarding Subs}} &
  \gcell{79.1} & \gcell{9017} &
  \gcell{70.1} & \gcell{9617} &
  \gcell{97.4} & \gcell{3317} &
  \gcell{94.5} & \gcell{981} &
  \gcell{66.3} & \gcell{3749} \\
& \textbf{RED (ours)} &
  80.0 & 5793 &
  72.2 & 5908 &
  97.9 & 1995 &
  94.6 & 443 &
  66.8 & 2477 \\
\hline

\multirow{4}{*}{\rotatebox{90}{\fontsize{8.6pt}{9.4pt}\selectfont \shortstack[c]{Dpsk-R1-Distill\\Qwen-7B}}}
& \gcell{\textbf{\textit{w/o} All (Vanilla)}} &
  \gcell{54.4} & \gcell{10438} &
  \gcell{43.3} & \gcell{11454} &
  \gcell{91.8} & \gcell{2887} &
  \gcell{92.4} & \gcell{442} &
  \gcell{49.2} & \gcell{8016} \\
& \textbf{\textit{w/o} Refining First} &
  54.5 & 4511 &
  43.2 & 6797 &
  92.0 & 1359 &
  92.5 & 289 &
  49.0 & 4527 \\
& \gcell{\textbf{\textit{w/o} Discarding Subs}} &
  \gcell{55.3} & \gcell{8679} &
  \gcell{44.8} & \gcell{8589} &
  \gcell{93.0} & \gcell{1979} &
  \gcell{93.6} & \gcell{391} &
  \gcell{50.4} & \gcell{5941} \\
& \textbf{RED (ours)} &
  57.8 & 4293 &
  47.8 & 5690 &
  94.1 & 1187 &
  94.1 & 271 &
  51.2 & 4109 \\
\hline

\multirow{4}{*}{\rotatebox{90}{\fontsize{8.6pt}{9.4pt}\selectfont \shortstack[c]{Dpsk-R1-Distill\\Qwen-32B}}}
& \gcell{\textbf{\textit{w/o} All (Vanilla)}} &
  \gcell{70.0} & \gcell{7873} &
  \gcell{58.9} & \gcell{8906} &
  \gcell{93.3} & \gcell{2337} &
  \gcell{93.9} & \gcell{438} &
  \gcell{60.8} & \gcell{6027} \\
& \textbf{\textit{w/o} Refining First} &
  70.2 & 4327 &
  59.0 & 5741 &
  93.2 & 1549 &
  94.0 & 257 &
  60.8 & 4091 \\
& \gcell{\textbf{\textit{w/o} Discarding Subs}} &
  \gcell{73.7} & \gcell{6317} &
  \gcell{60.8} & \gcell{7511} &
  \gcell{94.2} & \gcell{1962} &
  \gcell{94.2} & \gcell{339} &
  \gcell{61.1} & \gcell{5311} \\
& \textbf{RED (ours)} &
  74.4 & 3898 &
  63.3 & 5018 &
  95.2 & 1347 &
  94.6 & 209 &
  61.8 & 3497 \\
\hline

\multirow{4}{*}{\rotatebox{90}{\fontsize{8.6pt}{9.4pt}\selectfont \shortstack[c]{Dpsk-R1-Distill\\LLaMA-8B}}}
& \gcell{\textbf{\textit{w/o} All (Vanilla)}} &
  \gcell{45.6} & \gcell{10799} &
  \gcell{28.9} & \gcell{11548} &
  \gcell{86.2} & \gcell{3635} &
  \gcell{92.3} & \gcell{606} &
  \gcell{46.3} & \gcell{8341} \\
& \textbf{\textit{w/o} Refining First} &
  45.5 & 4879 &
  29.1 & 5487 &
  86.3 & 2267 &
  92.5 & 336 &
  46.1 & 4727 \\
& \gcell{\textbf{\textit{w/o} Discarding Subs}} &
  \gcell{46.3} & \gcell{8841} &
  \gcell{30.1} & \gcell{9587} &
  \gcell{87.6} & \gcell{3076} &
  \gcell{92.7} & \gcell{493} &
  \gcell{47.2} & \gcell{6661} \\
& \textbf{RED (ours)} &
  47.8 & 4507 &
  34.4 & 5039 &
  90.1 & 2009 &
  93.1 & 283 &
  47.8 & 3593 \\
\hline

\multirow{4}{*}{\rotatebox{90}{\fontsize{8.6pt}{9.4pt}\selectfont \shortstack[c]{Dpsk-R1-Distill\\LLaMA-70B}}}
& \gcell{\textbf{\textit{w/o} All (Vanilla)}} &
  \gcell{68.9} & \gcell{7766} &
  \gcell{47.8} & \gcell{8909} &
  \gcell{94.1} & \gcell{2433} &
  \gcell{94.0} & \gcell{432} &
  \gcell{64.5} & \gcell{5881} \\
& \textbf{\textit{w/o} Refining First} &
  69.0 & 4491 &
  48.0 & 5397 &
  94.0 & 1219 &
  94.1 & 303 &
  64.7 & 4077 \\
& \gcell{\textbf{\textit{w/o} Discarding Subs}} &
  \gcell{71.4} & \gcell{5583} &
  \gcell{49.2} & \gcell{6607} &
  \gcell{95.6} & \gcell{1969} &
  \gcell{94.6} & \gcell{381} &
  \gcell{65.0} & \gcell{4663} \\
& \textbf{RED (ours)} &
  73.3 & 3974 &
  50.0 & 4303 &
  96.2 & 932 &
  94.9 & 269 &
  66.2 & 3563 \\
\hline

\Xhline{1.2pt}
\end{tabular}
}
\caption{Ablation study.}
\label{tab:ably}
\end{table*}

\section{Hyperparameter Settings}
\label{sec:hyperparam}
\vspace{-0.3em}

\subsection{Prompt Templates for Dual-Consistency Probing.}
In Section \ref{sec:method-3}, we periodically probe the current hidden state by appending short prompts that request only the final answer.
Since the induced answer is typically fragile in early reasoning, even minor surface-form changes may elicit different answers when the model is not yet confident.
We therefore use $M{=}4$ stylistically diverse yet semantically equivalent probe templates and require cross-prompt agreement before triggering an early exit.
To reduce prompt length to enable fully batched probing, we length-normalize all templates to have the same token count.

\begin{itemize}[leftmargin=*, itemsep=0.25em, topsep=0.2em]
    \item \textbf{Template A (imperative / QA-style).}
    \textit{\detokenize{Stop now immediately and output the final answer only. Use one token. Format: Final Answer: <answer>. No explanation, no extras.}}
    \item \textbf{Template B (role / evaluator-style).}
    \textit{\detokenize{Switch to answer-mode. Output exactly one token inside \boxed{ }. Do not justify. Do not add words, symbols, or spaces.}}
    \item \textbf{Template C (schema / machine-readable).}
    \textit{\detokenize{Structured reply: answer=<token>. Output only the value token after '='. No braces, quotes, labels, or commentary. One token strictly now.}}
    \item \textbf{Template D (conversational / best-guess).}
    \textit{\detokenize{Quick check: what's your final answer? Reply with one token only. No reasoning, no repeats, no punctuation. Just answer now.}}
\end{itemize}

\noindent
\textbf{Rationale for Prompt Selection.}
The four templates are intentionally short but differ sharply in surface form and pragmatics while preserving the same semantic request (“return only the final answer token”).
They span distinct instruction styles: (A) a strict imperative with an explicit \texttt{Final Answer:} anchor; (B) a mode-switch framing with a Latex Style wrapper cue (\texttt{\textbackslash boxed}) to alter formatting habits; (C) a schema like key--value constraint that changes punctuation and discourse structure; and (D) an informal question that tests robustness under a different conversational intent (“quick check / best guess”).
These differences induce diverse decoding priors (register, anchors, and syntax), so requiring cross-prompt agreement makes the trigger a strong indicator that the latent answer has stabilized, rather than being a prompt-sensitive artifact of an early, brittle internal state.

\vspace{-0.2em}
\subsection{Hyperparameters in Refining \first.}
We set the sliding-window length to $L=15$ and define the trigger condition as the entropy falling within the local Top-$K$ ($K=3$)
For the entropy-variance threshold $T$, we perform a sensitivity study on Qwen3-32B-thinking over MATH500, reporting Pass@1 and average generated Token count. We fix all other hyperparameters, including the full Discarding \subs\ configuration, and vary only $T$.
As shown in Table~\ref{tab:sensitivity_T}, values around $2.0\!\sim\!2.5$ behave similarly, while $T{=}2.4$ yields the best joint outcome (highest Pass@1 and lowest tokens), which we adopt as default.

\begin{table}[t]
\centering
\renewcommand\tabcolsep{12pt} 
\renewcommand\arraystretch{1.2}

\resizebox{1.0\linewidth}{!}{
\begin{tabular}{c|l|cc}
\Xhline{1.2pt}
\rowcolor{MorandiHeader}
\textbf{Model} & \textbf{Setting} & \multicolumn{2}{c}{\textbf{MATH500}} \\
\rowcolor{MorandiHeader}
& & \textbf{Pass@1$\uparrow$} & \textbf{Token$\downarrow$} \\
\Xhline{1.2pt}

\multirow{6}{*}{\makecell{Qwen3-32B-\\thinking}}
& \gcell{$T{=}2.0$} & \gcell{97.6} & \gcell{2058} \\ 
& $T{=}2.1$           & 97.7           & 2042           \\ 
& \gcell{$T{=}2.2$} & \gcell{97.6} & \gcell{2027} \\ 
& $T{=}2.3$           & 97.8           & 2011           \\ 
& \gcell{\textbf{$T{=}2.4$}} & \gcell{\textbf{97.9}} & \gcell{\textbf{1995}} \\ 
& $T{=}2.5$           & 97.5           & 2008           \\ 
\Xhline{1.2pt}
\end{tabular}
}
\caption{Sensitivity of the entropy-variance threshold $T$ for Refine \first\ on MATH500 (Qwen3-32B-thinking). Best results are highlighted in \textbf{bold}.}
\label{tab:sensitivity_T}
\end{table}

\subsection{Hyperparameters in Discarding \subs.}
We probe every $K{=}2$ steps, use $M{=}4$ prompt templates, and draw $N{=}12$ parallel samples per template (fixed throughout).
We ablate only the intra-prompt-template internal consistency threshold $P$ while keeping all other settings fixed (including Refine \first).
As shown in Table~\ref{tab:sensitivity_P}, $P{=}0.6$ yields the highest Pass@1 (97.9) and is notably more token-efficient than the stricter $P{=}0.7$; lower thresholds ($P{=}0.4,0.5$) save some tokens but incur a large accuracy drop.
We therefore use $P{=}0.6$ by default.

\begin{table}[!t]
\centering
\renewcommand\tabcolsep{15pt} 
\renewcommand\arraystretch{1.2}

\resizebox{1.0\linewidth}{!}{
\begin{tabular}{c|l|cc}
\Xhline{1.2pt}
\rowcolor{MorandiHeader}
\textbf{Model} & \textbf{Setting} &
\multicolumn{2}{c}{\textbf{MATH500}} \\
\rowcolor{MorandiHeader}
& &
\textbf{Pass@1$\uparrow$} & \textbf{Token$\downarrow$} \\
\Xhline{1.2pt}

\multirow{4}{*}{\makecell{Qwen3-32B-\\thinking}}
& \gcell{$P{=}0.4$} & \gcell{87.1} & \gcell{\textbf{1741}} \\ 
& $P{=}0.5$           & 87.4           & 1877           \\ 
& \gcell{\textbf{$P{=}0.6$}} & \gcell{\textbf{97.9}} & \gcell{1995} \\ 
& $P{=}0.7$           & 97.4           & 2434           \\ 
\Xhline{1.2pt}
\end{tabular}
}
\caption{Sensitivity of the internal consistency rate $P$ for Discarding \subs\ on MATH500 (Qwen3-32B-thinking). Best results are highlighted in \textbf{bold}.}
\label{tab:sensitivity_P}
\end{table}

\section{Baselines: Further Details}
\label{sec:baseline}
\subsection{Introduction of Baselines}
To comprehensively evaluate the effectiveness of \ourmethod, we compare it against a diverse set of baselines. These methods are categorized into three distinct groups: the vanilla backbone model, training-free intervention methods, and reinforcement learning (RL) based strategies.

\paragraph{I) Vanilla Model.} We utilize the original backbone Large Reasoning Model (LRM) as the fundamental baseline. It generates Chain-of-Thought (CoT) reasoning and answers using standard decoding without any additional training or dynamic intervention mechanisms.

\paragraph{II) Training-free Methods.} These approaches introduce inference-time heuristics to improve efficiency without updating model parameters:
\begin{itemize}
    \item \textbf{DEER}~\citep{yang2025dynamicearlyexitreasoning} is a dynamic early-exit mechanism that terminates generation based on the geometric mean of token probabilities within a tentative answer.
    \item \textbf{Think or Not}~\citep{yong2025think} employs an entropy-based adaptive stopping strategy, halting reasoning when the generation entropy exceeds a specific threshold.
    \item \textbf{AlphaOne}~\citep{zhang-etal-2025-alphaone} introduces a dynamic switching mechanism using a Bernoulli process to transition between "slow thinking" (reasoning) and "fast thinking" (answering) modes at test time.
\end{itemize}

\paragraph{III) RL-based Strategies.} These methods employ reinforcement learning to explicitly optimize the trade-off between reasoning capability and efficiency:
\begin{itemize}
    \item \textbf{DAST}~\citep{shen-etal-2025-dast} (Difficulty-Adaptive Slow-Thinking) utilizes a difficulty-aware token budget and reward optimization to adaptively adjust reasoning length based on problem complexity.
    \item \textbf{RL + Length Penalty}~\citep{arora2025training} modifies the reward function to penalize correct answers that deviate from the average length, encouraging concise reasoning.
    \item \textbf{GRPO}~\citep{deepseekai2025deepseekr1incentivizingreasoningcapability} (DeepSeek-R1) applies Group Relative Policy Optimization to incentivize reasoning capabilities by comparing a group of candidate outputs and updating the policy based on normalized advantages.
    \item \textbf{S-GRPO}~\citep{dai2025sgrpo} extends the GRPO framework with a Serial Group Decay Reward strategy, assigning decaying rewards to different reasoning steps to train the model in identifying optimal early-exit points.
\end{itemize}

\subsection{Baseline Settings}
For all baselines, we strictly adhere to the settings detailed in their respective original papers. To ensure fair comparison, we keep the same inference hyperparameters across methods (our vanilla decoding): sampling with temperature $T{=}0.6$ and top-$p{=}0.95$, max-token${=}16000$, and we report Pass@1 averaged over three repeated runs. To ensure optimal performance, we utilize the official chat templates corresponding to each model family (e.g., applying the specific formats for Qwen and Llama, respectively).
For RL-based baselines, training is conducted on a cluster of NVIDIA A100 GPUs with bf16 mixed precision, DeepSpeed ZeRO-3 sharding, gradient checkpointing, and identical reward extraction / answer parsing logic across methods.

\section{Experimental Results of Rollback in Subs.}
\label{sec:rollback}

\begin{figure}[htbp]
\centering
\includegraphics[width=1.0\linewidth]{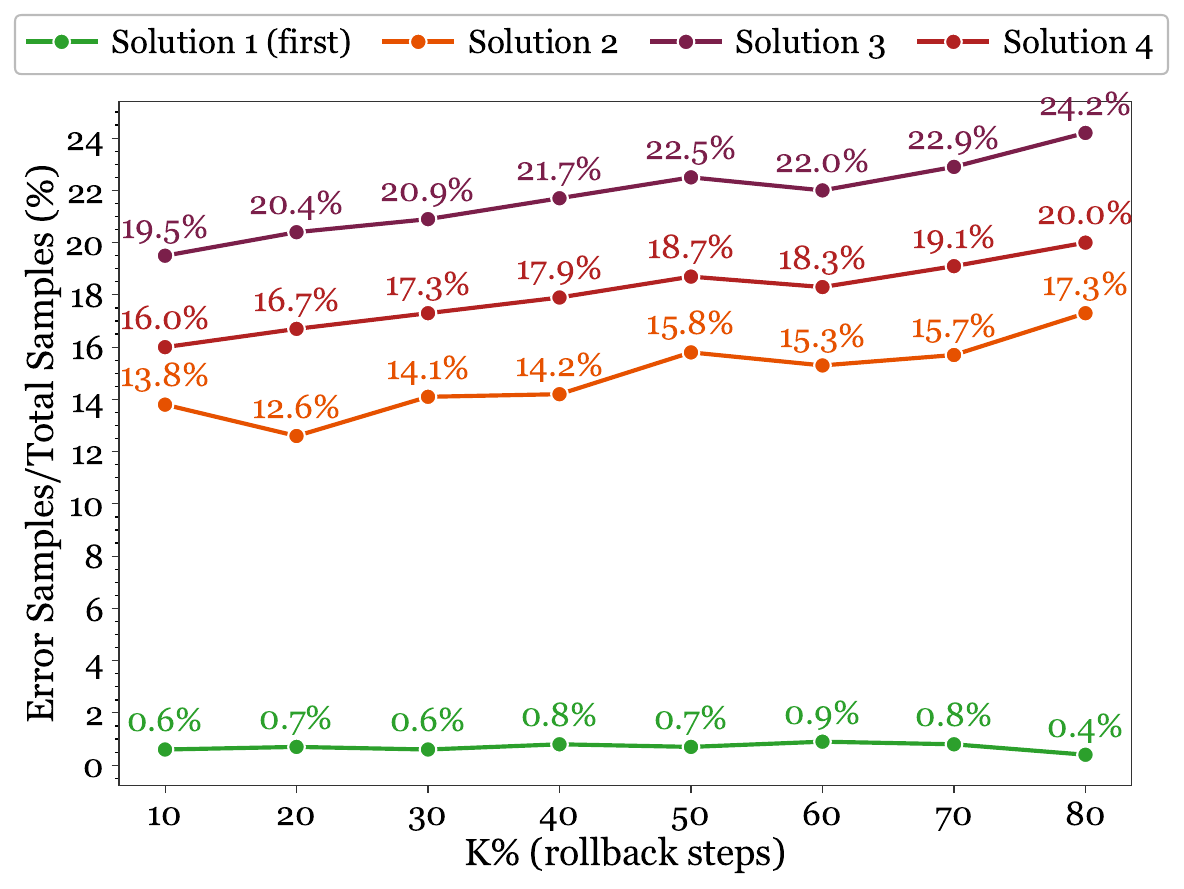} 
\caption{
Average sampling error rate (\%, lower is better) under rollback sampling.
For each solution variant, we truncate an originally correct trajectory at rollback ratio $K\in\{10,20,\ldots,80\}\%$, restore the KV cache at the truncation point, and re-sample the remaining continuation $N=100$ times; Curves report error rates averaged over problems.
\textsc{First} remains highly robust (error $<1\%$ across all $K$), whereas \textsc{Subs} exhibit substantially higher and largely $K$ invariant error rates, revealing persistent latent instability even when the original final answers are correct.
} 
\label{fig:rollback}
\end{figure}

\paragraph{Observations and Motivations.}
Through case study we find that subsequent solutions (\subs) can still contain noticeably unreliable or even incorrect intermediate steps, even when their \emph{final} answers appear correct. Therefore, a one-shot correct outcome does not necessarily imply a stable reasoning trajectory. This raises a latent risk: if the generation process is perturbed (e.g., by interrupting and resampling the continuation), these ``apparently correct'' \subs may drift to an incorrect final answer more easily than the first solution.

\paragraph{Rollback-sampling Experiment.}
To quantify the robustness of each solution's reasoning process, we design a \emph{rollback sampling} experiment on BS-17k-subset using Qwen3-8. The core idea is to truncate a solution that is correct in the original run, and then repeatedly re-sample its continuation from the same intermediate state, measuring how often the final answer becomes incorrect. We evaluate the first four solutions for each problem, denoted as $\text{Solution }1$--$\text{Solution }4$, where $\text{Solution }1$ corresponds to \first and $\text{Solution }2$--$\text{Solution }4$ are \subs. For metrics involving $\text{Solution }k$, the average is computed over the subset of problems for which at least $k$ solutions were successfully sampled. The procedure is as follows.

\paragraph{Instance Selection.} We collect problems for which the model can produce a correct final answer in a single pass, and for each solution variant under evaluation, we only keep instances where that solution is correct in the original run. This isolates stability under correctness.

\paragraph{Rollback Point.}
  For a solution $s$ with reasoning length $T_s$, we choose a rollback ratio
  $K\in\{10,20,\ldots,80\}$ and truncate the trajectory at
  \begin{equation}
  t = \left\lfloor \frac{K}{100}\,T_s \right\rfloor .
  \end{equation}
  We then roll back the generation state to this truncation point and keep the prefix up to step $t$.
  In practice, we store the KV cache during decoding and restore it at the rollback point, so that we can resume sampling from the identical intermediate state without recomputing the prefix.

 \paragraph{Re-sampling and Error-rate Estimation.} Starting from the restored intermediate state, we independently sample the continuation $N=100$ times until completion, producing $100$ final answers. We define the sampling error rate of solution $s$ at rollback ratio $K$ as
  \begin{equation}
  \widehat{e}_s(K) = \frac{1}{N}\sum_{n=1}^{N}\mathbf{1}\!\left[\hat{y}^{(n)}_{s,K}\neq y^\star\right],
  \end{equation}
  where $y^\star$ is the ground-truth answer and $\hat{y}^{(n)}_{s,K}$ is the final answer from the $n$-th re-sampled continuation. Averaging over all retained problems yields $\bar e_s(K)$.

\paragraph{Experimental Results.}
Figure~\ref{fig:rollback} plots $\bar e_s(K)$ against $K$.
\first is extremely robust: across all rollback ratios, the error rate is consistently below $1\%$ (mean $\approx 0.69\%$, range $0.4\%\sim 0.9\%$). This indicates that the intermediate states along \first reliably constrain the continuation toward the correct answer, and re-sampling rarely deviates from the correct trajectory. In contrast, \subs exhibit pronounced fragility. Solution~2 has a mean error rate of about $14.85\%$ (range $12.6\%\sim 17.3\%$), Solution~3 reaches about $21.76\%$ (range $19.5\%\sim 24.2\%$), and Solution~4 stays around $18.0\%$ (range $16.0\%\sim 20.0\%$). Importantly, these error rates remain consistently high across rollback points: sweeping $K$ from $10\%$ to $80\%$ changes the error probabilities only mildly. This suggests that the risk is not concentrated in a small set of ``critical'' steps near the end; instead, later solutions follow comparatively risky reasoning paths throughout the trajectory. If we aggregate Solutions~2--4 as \subs, the mean error rate is $\approx 18.20\%$.
Compared with \first (mean $\approx 0.69\%$), the relative error probability under interruption-and-resampling is
\begin{equation}
\frac{\bar e_{\first}}{\bar e_{\subs}}
\approx \frac{0.69}{18.20}
\approx 3.8\%,
\end{equation}
which is consistent with the main-text claim that \first incurs only a small fraction of the rollback-induced error probability of \subs.

\paragraph{Conclusion Aligned with Observation}
Rollback sampling directly validates the ``latent risk'' in \subs: even when \subs happens to be correct in a single run, its intermediate states provide weaker ``lock-in'' toward the correct answer, so re-sampling from an interruption point can readily slip to an incorrect final answer.

\section{FoE Initialization: Further Details}

\subsection{Human-LLM Agreement of Few-Shot PCS Judging}
\label{app:human_llm_agreement}

\paragraph{Purpose.}
In Section \ref{sec:foe-1}, we initialize the Forest of Errors (FoE) by defining a parent-child association score $\mathrm{Score}(e_i,e_j)$ on a 1-5 scale, and using an advanced LLM with few-shot prompting to score candidate parents for each child error in a near-to-far order, accepting the first candidate whose score exceeds a threshold $\tau$.
Since this few-shot judging step directly determines FoE edges, we conduct a human-LLM agreement study to validate that the LLM judge aligns with expert causal judgments.

\paragraph{Task and inputs.}
Formally, we define the input space for our experiments such that each evaluation instance is structured as a triple $(\text{CONTEXT\_UP\_TO\_CHILD}, e_i, e_j)$:

(i) the full reasoning prefix up to and including the child error node $e_j$,
(ii) an earlier candidate parent error node $e_i$,
and (iii) the target child error node $e_j$.
Both humans and the LLM judge output a PCS (Parent-Children Score) in $[1.0, 5.0]$ with one decimal place, following the same rubric of \emph{direct artifact reuse and directness}.

\paragraph{Data.}
We sample instances from the same benchmark suite used throughout the paper (AIME25, MATH500, GSM8K, and GPQA-Diamond).
To reflect the near-to-far evaluation regime in Section \ref{sec:foe-1}, we stratify sampling by temporal distance between $e_i$ and $e_j$:
(1) \textbf{Near} pairs, where $e_i$ is among the 1-3 nearest preceding error nodes of $e_j$; and
(2) \textbf{Far} pairs, where $e_i$ occurs at least 6 error nodes before $e_j$.
Our final evaluation set contains $N=500$ pairs, balanced across math/science domains and near/far strata.

\paragraph{Human annotation protocol.}
We recruit four expert annotators, covering both mathematical and scientific reasoning backgrounds.
Annotators independently assign PCS scores using a shared guideline sheet (the same definitions/anchors used by the few-shot prompt).
We use the median of four scores as the human reference.
To quantify human consistency, we report Krippendorff's $\alpha$ (ordinal) across the four annotators.
We also compute a leave-one-out (LOO) human agreement at the decision threshold $\tau=4.0$ (each annotator vs.\ the majority of the other three), to contextualize the LLM agreement relative to typical human variance.

\paragraph{LLM judging.}
We run the PCS few-shot prompt (Appendix~A) with a frontier LLM under deterministic decoding (temperature 0).
The judge sees exactly the same $(\texttt{CONTEXT\_UP\_TO\_CHILD}, e_i, e_j)$ tuple as the human experts, and outputs only the numeric PCS score.

\paragraph{Metrics.}
We report:
(i) \textbf{score-level correspondence} via Spearman correlation $\rho$ and mean absolute error (MAE) against the median human score;
(ii) \textbf{ordinal agreement} via quadratic weighted kappa (QWK) after mapping PCS to the nearest integer in $\{1,2,3,4,5\}$; and
(iii) \textbf{decision agreement} at $\tau=4.0$, treating PCS$\ge 4.0$ as a positive edge (accuracy and F1).

\begin{figure}[t]
\centering
\includegraphics[width=1\linewidth]{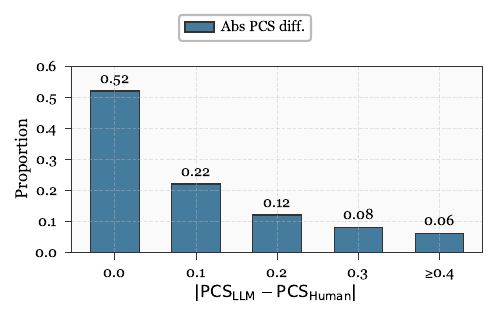} 
\caption{Distribution of absolute PCS disagreement between the LLM judge and the median human score.}
\label{fig:pcs_abs_err}
\end{figure}

\paragraph{Results.}
Table~\ref{tab:pcs_human_llm} summarizes the agreement performance. 
\textbf{Separately, to validate the consistency of the human reference,} we measured that human experts achieved a Krippendorff's $\alpha$ of 0.78, establishing a reliable consensus baseline for the PCS rubric. 
The LLM judge aligns strongly with this baseline, yielding a Spearman's $\rho=0.88$ and a low MAE of $0.28$. 
Crucially for FoE construction, at the $\tau=4.0$ threshold, the LLM reaches **94\% accuracy and 0.92 F1**, matching (and slightly exceeding) the **93\% LOO human agreement**. 
As shown in Figure~\ref{fig:pcs_abs_err}, the error distribution is heavily skewed toward zero, confirming that few-shot prompting effectively emulates expert reasoning in identifying artifact reuse.

\paragraph{Single-case sanity check (near-to-far parent selection).}
Table~\ref{tab:pcs_case} validates this near-to-far selection: both humans and the LLM assign PCS$\ge\tau$ solely to the closest candidate containing the reused artifact, yielding identical parent assignment.

\begin{table}[t]
\centering
\renewcommand\tabcolsep{5.0pt}
\renewcommand\arraystretch{1.08}

\resizebox{\columnwidth}{!}{
\begin{tabular}{l|c|cccc|cc}
\Xhline{1.2pt}
\rowcolor{MorandiHeader}
\textbf{Dataset} & \textbf{\#Pairs} & $\boldsymbol{\rho}$ & \textbf{QWK} & \textbf{MAE} & \textbf{Acc@$\tau$} & \textbf{F1@$\tau$} & \textbf{} \\
\Xhline{1.2pt}

\gcell{AIME25}       & \gcell{150} & \gcell{0.90} & \gcell{0.84} & \gcell{0.24} & \gcell{0.94} & \gcell{0.93} & \gcell{} \\
MATH500              & 150         & 0.87         & 0.82         & 0.27         & 0.93         & 0.92         &  \\
\gcell{GSM8K}        & \gcell{100} & \gcell{0.85} & \gcell{0.79} & \gcell{0.30} & \gcell{0.92} & \gcell{0.91} & \gcell{} \\
GPQA-Diamond         & 100         & 0.84         & 0.80         & 0.32         & 0.92         & 0.90         &  \\
\Xhline{0.9pt}

\gcell{Overall}      & \gcell{500} & \gcell{0.88} & \gcell{0.82} & \gcell{0.28} & \gcell{0.94} & \gcell{0.92} & \gcell{} \\

\Xhline{1.2pt}
\end{tabular}
}
\caption{Human-LLM agreement for PCS few-shot judging. Spearman's $\rho$ and MAE are computed on raw one-decimal PCS scores. QWK is computed after rounding PCS to the nearest integer in $\{1,2,3,4,5\}$. Acc@$\tau$ and F1@$\tau$ treat PCS$\ge 4.0$ as a positive edge.}
\label{tab:pcs_human_llm}
\end{table}

\paragraph{Conclusion.}
The human-LLM agreement results suggest that the proposed PCS few-shot judging protocol produces evaluations that are largely consistent with expert human judgments.
Across both mathematical and scientific benchmarks, the LLM judge exhibits agreement levels comparable to human inter-annotator consistency, at both the continuous score level and the thresholded decision level used for FoE edge construction ($\tau{=}4.0$).
These findings indicate that the few-shot prompt captures key signals of direct error induction via artifact reuse, rather than relying solely on surface proximity or topical overlap.
While the LLM judge is not intended to replace human expertise in all settings, the observed agreement provides empirical support for its use as a practical and scalable component in FoE initialization within our framework.

\begin{table}[t]
\centering
\renewcommand\tabcolsep{4.8pt}
\renewcommand\arraystretch{1.05}

\resizebox{\columnwidth}{!}{
\begin{tabular}{c|l|p{3.4cm}|c|c}
\Xhline{1.2pt}
\rowcolor{MorandiHeader}
\textbf{Order} & \textbf{Candidate} & \textbf{Short description} & \textbf{Human PCS} & \textbf{LLM PCS} \\
\Xhline{1.2pt}

\gcell{Nearest} & \gcell{$e_{j-1}$} & \gcell{Wrong binding reused by child ($T{=}13$)} & \gcell{4.9} & \gcell{5.0} \\
2nd            & $e_{j-2}$         & Ancestor value error (total $=13$)              & 3.6         & 3.7         \\
\gcell{Far}     & \gcell{$e_{j-6}$} & \gcell{Same segment, no artifact reuse}          & \gcell{2.1} & \gcell{2.0} \\

\Xhline{1.2pt}
\end{tabular}
}
\caption{Single-case example of near-to-far candidate scoring for one child error node. The LLM and humans agree on the first candidate reaching $\tau{=}4.0$, which becomes the parent.}
\label{tab:pcs_case}
\end{table}

\subsection{Forest Modeling: Complete Details}
\label{sec:modeling}

\paragraph{Modeling details.}

Given a set of identified error nodes within a reasoning trace (e.g., via the o1-based annotation method~\cite{yang-etal-2025-beyond-first}), we denote the chronologically ordered sequence as $E=[e_1, e_2, \ldots, e_n]$. To reconstruct the causal structure, we evaluate the likelihood that an earlier error $e_i$ ($i < j$) directly induces a subsequent error $e_j$. We define a parent--child association score $s_{ij}$, quantified via the few-shot prompting protocol detailed in Appendix \ref{sec:few-shot}:
\begin{equation}
s_{ij} \triangleq \textsc{Score}(e_i, e_j).
\label{eq:score-def}
\end{equation}
Original scores are assessed on a scale of $[1, 5]$. For computational consistency, we normalize these values to $[0.2, 1]$. We designate a predecessor as a valid candidate parent only if the score meets a strict threshold $\tau$. As specified in our design, we set $\tau=0.8$ (corresponding to a raw score of 4.0). Accordingly, the set of valid candidate parents $\mathcal{P}_j$ for node $e_j$ is defined as:
\begin{equation}
\mathcal{P}_j \triangleq \{\, i \in [1, j\!-\!1] : s_{ij} \ge \tau \,\}.
\label{eq:parent-set}
\end{equation}
To ensure a forest structure where each node has at most one parent, we select the temporally closest candidate (i.e., the largest index) from $\mathcal{P}_j$. The parent index $\pi(j)$ is formally determined by:
\begin{equation}
\pi(j) \triangleq
\begin{cases}
\max \mathcal{P}_j, & \mathcal{P}_j \neq \emptyset,\\
0, & \mathcal{P}_j = \emptyset.
\end{cases}
\label{eq:parent-map}
\end{equation}
Here, $\pi(j)=0$ indicates that $e_j$ has no antecedent satisfying the threshold condition. In such cases, $e_j$ is treated as a root node, instantiating a new \emph{Tree of Errors} (ToE) within the \emph{Forest of Errors} (\foe). Otherwise, a directed edge $e_{\pi(j)} \rightarrow e_j$ is added to the ToE containing $e_{\pi(j)}$. By iterating $j$ from $1$ to $n$, this procedure constructs the complete forest, as summarized in Algorithm~\ref{alg:foe-toe-construction}.

\begin{algorithm*}[t]
\caption{Constructing the \foe\ by linking each error to its nearest inducing predecessor.}
\label{alg:foe-toe-construction}
\begin{algorithmic}[1]
\Require Chronological error list $E=[e_1,\ldots,e_n]$; threshold $\tau$; PCS scorer $\textsc{Score}(\cdot,\cdot)\in[1,5]$.
\Ensure Forest $\mathcal{F}$ (a set of ToEs) and parent map $\pi:\{1,\ldots,n\}\rightarrow\{0,\ldots,n\}$.
\State $\mathcal{F}\gets\emptyset$; $m\gets 0$ \Comment{$m$ is the number of ToEs in $\mathcal{F}$}
\State \textsc{TreeID}$[1..n]$ \Comment{maps each node to its ToE index}
\For{$j\gets 1$ \textbf{to} $n$}
  \State $p\gets 0$ \Comment{$0$ denotes ``no parent'' (root)}
  \For{$i\gets j-1$ \textbf{down to} $1$} \Comment{nearest-to-farthest}
    \State $s\gets \textsc{Score}(e_i,e_j)$
    \If{$s\ge\tau$}
      \State $p\gets i$; \textbf{break}
    \EndIf
  \EndFor
  \If{$p=0$}
     \State $m\gets m+1$
     \State Initialize a new ToE $\mathcal{T}_m$ with root $e_j$
     \State $\mathcal{F}\gets \mathcal{F}\cup\{\mathcal{T}_m\}$; \textsc{TreeID}$[j]\gets m$
  \Else
     \State Add edge $e_p \rightarrow e_j$ to ToE $\mathcal{T}_{\textsc{TreeID}[p]}$
     \State \textsc{TreeID}$[j]\gets$ \textsc{TreeID}$[p]$
  \EndIf
  \State $\pi(j)\gets p$
\EndFor
\State \Return $\mathcal{F},\pi$
\end{algorithmic}
\end{algorithm*}

\subsection{Empirical Findings on Error Node Fixing}
\label{sec:findings-fix}
This subsection provides controlled intervention evidence to support the observations in Section \ref{sec:foe-1}:
\emph{merely rectifying child nodes leaves the (uncleared) ancestor error active, which continues spawning new offspring, whereas correcting the root node substantially decelerates subsequent error-node generation}.

\paragraph{Common setup.}
We operate on the constructed Forest of Errors (FoE) in a given \textit{Sub} solution and use the same parent--child scoring function as in \S\ref{sec:foe-1}.
For convenience, we report a normalized score $\textsc{nPCS} \triangleq \textsc{PCS}/5 \in [0.2,1]$ so that the edge threshold in the main paper ($\textsc{PCS}\ge 4.0$) corresponds to $\textsc{nPCS}\ge \tau$ with $\tau=0.8$.

\paragraph{Online correction of descendants induces sibling regeneration.}
We test whether correcting a non-root node can stop error propagation when its ancestor remains uncorrected.
Given a formed local ToE $R \rightarrow C_1 \rightarrow G_1$ (root/child/leaf), once the leaf error $G_1$ appears, we \emph{immediately correct} $G_1$ (e.g., replace the erroneous intermediate artifact with the correct one or insert a corrective instruction), while keeping $C_1$ and $R$ intact.
We then let the model continue for a fixed post-fix window and detect newly generated errors.
If any newly generated error $e$ satisfies $\textsc{nPCS}(C_1, e)\ge \tau$, we regard it as a regenerated sibling under the same parent (i.e., $C_1$ keeps spawning offspring despite the leaf fix).
We further repeat the same procedure at the next stage: correct $C_1$ (while keeping $R$ intact) and test whether $R$ spawns a new child $C_2$ such that $\textsc{nPCS}(R, C_2)\ge \tau$.

To quantify the prevalence of this phenomenon beyond individual cases, we define the \textbf{Error Spawning Rate} as:
\begin{equation}
    \textsc{Spawn}@\Delta \triangleq \mathbb{E}_{u \sim \mathcal{I}} \left[ \mathbf{1}_{\mathcal{A}(u)} \right],
\end{equation}
where $\mathcal{A}(u)$ denotes the event that there exists at least one new error node $e \in E^{\text{post}}_{\Delta}(u)$ such that $\textsc{nPCS}(\mathrm{Anc}(u), e) \ge \tau$. 
Here, $\mathcal{I}$ is the set of intervention points (each involving the correction of a non-root node $u$ while leaving its nearest uncorrected ancestor $\mathrm{Anc}(u)$ intact), and $E^{\text{post}}_{\Delta}(u)$ represents the set of newly generated error nodes within the subsequent $\Delta$ decoding steps following the correction. Intuitively, $\textsc{Spawn}@\Delta$ measures the probability that an \emph{uncleared} ancestor remains active and generates at least one new erroneous child after only its descendant has been corrected.

Our empirical results on the BS-17k-subset yield a $\textsc{Spawn}@15$ value of 0.842, indicating that in over 84\% of cases, rectifying a descendant without addressing its root cause fails to halt the error cascade. This high spawning rate confirms that error nodes are structurally coupled with their ancestors rather than isolated events. The uncorrected node persisting in the context acts as a persistent "error factory," continuously driving the model toward new erroneous states.

\paragraph{Correcting the root of a formed tree suppresses subsequent growth.}
We next test whether root correction can decelerate node generation even when the tree has already formed.
During normal decoding, when we identify a small formed error tree, we duplicate the KV cache at that moment and branch into two continuations:
(i) \textbf{Fix-Root} branch: correct the root error immediately at the branching point;
(ii) \textbf{No-Fix} branch: keep the root unchanged.
We then let both branches continue decoding forward until a stop token is reached and remodeling the resulting Tree in each branch.

\begin{table}[t]
\centering
\renewcommand\tabcolsep{35pt} 
\renewcommand\arraystretch{1.2}

\resizebox{1.0\linewidth}{!}{
\begin{tabular}{l|c|c}
\Xhline{1.2pt}
\rowcolor{MorandiHeader}
\textbf{Metric} & \textbf{Fix-Root} & \textbf{No-Fix} \\
\Xhline{1.2pt}

\gcell{$|V|$} & \gcell{\textbf{0.42$\times$}} & \gcell{1.00$\times$} \\
$\bar{R}$      & 0.35$\times$                  & 1.00$\times$         \\
\gcell{$D$}   & \gcell{0.51$\times$}          & \gcell{1.00$\times$} \\

\Xhline{1.2pt}
\end{tabular}
}
\caption{
Quantitative impact of root-node intervention on the \foe structure. All metrics are reported as relative ratios to the \textbf{No-Fix} baseline ($1.00\times$), averaged over the BS-17k-subset. \textbf{Metric definitions:} $|V|$ denotes the total number of error nodes per tree; $\bar{R}$ is the error reproduction rate; and $D$ represents the average tree depth.
}
\label{tab:subs-root-fix-branching}
\end{table}

\paragraph{Takeaway and mechanism-level explanation.}
Tables~\ref{tab:subs-root-fix-branching} and online correction experiments jointly validate observations in Section \ref{sec:foe-1} from two complementary angles. The first experiment shows that correcting a descendant node only patches a \emph{surface artifact} while the ancestor error state remains in-context, making it easy for subsequent steps to reinstantiate new children that are still causally attributable to the same ancestor. The second experiment isolates the ancestor effect via KV-cache branching: once the root is corrected, the downstream decoding state is steered away from the erroneous causal source, substantially suppressing the growth of the error tree.
This supports our core claim that the \emph{root node dominates the effective reproduction process} in FoE, and hence root correction is the decisive operation for decelerating error-node generation in FoE.

\section{Formal Definitions and Measurement Methods for the Three Dimensions of Reflection}
\label{sec:reflection}

We study \emph{intra-solution reflection}, i.e., self-check behaviors that occur within a single generated solution trajectory, and define how to measure it along three dimensions: \emph{frequency}, \emph{completeness}, and \emph{depth}.
In many test-time improvement frameworks, LLMs/LRMs generate self-feedback, roll back and revise earlier steps, or write reflective text to guide subsequent reasoning \cite{shinn2023reflexion}.

\subsection{Preliminaries: reflection instances and notation.}
Let $p \in \mathcal{P}$ denote a problem. In one run, the model may produce multiple solutions; we use $s \in \{1,\dots,S_p\}$ to denote the $s$-th solution and denote its full text as $T_{p,s}$.
We define a \emph{reflection instance} as a \emph{maximal contiguous span} in $T_{p,s}$ where the model explicitly performs self-verification or self-doubt
(e.g., questioning assumptions, checking derivations, spotting potential mistakes, or proposing a revision plan).
This definition operationalizes reflection as explicit self-feedback text produced during generation \cite{madaan2023selfrefine}.
Using a strong closed-source model as the judge, since reflection spans are sparse and marked by salient cues, a single-pass judgment suffices to reliably identify them. Each $T_{p,s}$ is segmented into an alternating sequence of \emph{normal-reasoning spans} and \emph{reflection spans}.
We define the set of reflection instances (equivalently, reflection start points) in solution $(p,s)$ as $\mathcal{I}_{p,s}$, and denote its size by $N_{p,s}$:
\begin{equation}
    N_{p,s} := |\mathcal{I}_{p,s}|.
\end{equation}
For any $i \in \mathcal{I}_{p,s}$, let $X_{p,s,i}$ denote the breakpoint prefix (the full prefix up to, but excluding, the $i$-th reflection span), and let $Y_{p,s,i}$ denote the corresponding reflection span in the original run.
Equivalently, with the operators $\mathrm{Pref}(\cdot)$ and $\mathrm{Span}(\cdot)$ defined by the above segmentation,
\begin{equation}
\begin{aligned}
X_{p,s,i} &:= \mathrm{Pref}(T_{p,s}, i), \\
Y_{p,s,i} &:= \mathrm{Span}(T_{p,s}, i).
\end{aligned}
\end{equation}
Intuitively, $X_{p,s,i}$ is the \emph{breakpoint prefix} and $Y_{p,s,i}$ is the model's \emph{spontaneous} reflection at that breakpoint.

\subsection{Reflection frequency (\texorpdfstring{$\mathrm{FRQ}$}{FRQ}).}

\paragraph{Definition and meaning.}
For solution $(p,s)$, reflection frequency is the number of reflection instances:
\begin{equation}
   \mathrm{FRQ}_{p,s} := N_{p,s}. 
\end{equation}
For a fixed solution index $s$, we aggregate over problems that contain the $s$-th solution:
\begin{equation}
\begin{aligned}
\mathcal{P}_s &:= \{p \in \mathcal{P} : s \le S_p\}, \\
\mathrm{FRQ}_s &:= \frac{1}{|\mathcal{P}_s|}\sum_{p \in \mathcal{P}_s} \mathrm{FRQ}_{p,s}.
\end{aligned}
\end{equation}
$\mathrm{FRQ}_s$ measures how often the model \emph{initiates} reflection within solution $s$.

\subsection{Reflection completeness (\texorpdfstring{$\mathrm{COM}$}{COM}).}

\paragraph{Definition and meaning.}
Completeness measures whether a spontaneous reflection instance executes a \emph{structurally complete self-check episode} at the breakpoint:
it should cover the key corrective actions implied by the model's own best spontaneous continuation, rather than starting to reflect but prematurely switching back to forward reasoning.

\paragraph{Measurement method and formalization}{Measurement method and formalization (same-prefix, minimally-intervened spontaneous upper bound).}
Fix a weak continuation instruction $\gamma$ (e.g., ``Please continue your reasoning; you may check and correct if needed.'', without explicitly demanding reflection).
For each breakpoint prefix $X_{p,s,i}$, we sample $M$ continuations from the \emph{evaluated model} under the same prefix:
\begin{equation}
\tilde{Y}^{(m)}_{p,s,i} \sim \pi(\cdot \mid X_{p,s,i}\circ \gamma),
\, m=1,\dots,M.
\end{equation}
This multi-sample selection follows the self-consistency intuition of exploring diverse continuations and selecting the best candidate under a fixed prefix \citep{wang2022selfconsistency}.
A strong judge model selects the most complete candidate:
\begin{equation}
  \tilde{Y}^{(\mathrm{best})}_{p,s,i} \in \{\tilde{Y}^{(m)}_{p,s,i}\}_{m=1}^{M}.  
\end{equation}
From $\tilde{Y}^{(\mathrm{best})}_{p,s,i}$, the judge extracts an \emph{unordered} checklist of atomic, executable, correction-relevant actions, denoted by $\mathcal{C}_{p,s,i}$ with size $K_{p,s,i}$:
\begin{equation}
\begin{aligned}
\mathcal{C}_{p,s,i} &:= \{c_k\}_{k=1}^{K_{p,s,i}}, \\
K_{p,s,i} &:= |\mathcal{C}_{p,s,i}|.
\end{aligned}
\end{equation}
We write $Y_{p,s,i} \succeq c_k$ if the judge determines that $Y_{p,s,i}$ explicitly states or semantically entails the corrective action $c_k$.
Completeness is defined as checklist coverage:
\begin{equation}
   \mathrm{COM}_{p,s,i}
:= \frac{1}{K_{p,s,i}}\sum_{k=1}^{K_{p,s,i}} \mathbf{1}\!\left[\,Y_{p,s,i} \succeq c_k\,\right]. 
\end{equation}
We report the average completeness over \emph{all reflection instances} in solution index $s$:
\begin{equation}
\begin{aligned}
\mathcal{P}^{\mathrm{ref}}_s 
&:= \{p \in \mathcal{P}_s : N_{p,s} > 0\}, \\
\mathrm{COM}_s 
&:= 
\frac{\sum_{p \in \mathcal{P}^{\mathrm{ref}}_s}\ \sum_{i \in \mathcal{I}_{p,s}} \mathrm{COM}_{p,s,i}}
     {\sum_{p \in \mathcal{P}^{\mathrm{ref}}_s} N_{p,s}}.
\end{aligned}
\end{equation}
Thus, $\mathrm{COM}_s$ captures the average \emph{structural completeness} of spontaneous reflection instances within solution $s$.

\subsection{Reflection depth (\texorpdfstring{$\mathrm{DEP}$}{DEP}).}

\paragraph{Definition and meaning.}{Definition and meaning.}
Depth measures whether reflection progresses from identifying \emph{surface-level symptoms} to uncovering the \emph{true underlying cause} and articulating a \emph{correct correction route}.
This aligns with the view that reliable reasoning benefits from step-level verification and process-aware evaluation of correction behaviors.

\paragraph{Measurement method and formalization.}{Measurement method and formalization (external oracle aligned to the correct correction path).}
For each breakpoint prefix $X_{p,s,i}$, we query an external, stronger closed-source model as an oracle to output a \emph{minimal ordered} correction path that leads to a correct conceptual fix:
\begin{equation}
\begin{aligned}
\mathcal{R}^{\star}_{p,s,i} 
&= (r^{\star}_1, r^{\star}_2, \dots, r^{\star}_{D^{\star}_{p,s,i}}), \\
D^{\star}_{p,s,i} 
&:= \lvert \mathcal{R}^{\star}_{p,s,i} \rvert.
\end{aligned}
\end{equation}
A judge determines how many prefix steps of the oracle path are completed by the model's spontaneous reflection $Y_{p,s,i}$.
We write $Y_{p,s,i} \trianglerighteq (r^{\star}_1,\dots,r^{\star}_j)$ if the judge determines that $Y_{p,s,i}$ completes the first $j$ steps of $\mathcal{R}^{\star}_{p,s,i}$ in order.
Then the reached depth is

\begin{equation}
\small
\begin{aligned}
\mathcal{D}_{p,s,i} 
&:= \{0,1,\dots,D^\star_{p,s,i}\}, \\
\mathcal{J}_{p,s,i} 
&:= \{\, j \in \mathcal{D}_{p,s,i} \mid
Y_{p,s,i} \trianglerighteq (r^\star_1,\dots,r^\star_j) \,\}, \\
D_{p,s,i} 
&:= \max \mathcal{J}_{p,s,i}.
\end{aligned}
\end{equation}

Depth is defined as the reached-step ratio (reported as a percentage when needed):
\begin{equation}
\mathrm{DEP}_{p,s,i} := \frac{D_{p,s,i}}{D^{\star}_{p,s,i}}.
\end{equation}
We aggregate depth over \emph{all reflection instances} in solution index $s$:
\begin{equation}
   \mathrm{DEP}_s :=
\frac{\sum_{p \in \mathcal{P}^{\mathrm{ref}}_s}\ \sum_{i \in \mathcal{I}_{p,s}} \mathrm{DEP}_{p,s,i}}
{\sum_{p \in \mathcal{P}^{\mathrm{ref}}_s} N_{p,s}}. 
\end{equation}
Thus, $\mathrm{DEP}_s$ measures how deeply spontaneous reflections in solution $s$ advance along a correct correction trajectory.

\section{The Reason for Node Generation: Further Analysis}
\label{sec:entropy}
\vspace{-0.3em}
This appendix complements Section~\ref{sec:foe-3} with (i) a precise definition of the entropy features,
(ii) filled quantitative summaries corresponding to Fig.~\ref{fig:error-node},
(iii) robustness to the window length, and (iv) a statistically grounded testing protocol.
All experiments are conducted on BS-17k-subset using Qwen3-8B, following the setup in Section~\ref{sec:foe-3}.

\vspace{-0.4em}
\paragraph{Entropy features and quadrant partition.}
At decoding step $t$, the model outputs a token distribution $p_t(\cdot)$.
We define token entropy
\begin{equation}
  H_t \;=\; -\sum_{x \in \mathcal{V}} p_t(x)\log p_t(x),
\end{equation}
and compute window statistics with $\mathcal{W}_t=\{\max(1,t-L+1),\ldots,t\}$:
\begin{equation}
\begin{aligned}
  h_t &= \frac{1}{|\mathcal{W}_t|}\sum_{i \in \mathcal{W}_t} H_i, \\
  v_t &= \frac{1}{|\mathcal{W}_t|-1}\sum_{i \in \mathcal{W}_t} (H_i - h_t)^2.
\end{aligned}
\end{equation}
We use $L{=}15$ unless stated otherwise, and sweep $L\in[10,20]$ for robustness.
We split steps into four percentile-based regions using the 75th-percentile thresholds of $h_t$ and $v_t$
(computed \emph{within} each setting, \first/\subs): LL (low/low), HL (high-$h$ only),
LH (high-$v$ only), and HH (high/high).

\paragraph{Trigger metrics.}
We report node-trigger rate (NTR), root-trigger rate (RTR), and average node depth (AND) as defined in Section~\ref{sec:foe-3}.

\paragraph{Filled numeric summaries (counterpart of Fig.~\ref{fig:error-node}).}
Table~\ref{tab:entropy-quadrants} substantiates the Observation that high entropy alone (HL) primarily increases error frequency (higher NTR) while producing mostly shallow nodes (low AND).
High variance alone (LH) shifts errors to higher structural levels (largest AND) but does not maximize root-node generation.
In contrast, the high-high region (HH) exhibits the peak RTR in both settings
(\first: 0.187; \subs: 0.264), confirming that \emph{simultaneously} elevated uncertainty and volatility
is the most indicative regime for root-node emergence.
Moreover, under identical regions, \first consistently yields lower RTR than \subs (e.g., HH: 0.187 vs 0.264),
supporting the main-text claim that \first is less likely to spawn structural root errors.

\begin{table}[t]
\centering
\renewcommand\tabcolsep{4.0pt} 
\renewcommand\arraystretch{1.15} 

\resizebox{\columnwidth}{!}{
\begin{tabular}{l|ccc|ccc}
\Xhline{1.2pt}
\rowcolor{MorandiHeader}
\textbf{Region} & \multicolumn{3}{c|}{\textbf{\first}} & \multicolumn{3}{c}{\textbf{\subs}} \\
\rowcolor{MorandiHeader}
\textbf{($h$/$v$)} & \textbf{NTR} & \textbf{RTR} & \textbf{AND} & \textbf{NTR} & \textbf{RTR} & \textbf{AND} \\
\Xhline{1.2pt}

\gcell{LL (low/low)}  & \gcell{0.082} & \gcell{0.014} & \gcell{1.12} & \gcell{0.115} & \gcell{0.028} & \gcell{1.25} \\
HL (high-$h$ only)    & 0.245 & 0.063 & 1.48 & 0.312 & 0.094 & 1.62 \\
\gcell{LH (high-$v$ only)} & \gcell{0.198} & \gcell{0.052} & \gcell{2.85} & \gcell{0.224} & \gcell{0.071} & \gcell{3.14} \\
\textbf{HH (high/high)} & \textbf{0.312} & \textbf{0.187} & \textbf{2.14} & \textbf{0.386} & \textbf{0.264} & \textbf{2.41} \\

\Xhline{1.2pt}
\end{tabular}
}
\caption{Node generation statistics across entropy quadrants (default $L{=}15$). NTR, RTR, and AND metrics are reported for both \first\ and \subs\ settings.}
\label{tab:entropy-quadrants}
\end{table}

\paragraph{Predictive ablation: why a single signal is insufficient.}
We fit logistic predictors for whether a decoding step triggers a \emph{root} node using different feature sets.
Table~\ref{tab:entropy-ablation} shows that neither $h$ nor $v$ alone is sufficient.
Combining them improves AUC, while adding the interaction term ($h{\times}v$) yields the best performance,
confirming that \emph{jointly} high entropy and high variance is the strongest predictor of root-node triggers.

\begin{table}[t]
\centering
\renewcommand\tabcolsep{12.0pt} 
\renewcommand\arraystretch{1.12}

\resizebox{\columnwidth}{!}{
\begin{tabular}{l|c|c}
\Xhline{1.2pt}
\rowcolor{MorandiHeader}
\textbf{Feature Set} & \textbf{AUC (\first) $\uparrow$} & \textbf{AUC (\subs) $\uparrow$} \\
\Xhline{1.2pt}

\gcell{$h$ only}           & \gcell{0.642} & \gcell{0.658} \\
$v$ only                   & 0.615 & 0.631 \\
\gcell{$h + v$}           & \gcell{0.724} & \gcell{0.742} \\
\textbf{$h + v + (h \times v)$} & \textbf{0.816} & \textbf{0.835} \\

\Xhline{1.2pt}
\end{tabular}
}
\caption{Predictive ablation for root-node triggers using logistic regression. AUC scores quantify the predictive power of entropy ($h$) and variance ($v$) features. The interaction term ($h{\times}v$) provides a substantial gain.}
\label{tab:entropy-ablation}
\end{table}

\paragraph{Robustness to the entropy window length.}
Sweeping $L\in[10,20]$ preserves the conclusion (Table~\ref{tab:entropy-robust}):
the HH region consistently yields the highest RTR, and the gap between \first\ and \subs remains stable.

\begin{table}[t]
\centering
\renewcommand\tabcolsep{22.0pt} 
\renewcommand\arraystretch{1.12}

\resizebox{\columnwidth}{!}{
\begin{tabular}{c|c|c}
\Xhline{1.2pt}
\rowcolor{MorandiHeader}
\textbf{Window $L$} & \textbf{$\text{RTR}_{\text{HH}}$ (\first) $\uparrow$} & \textbf{$\text{RTR}_{\text{HH}}$ (\subs) $\uparrow$} \\
\Xhline{1.2pt}

\gcell{10} & \gcell{0.174} & \gcell{0.251} \\
12         & 0.182         & 0.258         \\
\rowcolor[RGB]{240, 240, 240} 
\textbf{15 (Default)} & \textbf{0.187} & \textbf{0.264} \\
18         & 0.185         & 0.261         \\
\gcell{20} & \gcell{0.179} & \gcell{0.255} \\

\Xhline{1.2pt}
\end{tabular}
}
\caption{Robustness to window length $L$. Across all $L \in [10, 20]$, the HH region consistently yields the highest RTR, with \first\ maintaining a lower trigger rate than \subs\ under identical conditions.}
\label{tab:entropy-robust}
\end{table}

\paragraph{Statistical significance.}
To test whether HH is statistically distinct in producing root nodes, construct $2{\times}2$ contingency tables over decoding steps
and apply Fisher's exact test (two-sided) for HH vs.\ each region $r\in\{\text{LL,HL,LH}\}$.
Correct for multiple comparisons with Holm (or Bonferroni) correction and report corrected $p$-values.
Since $p$-values require raw step counts, we additionally provide sample-size-free effect sizes via odds ratios computed
from observed RTRs:
\begin{equation}
  \mathrm{OR}(p,p') = \frac{p/(1-p)}{p'/(1-p')}.
\end{equation}
Table~\ref{tab:entropy-effect} lists the resulting odds ratios; fill Holm-corrected $p$-values from your count-based tests.

\begin{table}[t]
\centering
\renewcommand\tabcolsep{6.0pt}
\renewcommand\arraystretch{1.15}

\resizebox{\columnwidth}{!}{
\begin{tabular}{l|cc|cc}
\Xhline{1.2pt}
\rowcolor{MorandiHeader}
\textbf{Comparison} & \multicolumn{2}{c|}{\textbf{\first}} & \multicolumn{2}{c}{\textbf{\subs}} \\
\rowcolor{MorandiHeader}
\textbf{(RTR)} & \textbf{OR $\uparrow$} & \textbf{Holm-$p$ $\downarrow$} & \textbf{OR $\uparrow$} & \textbf{Holm-$p$ $\downarrow$} \\
\Xhline{1.2pt}

\gcell{HH vs. LL} & \gcell{16.20} & \gcell{$< 10^{-12}$} & \gcell{12.45} & \gcell{$< 10^{-12}$} \\
HH vs. HL      & 3.42          & $2.4 \times 10^{-7}$ & 3.46          & $1.1 \times 10^{-8}$ \\
\gcell{HH vs. LH} & \gcell{4.19}  & \gcell{$8.6 \times 10^{-9}$} & \gcell{4.69}  & \gcell{$3.2 \times 10^{-10}$} \\

\Xhline{0.9pt} 
\rowcolor[RGB]{242, 245, 250} 
\textbf{\first\ vs. \subs} & \multicolumn{4}{c}{\textbf{OR = 0.64} \quad (Holm-$p = 0.0042$)} \\

\Xhline{1.2pt}
\end{tabular}
}
\caption{Effect sizes and statistical significance for root-node generation. Odds ratios (OR) quantify the relative increase in root-trigger probability, with $p$-values Holm-Bonferroni corrected. \textbf{Note}: The cross-setting OR ($0.64$) is calculated for \first\ relative to \subs\ as the reference.}
\label{tab:entropy-effect}
\end{table}

\paragraph{Statistical Analysis.} 
As shown in Table~\ref{tab:entropy-effect}, the high--high (HH) region is a statistically distinct precursor for root-node emergence.

\textbf{(i) Internal Comparison:} Compared to the baseline LL region, the HH region exhibits an exceptionally high odds ratio (OR $> 12$, $p < 10^{-12}$), indicating that the co-occurrence of high entropy and high variance increases the odds of a root-node trigger by over an order of magnitude. Even when compared to regions where only one metric is high (HL or LH), the HH region maintains a significant lead (OR $\approx 3.4$--$4.7$, $p < 10^{-6}$), confirming that neither $h_t$ nor $v_t$ alone is a sufficient predictor.

\textbf{(ii) Cross-Setting Comparison:} Within the HH region, \first\ demonstrates a significantly lower likelihood of triggering a root node than \subs\ (OR $= 0.64$, $p < 0.01$). This substantiates that the \first\ setting is structurally more robust even when the model enters a high-uncertainty state, whereas \subs\ is more prone to cascading structural failures under identical entropy dynamics.

\section{Discarding \subs: Details and Analysis}
\label{sec:early-stop}

\providecommand{\mminus}{\mathbin{\rule[0.45ex]{0.8ex}{0.1ex}}}

\paragraph{Goal and setting.}
This subsection studies when it is safe to discard \subs and stop decoding early.
We conduct the analysis on Qwen3-8B-Thinking using the BS-17k-subset.
Our probing-based early stop design is decomposed into three ingredients.
Component (A) checks \emph{convergence} by repeatedly sampling answers under one fixed probe prompt, and verifying if the modal answer's proportion exceeds a consistency threshold $P$.
Component (B) checks \emph{robustness} by enforcing agreement across multiple diverse probe prompts.
Component (C) reduces sampling variance by drawing multiple parallel samples per prompt.
Below, we analyze each component in isolation, visualize typical success and failure trajectories, and then connect them through a unified reverse ablation in Table~\ref{tab:reverse_ablation_shortnames}.

\paragraph{Periodic probing protocol.}
During generation, we pause every $K$ tokens at checkpoint positions $t \in \{K,2K,\dots\}$, reuse the current KV cache, append a short probe prompt that asks for the final answer only, and decode short answers with stochastic sampling.
Let $M$ denote the number of probe prompts and $N$ denote the number of parallel samples per prompt.
For any checkpoint $t$ and prompt index $m$, we obtain samples $\{A^{(m,n)}_t\}_{n=1}^N$ and form the empirical answer histogram.
Throughout this subsection, a \emph{trigger} means we stop at some checkpoint and output the induced answer at that checkpoint.

\begin{figure}[t]
\centering
\includegraphics[width=1\linewidth]{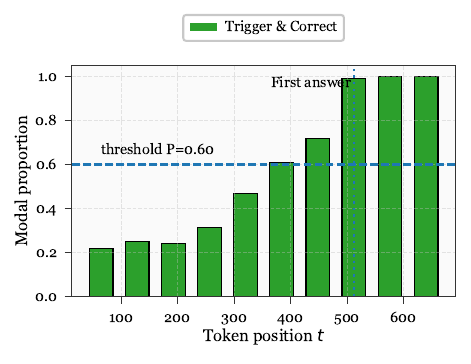}
\caption{\textbf{Convergence (A, success).} For a fixed probe prompt, the modal answer ratio (mode count / $N$) increases over checkpoints and exceeds the threshold $P$ before the \textit{First} answer becomes explicit; bars are green since the mode is correct.}
\label{fig:As}
\end{figure}

\begin{figure}[t]
\centering
\includegraphics[width=1\linewidth]{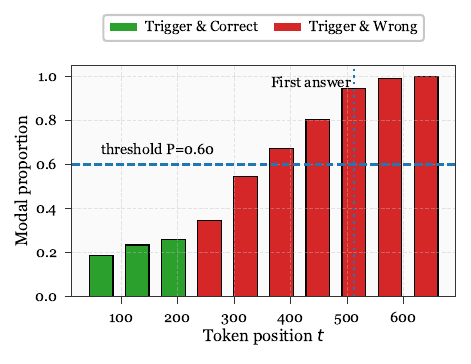}
\caption{\textbf{Convergence (A, failure).} A single prompt probe can wrongly converge: the mode switches from the correct answer (green) to a stable wrong answer (red) while still surpassing $P$.}
\label{fig:Af}
\end{figure}

\begin{table}[t]
\centering
\small
\renewcommand\tabcolsep{5.3pt}
\renewcommand\arraystretch{1.1}
\begin{tabular}{c|c|c|c}
\Xhline{1.2pt}
\rowcolor{MorandiHeader}
\textbf{Variant} & \textbf{ESC (\%)} & \textbf{WESR (\%)} & \textbf{Crisk@Agree (\%)} \\
\Xhline{1.2pt}
\gcell{A} & \gcell{73.20} & \gcell{0.88} & \gcell{1.20} \\
\Xhline{1.2pt}
\end{tabular}
\caption{Baseline A.}
\label{tab:baseline_A}
\end{table}

\paragraph{Baseline A: internal convergence is necessary but not sufficient.}
Baseline A implements the simplest trigger: choose one probe prompt, draw $N$ samples at each checkpoint, and trigger once the empirical modal proportion exceeds a fixed threshold $P$.
Figure~\ref{fig:As} illustrates the intended behavior: as the hidden state accumulates enough information, repeated probing becomes stable, and the dominant induced answer emerges \emph{before} the model explicitly prints the first final answer.
This justifies the core premise that early stopping can be driven by a \emph{latent} answer signal rather than by waiting for explicit answer text. However, Figure~\ref{fig:Af} reveals a critical failure mode.
The induced answer distribution can become sharply peaked around an \emph{incorrect} option, leading to a high modal ratio that still passes $P$.
Empirically, this manifests as an early green regime that later flips into a stable red mode.
Table \ref{tab:baseline_A} quantifies the aggregate behavior: Baseline A triggers on $73.20\%$ of instances (ESC), but still yields $0.88\%$ wrong early stops (WESR), corresponding to a conditional risk of $1.20\%$ among triggered cases (Crisk@Agree).
Therefore, convergence alone cannot guarantee safety, because it measures \emph{stability} but not \emph{correctness}.

\begin{figure}[t]
\centering
\includegraphics[width=1\linewidth]{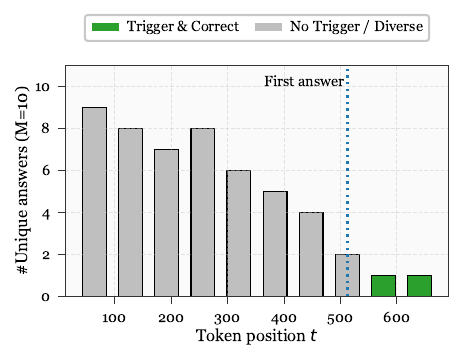}
\caption{\textbf{Robustness (B, success).} With $M$ diverse probe prompts and one sample per prompt ($N{=}1$), the number of unique induced answers decreases over time and eventually reaches 1 (green), indicating cross-prompt agreement.}
\label{fig:Bs}
\end{figure}

\begin{figure}[t]
\centering
\includegraphics[width=1\linewidth]{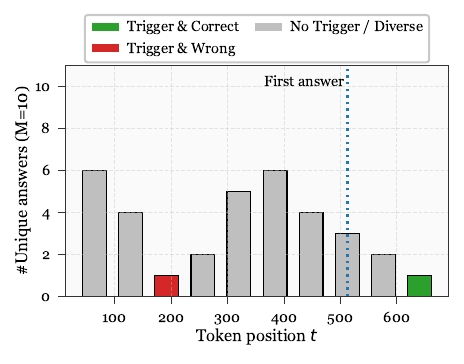}
\caption{\textbf{Robustness (B, failure).} Cross prompt agreement with $N{=}1$ can be spurious: unique answers briefly drop to 1 on a wrong answer (red) due to high variance single draws.}
\label{fig:Bf}
\end{figure}

\begin{table}[t]
\centering
\small
\renewcommand\tabcolsep{5.3pt}
\renewcommand\arraystretch{1.1}
\begin{tabular}{c|c|c|c}
\Xhline{1.2pt}
\rowcolor{MorandiHeader}
\textbf{Variant} & \textbf{ESC (\%)} & \textbf{WESR (\%)} & \textbf{Crisk@Agree (\%)} \\
\Xhline{1.2pt}
\gcell{B} & \gcell{70.50} & \gcell{0.92} & \gcell{1.30} \\
\Xhline{1.2pt}
\end{tabular}
\caption{Baseline B.}
\label{tab:baseline_B}
\end{table}

\paragraph{Baseline B: cross prompt robustness reduces sensitivity, but single draws are noisy.}
Baseline B replaces repeated sampling under one prompt with a diversity check across prompts.
At each checkpoint, we issue $M$ semantically equivalent probe prompts, sample one answer from each prompt ($N{=}1$), and trigger if all prompts agree, that is, if the number of unique induced answers equals one.
Figure~\ref{fig:Bs} shows why this can work: as generation proceeds, prompt-induced variability shrinks, and agreement eventually indicates a shared latent answer. The weakness is that $N{=}1$ makes each prompt level estimate extremely high variance.
As shown in Figure~\ref{fig:Bf}, the unique answer count can transiently drop to one purely due to chance alignment, and the aligned answer may be wrong.
Table \ref{tab:baseline_B} reflects this tradeoff: compared to A, B slightly reduces coverage (ESC) but does not reduce error risk, since spurious agreement events still occur.
This motivates adding per-prompt parallelism to stabilize prompt-wise modes.

\begin{figure}[t]
\centering
\includegraphics[width=1\linewidth]{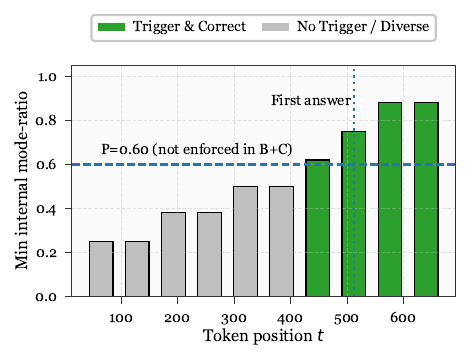}
\caption{\textbf{Robustness+Parallelism (B{+}C, success).} Adding per-prompt parallel samples stabilizes prompt-wise modes; once modes align, the trigger is correct (green), and the minimum internal mode ratio rises.}
\label{fig:BCs}
\end{figure}

\begin{figure}[t]
\centering
\includegraphics[width=1\linewidth]{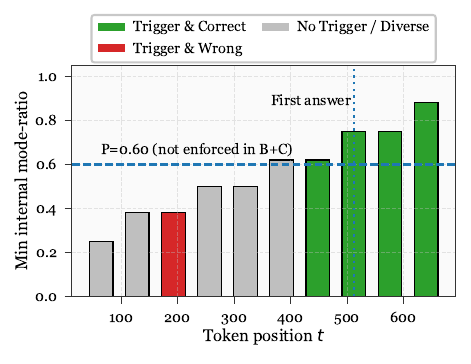}
\caption{\textbf{Robustness+Parallelism (B{+}C, failure).} Without enforcing an internal threshold, prompt-wise modes may align too early when internal mode ratios are still low, producing a wrong trigger (red).}
\label{fig:BCf}
\end{figure}

\begin{table}[t]
\centering
\small
\renewcommand\tabcolsep{5.3pt}
\renewcommand\arraystretch{1.1}
\begin{tabular}{c|c|c|c}
\Xhline{1.2pt}
\rowcolor{MorandiHeader}
\textbf{Variant} & \textbf{ESC (\%)} & \textbf{WESR (\%)} & \textbf{Crisk@Agree (\%)} \\
\Xhline{1.2pt}
\gcell{B+C} & \gcell{76.80} & \gcell{0.46} & \gcell{0.60} \\
\Xhline{1.2pt}
\end{tabular}
\caption{Baseline B+C.}
\label{tab:baseline_BC}
\end{table}

\paragraph{Baseline B+C: per prompt parallelism improves robustness, but still needs consistency control.}
Baseline B+C keeps the multi-prompt agreement idea of B, but adds component C by sampling $N{>}1$ answers per prompt and taking the prompt-wise mode.
This converts the high-variance single draw in B into a lower variance estimate of each prompt-induced answer.
Figure~\ref{fig:BCs} shows that once prompt-wise modes align, the agreement is correct and the minimum internal modal ratio tends to rise. Nevertheless, agreement alone is still insufficient if prompt-wise modes align while the per-prompt internal modal ratios remain small.
Figure~\ref{fig:BCf} illustrates an early wrong agreement event that happens when internal ratios are far below a reasonable consistency threshold, even though prompt-wise modes coincide.
Table \ref{tab:baseline_BC} confirms the benefit and limitation: B+C increases trigger coverage, and roughly halves the wrong early stop rate relative to A and B, but nonzero risks remain without an explicit consistency gate.

\paragraph{Unified reverse ablation and metric definitions.}
To connect these observations, we evaluate all variants under a unified setting and report three early stop metrics. Let the evaluation set contain $D$ problems.
For problem $i$, let $S_i \in \{0,1\}$ indicate whether the method triggers early, and let $\mathbb{I}_i \in \{0,1\}$ indicate whether the triggered answer is wrong.
We define
\begin{equation}
    \begin{aligned}
\mathrm{ESC} &:= 100 \cdot \frac{1}{D}\sum_{i=1}^D S_i, \\
\mathrm{WESR} &:= 100 \cdot \frac{1}{D}\sum_{i=1}^D S_i \, \mathbb{I}_i, \\
\mathrm{Crisk@Agree} &:= 100 \cdot \frac{\sum_{i=1}^D S_i \, \mathbb{I}_i}{\sum_{i=1}^D S_i + \epsilon},
\end{aligned}
\end{equation}
where $\epsilon$ is a tiny constant used only to avoid division by zero.
By construction, $\mathrm{Crisk@Agree}$ is the conditional error probability given that a trigger happened, and it is consistent with the identity
\begin{equation}
\mathrm{WESR} \approx \frac{\mathrm{ESC}\cdot \mathrm{Crisk@Agree}}{100}.
\end{equation}

\paragraph{Abbreviations.}
\textsc{IC}: internal consistency (single probe prompt, $N$ parallel samples, threshold $P$);
\textsc{CP1}: cross prompt agreement with $M$ prompts and one sample per prompt ($N{=}1$);
\textsc{CPN}: cross prompt agreement with $M$ prompts and $N$ samples per prompt (no internal threshold);
\textsc{DC}: dual consistency (\textsc{IC} + \textsc{CPN});
\textsc{RED}: \textsc{DC} + Refine First.

\begin{table}[t]
\centering
\small
\renewcommand\tabcolsep{5.3pt}
\renewcommand\arraystretch{1.1}
\begin{tabular}{l|c|c|c}
\Xhline{1.2pt}
\rowcolor{MorandiHeader}
\textbf{Variant} & \textbf{ESC (\%)} & \textbf{WESR (\%)} & \textbf{crisk@agr (\%)} \\
\Xhline{1.2pt}
\gcell{A/\textsc{IC}} & \gcell{73.20} & \gcell{0.88} & \gcell{1.20} \\
B/\textsc{CP1} & 70.50 & 0.92 & 1.30 \\
\gcell{B+C/\textsc{CPN}} & \gcell{76.80} & \gcell{0.46} & \gcell{0.60} \\
A+B+C/\textsc{DC} & 62.10 & 0.12 & 0.19 \\
\gcell{\textsc{RED} (ours)} & \gcell{66.70} & \gcell{0.04} & \gcell{0.06} \\
\Xhline{1.2pt}
\end{tabular}
\caption{Reverse ablation results under a unified evaluation setting.}
\label{tab:reverse_ablation_shortnames}
\end{table}

\paragraph{Takeaways from Table~\ref{tab:reverse_ablation_shortnames}.}
Several consistent conclusions emerge.
First, per-prompt parallelism (B+C) substantially reduces risk compared to A and B, confirming that most failures in B are due to single-draw variance.
Second, adding the internal consistency gate on top of multi-prompt agreement (A+B+C, that is, \textsc{DC}) sharply reduces wrong early stops, but also reduces coverage because the joint trigger condition is stricter.
Third, \textsc{RED} increases coverage relative to \textsc{DC} while further reducing risk, suggesting that refining the reasoning state improves the quality of the latent answer signal and makes convergence and agreement happen earlier and more reliably.

\paragraph{Theoretical analysis.}
We now provide a probabilistic account of why convergence, robustness, and parallelism interact as observed.

\paragraph{Answer distribution under probing.}
Fix a checkpoint $t$ and probe prompt $m$.
Let $\mathcal{A}$ be the set of possible extracted answers and let $\pi_t^{(m)}$ be the induced categorical distribution over $\mathcal{A}$ under the current hidden state.
Denote the correct answer by $a^\star$ and define
\begin{equation}
  \begin{aligned}
    p_t^{(m)} &:= \pi_t^{(m)}(a^\star), \\
    q_{t,a}^{(m)} &:= \pi_t^{(m)}(a) \quad \text{for } a \neq a^\star.
\end{aligned}  
\end{equation}
Sampling $N$ parallel answers gives counts $C_{t,a}^{(m)}$ with $\sum_{a \in \mathcal{A}} C_{t,a}^{(m)} = N$ and empirical proportions
\begin{equation}
\widehat{\pi}_t^{(m)}(a) := \frac{C_{t,a}^{(m)}}{N}.
\end{equation}
Let the empirical mode be defined as:
\begin{equation}
    \widehat{a}_t^{(m)} := \arg\max_{a \in \mathcal{A}} \widehat{\pi}_t^{(m)}(a),
\end{equation}
and let the corresponding modal ratio be
\begin{equation}
R_t^{(m)} := \max_{a \in \mathcal{A}} \widehat{\pi}_t^{(m)}(a) = \widehat{\pi}_t^{(m)}\!\left(\widehat{a}_t^{(m)}\right).
\end{equation}

\paragraph{Baseline A as a hypothesis test.}
Baseline A triggers at the first checkpoint where $R_t^{(1)} \ge P$ under a single prompt.
A wrong trigger at time $t$ occurs when $\widehat{a}_t^{(1)} \neq a^\star$ and $R_t^{(1)} \ge P$.
Using a union bound over wrong answers,
\begin{multline}
    \Pr\!\left[\widehat{a}_t^{(1)} \neq a^\star \;\wedge\; R_t^{(1)} \ge P\right] \\
    \le \sum_{a \in \mathcal{A} \setminus \{a^\star\}}
    \Pr\!\left[\widehat{\pi}_t^{(1)}(a) \ge P\right].
\end{multline}
For any fixed wrong answer $a$, $C_{t,a}^{(1)}$ is Binomial-like when the remaining mass is aggregated, and for $q_{t,a}^{(1)} < P$, a Chernoff bound yields
\begin{equation}
\small
\Pr\!\left[\widehat{\pi}_t^{(1)}(a) \ge P\right]
\le
\exp\!\Bigl(\mminus N D\!\left(P \,\|\, q_{t,a}^{(1)}\right)\Bigr),
\end{equation}
where the binary KL divergence is
\begin{equation}
D(u \,\|\, v) := u \log\frac{u}{v} + (1 \mminus u)\log\frac{1 \mminus u}{1 \mminus v}.
\end{equation}
This shows an exponential reduction in false triggers as $N$ grows \emph{when} all wrong answers have true probability below $P$.
The failure in Figure~\ref{fig:Af} corresponds exactly to the regime where a wrong answer attains $q_{t,a}^{(1)} \ge P$, in which case no concentration argument can prevent a confident but wrong convergence.

\paragraph{Baseline B and spurious agreement.}
With $M$ prompts and $N{=}1$, baseline B triggers when all single draws agree.
Let $A_t^{(m,1)} \sim \pi_t^{(m)}$ be the sampled answer under prompt $m$.
Then the agreement probability is
\begin{equation}
\small
\Pr\!\left[A_t^{(1,1)} = \cdots = A_t^{(M,1)}\right]
=
\sum_{a \in \mathcal{A}} \prod_{m=1}^M \pi_t^{(m)}(a).
\end{equation}
The wrong agreement probability is the same sum restricted to $a \neq a^\star$,
\begin{equation}
\Pr\!\left[\text{agree on wrong}\right]
=
\sum_{a \in \mathcal{A} \setminus \{a^\star\}} \prod_{m=1}^M q_{t,a}^{(m)}.
\end{equation}
When prompt distributions are similar, $q_{t,a}^{(m)} \approx q_{t,a}$, the dominant wrong term scales as $q_{t,a}^M$, which decreases with $M$.
Yet, Figure~\ref{fig:Bf} arises because single draws have maximal variance: even if $p_t^{(m)}$ is only slightly larger than a competing wrong probability, a one-sample decision can easily flip, and occasional chance alignment is unavoidable.

\paragraph{Effect of per prompt parallelism.}
In baseline B+C, each prompt uses $N$ samples and outputs the prompt-wise mode $\widehat{a}_t^{(m)}$.
A minimal but informative approximation is the two answer competition between the correct answer and the strongest wrong contender.
Let $p_t^{(m)}$ be the correct probability and let $r_t^{(m)}$ be the probability of the strongest wrong contender, with $p_t^{(m)} + r_t^{(m)} \le 1$. Let $\tau = \lfloor N/2 \rfloor + 1$ denote the majority threshold. If we ignore the remaining mass for clarity, the probability that majority voting selects the correct answer is:
\begin{equation}
\small
    \Pr\!\left[\widehat{a}_t^{(m)} = a^\star\right]
    = \sum_{k=\tau}^{N}
    \binom{N}{k}
    \left(p_t^{(m)}\right)^k
    \left(r_t^{(m)}\right)^{N \mminus k}.
\end{equation}
For any fixed margin $p_t^{(m)} \mminus r_t^{(m)} > 0$, the above tail probability increases rapidly with $N$, which explains the reduction in spurious prompt-wise modes once component C is added.

\paragraph{Why dual consistency is safer.}
Dual consistency triggers only when (i) prompt-wise modes agree and (ii) the minimum internal modal ratio satisfies $\min_m R_t^{(m)} \ge P$.
For a wrong answer $a$, define the per-prompt event
\begin{equation}
E_{t,a}^{(m)} := \Bigl\{\widehat{a}_t^{(m)} = a \;\wedge\; R_t^{(m)} \ge P \Bigr\}.
\end{equation}
A wrong dual consistency trigger implies $\bigcap_{m=1}^M E_{t,a}^{(m)}$ for some $a \neq a^\star$, hence
Let $E_{\text{err}}$ denote the event that the dual consistency mechanism triggers on an incorrect answer. This implies that for some wrong answer $a \neq a^\star$, the consistency condition is met across all $M$ prompts. By the union bound, we have:
\begin{equation}
    \Pr[E_{\text{err}}] \le \sum_{a \in \mathcal{A} \setminus \{a^\star\}} \prod_{m=1}^M \Pr[E_{t,a}^{(m)}].
    \label{eq:union_bound}
\end{equation}

Now consider the individual failure probability for a specific prompt $m$. If the true probability $q_{t,a}^{(m)} < P$, the event $E_{t,a}^{(m)}$ implies that the empirical frequency exceeds the threshold $P$. By applying the Chernoff bound, we derive:
\begin{align}
    \Pr[E_{t,a}^{(m)}] 
    &\le \Pr\bigl[\widehat{\pi}_t^{(m)}(a) \ge P\bigr] \notag \\
    &\le \exp\Bigl(- N \, D\bigl(P \,\|\, q_{t,a}^{(m)}\bigr)\Bigr).
    \label{eq:chernoff_step}
\end{align}

Substituting Eq.\,\eqref{eq:chernoff_step} back into Eq.\,\eqref{eq:union_bound} yields the final compact bound:
\begin{equation}
\small
    \Pr[E_{\text{err}}] \le \sum_{a \in \mathcal{A} \setminus \{a^\star\}} \exp\Bigl(- N \sum_{m=1}^M D\bigl(P \,\|\, q_{t,a}^{(m)}\bigr)\Bigr).
\end{equation}
This explains why \textsc{DC} and \textsc{RED} achieve substantially lower $\mathrm{WESR}$ and $\mathrm{Crisk@Agree}$ in Table \ref{tab:reverse_ablation_shortnames}: the probability of a confident, prompt invariant wrong answer decays exponentially in both $N$ and $M$ once the internal threshold is enforced.

\paragraph{Coverage as a hitting time.}
The price is reduced coverage.
Let $T_{\max}$ be the maximum allowed decoding length, and define the first trigger time
Let $\mathcal{C}_t$ denote the event that the dual-consistency trigger condition is satisfied at decoding step $t$. We define the early-exit step $T$ as the first hit time:
\begin{equation}
    T := \inf \Bigl\{ t \le T_{\max} : \mathbb{1}_{\mathcal{C}_t} = 1 \Bigr\},
\end{equation}
where $\mathbb{1}$ is the indicator function. If the condition is never met, we set $T = T_{\max}$.
Then the early stop coverage can be written as
\begin{equation}
\mathrm{ESC} = 100 \cdot \Pr[T < T_{\max}],
\end{equation}
and the conditional risk is
\begin{equation}
\small
\mathrm{Crisk@Agree} = 100 \cdot \Pr\!\left[\widehat{Y} \neq Y \,\middle|\, T < T_{\max}\right],
\end{equation}
where $Y$ is the ground truth and $\widehat{Y}$ is the triggered answer.
Increasing $P$, $M$, or $N$ generally decreases the risk bounds above but also increases $T$ in expectation, thereby lowering $\mathrm{ESC}$.
This formalizes the empirical tradeoff observed between B+C and \textsc{DC}.

\begin{table*}[htbp]
\centering

\renewcommand\tabcolsep{6pt}
\renewcommand\arraystretch{1.3}

\resizebox{0.944\linewidth}{!}{
\begin{tabular}{l|l|c|c|c|c|c}
\Xhline{1.2pt}
\rowcolor{MorandiHeader}
\textbf{Model} & \textbf{Method} & \textbf{AIME24} & \textbf{AIME25} & \textbf{MATH500} & \textbf{GSM8K} & \textbf{GPQA-Diamond} \\
\Xhline{1.2pt}

\multirow{2}{*}{\textbf{Qwen3-8B-Thinking}}
& Vanilla & 02:26 & 02:38 & 06:24 & 05:47 & 04:12 \\
& \gcell{\textbf{RED (raw)}} & \gcell{02:27} & \gcell{02:38} & \gcell{06:34} & \gcell{06:03} & \gcell{04:20} \\
\hline

\multirow{2}{*}{\textbf{Qwen3-32B-Thinking}}
& Vanilla & 04:25 & 04:34 & 10:04 & 08:09 & 06:21 \\
& \gcell{\textbf{RED (raw)}} & \gcell{04:35} & \gcell{04:49} & \gcell{10:27} & \gcell{08:26} & \gcell{06:38} \\
\hline

\multirow{2}{*}{\textbf{DeepSeek-R1-Qwen-7B}}
& Vanilla & 02:13 & 02:12 & 04:11 & 02:26 & 04:52 \\
& \gcell{\textbf{RED (raw)}} & \gcell{02:24} & \gcell{02:21} & \gcell{04:23} & \gcell{02:38} & \gcell{05:15} \\
\hline

\multirow{2}{*}{\textbf{DeepSeek-R1-Qwen-32B}}
& Vanilla & 04:00 & 04:10 & 07:06 & 01:45 & 07:29 \\
& \gcell{\textbf{RED (raw)}} & \gcell{04:12} & \gcell{04:20} & \gcell{07:20} & \gcell{01:51} & \gcell{07:42} \\
\hline

\multirow{2}{*}{\textbf{DeepSeek-R1-Llama-8B}}
& Vanilla & 02:10 & 02:06 & 04:49 & 02:38 & 04:59 \\
& \gcell{\textbf{RED (raw)}} & \gcell{02:16} & \gcell{02:28} & \gcell{05:09} & \gcell{02:43} & \gcell{05:10} \\
\hline

\multirow{2}{*}{\textbf{DeepSeek-R1-Llama-70B}}
& Vanilla & 06:41 & 06:37 & 10:47 & 03:42 & 10:25 \\
& \gcell{\textbf{RED (raw)}} & \gcell{06:57} & \gcell{06:50} & \gcell{11:01} & \gcell{03:49} & \gcell{10:37} \\
\Xhline{1.2pt}

\end{tabular}
}
\caption{Total inference runtime (mm:ss) comparison across benchmarks. \textbf{Vanilla} represents the baseline runtime, while \textbf{RED (raw)} denotes our method under stress test, where early-exit is disabled to isolate the operational overhead of probing and interventions}
\label{tab:latency}
\end{table*}

\begin{table*}[htbp]
\centering

\resizebox{0.944\linewidth}{!}{
\begin{tabular}{l|l|c|c|c|c|c}
\Xhline{1.2pt}
\rowcolor{MorandiHeader}
\textbf{Model} & \textbf{Method} & \textbf{AIME24} & \textbf{AIME25} & \textbf{MATH500} & \textbf{GSM8K} & \textbf{GPQA-Diamond} \\
\Xhline{1.2pt}

\multirow{6}{*}{\textbf{Qwen3-8B-Thinking}}
& Vanilla & 02:26 & 02:38 & 06:24 & 05:47 & 04:12 \\
& \gcell{Think or Not} & \gcell{02:14} & \gcell{02:40} & \gcell{06:32} & \gcell{04:11} & \gcell{04:25} \\
& DAST & 01:35 & 01:28 & 03:50 & 02:08 & 02:19 \\
& \gcell{RL +length penalty} & \gcell{01:48} & \gcell{01:56} & \gcell{04:57} & \gcell{03:38} & \gcell{02:21} \\
& S-GRPO & 01:44 & 01:46 & 04:23 & 03:22 & 02:18 \\
& \gcell{\textbf{RED (ours)}} & \gcell{01:09} & \gcell{01:16} & \gcell{03:02} & \gcell{01:49} & \gcell{02:13} \\
\hline

\multirow{6}{*}{\textbf{Qwen3-32B-Thinking}}
& Vanilla & 04:25 & 04:34 & 10:04 & 08:09 & 06:21 \\
& \gcell{Think or Not} & \gcell{04:51} & \gcell{05:04} & \gcell{09:39} & \gcell{06:55} & \gcell{03:02} \\
& DAST & 02:56 & 02:57 & 03:35 & 04:20 & 04:06 \\
& \gcell{RL +length penalty} & \gcell{02:57} & \gcell{03:01} & \gcell{07:09} & \gcell{06:12} & \gcell{06:21} \\
& S-GRPO & 03:39 & 03:30 & 06:34 & 06:08 & 06:00 \\
& \gcell{\textbf{RED (ours)}} & \gcell{02:27} & \gcell{02:26} & \gcell{06:16} & \gcell{02:43} & \gcell{03:46} \\
\hline

\multirow{6}{*}{\textbf{DeepSeek-R1-Qwen-7B}}
& Vanilla & 02:13 & 02:12 & 04:11 & 02:26 & 04:52 \\
& \gcell{Think or Not} & \gcell{01:48} & \gcell{01:51} & \gcell{03:02} & \gcell{02:44} & \gcell{04:31} \\
& DAST & 01:42 & 01:59 & 02:01 & 01:15 & 02:15 \\
& \gcell{RL +length penalty} & \gcell{01:03} & \gcell{01:30} & \gcell{01:58} & \gcell{01:33} & \gcell{01:42} \\
& S-GRPO & 00:55 & 01:05 & 01:42 & 01:21 & 01:25 \\
& \gcell{\textbf{RED (ours)}} & \gcell{00:58} & \gcell{01:19} & \gcell{01:48} & \gcell{01:39} & \gcell{02:45} \\
\hline

\multirow{6}{*}{\textbf{DeepSeek-R1-Qwen-32B}}
& Vanilla & 04:00 & 04:10 & 07:06 & 01:45 & 07:29 \\
& \gcell{Think or Not} & \gcell{04:01} & \gcell{04:45} & \gcell{08:09} & \gcell{01:40} & \gcell{08:45} \\
& DAST & 02:28 & 03:07 & 04:18 & 00:59 & 05:27 \\
& \gcell{RL +length penalty} & \gcell{02:12} & \gcell{02:48} & \gcell{05:11} & \gcell{01:03} & \gcell{05:56} \\
& S-GRPO & 01:41 & 01:47 & 06:31 & 01:13 & 05:03 \\
& \gcell{\textbf{RED (ours)}} & \gcell{02:01} & \gcell{02:28} & \gcell{05:43} & \gcell{00:53} & \gcell{04:31} \\
\hline

\multirow{6}{*}{\textbf{DeepSeek-R1-Llama-8B}}
& Vanilla & 02:10 & 02:06 & 04:49 & 02:38 & 04:59 \\
& \gcell{Think or Not} & \gcell{02:46} & \gcell{02:39} & \gcell{04:53} & \gcell{02:07} & \gcell{04:02} \\
& DAST & 02:12 & 02:09 & 04:15 & 02:24 & 02:33 \\
& \gcell{RL +length penalty} & \gcell{01:18} & \gcell{01:23} & \gcell{04:17} & \gcell{02:06} & \gcell{01:57} \\
& S-GRPO & 00:59 & 01:08 & 03:36 & 02:05 & 02:29 \\
& \gcell{\textbf{RED (ours)}} & \gcell{01:15} & \gcell{01:03} & \gcell{02:57} & \gcell{02:05} & \gcell{02:08} \\
\hline

\multirow{6}{*}{\textbf{DeepSeek-R1-Llama-70B}}
& Vanilla & 06:41 & 06:37 & 10:47 & 03:42 & 10:25 \\
& \gcell{Think or Not} & \gcell{06:33} & \gcell{06:14} & \gcell{11:27} & \gcell{03:55} & \gcell{11:56} \\
& DAST & 05:36 & 05:04 & 08:26 & 02:37 & 08:19 \\
& \gcell{RL +length penalty} & \gcell{05:48} & \gcell{05:11} & \gcell{04:20} & \gcell{02:27} & \gcell{08:50} \\
& S-GRPO & 05:22 & 03:24 & 06:38 & 02:27 & 08:28 \\
& \gcell{\textbf{RED (ours)}} & \gcell{03:39} & \gcell{04:56} & \gcell{05:59} & \gcell{02:25} & \gcell{06:32} \\

\Xhline{1.2pt}
\end{tabular}
}
\caption{Total inference runtime (mm:ss) comparison across baselines. The experiments enable the early-exit mechanism.%
\textbf{Vanilla} represents the baseline runtime, while \textbf{RED (ours)} denotes our method.}
\label{tab:latency_full}
\end{table*}

\paragraph{Why refining the reasoning state helps.}
Finally, the role of Refine First can be modeled as improving the entire family of prompt-induced distributions by shifting probability mass toward $a^\star$ earlier.
One simple parameterization is a monotone growth model
\begin{equation}
\small
p_t^{(m)} = \sigma\!\bigl(\alpha_m (t \mminus \tau_m)\bigr),
\qquad
\sigma(x) := \frac{1}{1 + e^{\mminus x}},
\end{equation}
where $\tau_m$ is the prompt specific time when the latent answer becomes confident.
A successful refinement reduces these effective times, that is, $\tau_m \mapsto \tau_m \mminus \Delta$ with $\Delta > 0$,
which decreases the expected hitting time $T$ while leaving the concentration bounds for wrong triggers unchanged or improved.
This provides a mechanistic explanation for why \textsc{RED} simultaneously improves $\mathrm{ESC}$ and reduces $\mathrm{WESR}$ relative to \textsc{DC} in Table \ref{tab:reverse_ablation_shortnames}.

\section{Analysis of Latency Overhead}
\label{sec:stime}

To rigorously assess the latency overhead introduced by our \ourmethod (comprising both the entropy-based intervention in Section \ref{sec:method-2} and the probe-based early exit in Section \ref{sec:method-3}), we conducted comprehensive latency profiling across all evaluation benchmarks. All experiments were executed on a high-performance computing cluster equipped with NVIDIA A100-80G accelerators.

Crucially, to isolate the raw latency cost of these additional operations from the acceleration benefits of early exiting, we implemented a \textbf{``non-stopping'' stress test}.

In this configuration, the \textit{Refining} module remains fully active, performing real-time entropy variance monitoring and triggering negative sampling interventions when thresholds are breached. Simultaneously, the \textit{Discarding} module executes the parallel probing and consistency checks at every interval $K$.
However, we intentionally disable the final termination trigger even when the \textit{dual-consistency} condition is met.

This forces the model to traverse the full generation trajectory while bearing the maximum cumulative computational burden of both monitoring computations and periodic probing.
We recorded the \textit{total wall-clock time} required to complete inference over the \textbf{entire test set} (e.g., the full validation splits of GSM8K, MATH, and GPQA-Diamond) to quantify this worst-case latency overhead.

Table~\ref{tab:latency} presents the comparison between the vanilla decoding baseline and our method.
The results indicate that under the \textbf{``non-stopping'' stress test setting}---where the model is forced to execute all monitoring and probing operations without early exiting---we observe an \textbf{Average Relative Latency Overhead} of approximately \textbf{4.6\%} across all evaluated models and benchmarks.
This confirms that the intrinsic computational cost of our dual-mechanism framework is marginal, primarily attributed to the parallel design of the probe which minimizes interference with the main decoding pipeline.
Crucially, in practical deployment scenarios where the early-exit mechanism is active, this minimal overhead is easily amortized by the substantial reduction in generation steps (often saving $>$50\% of tokens as shown in Table \ref{tab:consolidated_results_wide}).
Consequently, the additional latency becomes negligible, and the system achieves a net gain in overall inference efficiency.

Table~\ref{tab:latency_full} presents the real-world inference runtime, excluding AlphaOne and GRPO due to their prohibitive computational costs. While \ourmethod introduces a minor operational overhead ($\sim$4.6\%), the results demonstrate that this cost is negligible compared to the massive acceleration achieved through adaptive termination, consistently yielding the lowest or near-lowest total wall-clock time across benchmarks.

Specifically, across all evaluated reasoning models and benchmarks, \ourmethod consistently achieves the lowest or near-lowest total wall-clock time. 
For instance, on the DeepSeek-R1-Llama-70B model, our method reduces the inference time for the AIME24 benchmark from 06:41 to 03:39---a speedup of nearly 1.8$\times$. 
This confirms that the substantial reduction in generated tokens effectively ``amortizes'' the per-step probing and intervention costs. 
Ultimately, the experimental data suggest that the minor latency introduced by intermediate probing is a highly efficient cost, yielding significant net savings in total wall-clock time.

\section{Theoretical Analysis:FoE Makes the First Solution the Best}
\label{sec:theory}
\noindent
This section provides an information-theoretic and probabilistic account of why \emph{extending} reasoning (thus producing subsequent solutions \subs) can increase the probability of error rather than decrease it.
The key mechanism is that errors in a reasoning trace are not isolated: once a wrong artifact enters the context, it can be repeatedly reused and amplified, forming a \emph{forest} of causally linked error nodes whose growth budget increases with test-time.

\subsection{FoE mechanism (informal, non-asymptotic).}
We represent the internal error structure of a solution trace as a \emph{directed forest} of wrong artifacts.
\textbf{FoE is a causal forest of errors:} each error node has at most one parent, defined as the nearest earlier error that \emph{directly induces} it via explicit artifact reuse (wrong values, wrong assumptions, wrong bindings, wrong rules). Because new root errors can arise over time, the resulting structure is a forest rather than a single chain.
\textbf{Root dominance and strong parent--child dependence:} fixing a descendant without fixing its ancestor typically fails to stop the creation of new errors because the uncorrected ancestor remains in-context as a persistent ``error factory''; in contrast, correcting the root of a formed tree strongly suppresses subsequent growth in both depth and breadth.
Empirically, \textbf{\first has a smaller and slower-growing FoE than \subs} under both static forest metrics (number of trees, nodes per tree, depth) and a dynamic reproduction-rate metric.
Moreover, \textbf{root-error genesis correlates with the \emph{joint} elevation of uncertainty and volatility:} neither a single uncertainty statistic (entropy) nor a single volatility statistic (entropy variance) suffices alone, while their joint elevation is most predictive for new root errors.
Finally, \textbf{self-reflection does not reliably prune FoE (and degrades in \subs):} later ``corrections'' are often refusals or ``fake corrections'' that edit surface text but preserve the underlying wrong artifact, hence additional exploration tends to preserve and amplify wrong artifacts rather than remove them.
Consistently, \textbf{latent instability is higher in \subs even when the final answer looks correct:} perturbing the generation by resampling from intermediate states causes \subs to drift to incorrect answers more easily, indicating weaker ``lock-in'' toward the correct solution trajectory.

\subsection{A probabilistic FoE model.}
Let $Y$ be the ground-truth answer for question $Q$. A reasoning model generates a sequence of tokens (or steps) forming a solution trace $S$, and outputs an extracted final answer $\hat{Y}(S)$.
We represent the error structure inside a trace $S$ by a directed forest
\begin{equation}
    \mathcal{F}(S) = (\mathcal{V}(S), \mathcal{E}(S)),
\end{equation}
where $\mathcal{V}(S)$ is the set of identified error nodes (artifacts) and $\mathcal{E}(S)$ links each error to its nearest inducing predecessor (or marks it as a root).

\subsubsection{Root-error triggering is driven by entropy and entropy-variance.}
Let $p_t(\cdot)$ denote the token distribution at generation step $t$, and define step entropy
\begin{equation}
H_t \;=\; -\sum_{x} p_t(x)\log p_t(x).
\label{eq:token-entropy}
\end{equation}
Over a local window $W_t$, define the mean entropy and entropy variance
\begin{equation}
 \begin{aligned}
h_t \;&=\; \frac{1}{|W_t|}\sum_{i\in W_t} H_i,
\\
v_t \;&=\; \frac{1}{|W_t|-1}\sum_{i\in W_t} (H_i-h_t)^2 .
\end{aligned}   
\end{equation}
Let $R_t$ be the indicator that a \emph{new root error} is created at step $t$.
The FoE findings imply that root triggering is \emph{supermodular} in $(h_t,v_t)$: joint elevation is more dangerous than either alone. A convenient statistical abstraction is a monotone interaction model
The probability is modeled as:
\begin{equation}
\Pr(R_t=1 \mid Q, S_t) \le \sigma\big( \eta_t \big),
\label{eq:root-trigger-model}
\end{equation}
where $S_t$ denotes the state at time $t$, and $\eta_t$ represents the logit term defined as:
\begin{equation}
\eta_t = \beta_0 + \beta_h h_t + \beta_v v_t + \beta_{hv} h_t v_t.
\end{equation}
where $\sigma(\cdot)$ is a sigmoid link and $\beta_{hv}>0$ captures the empirically observed synergy between entropy and entropy-variance in producing \emph{root} errors.

\subsubsection{FoE growth as a branching process (root dominance formalized).}
Once a root error is created, it can induce descendant errors through artifact reuse. We approximate each root-initiated tree as a truncated Galton--Watson process with solution-dependent mean reproduction $\rho$ (capturing the FoE reproduction metric).
Let $G(\ell;\rho)$ be the expected total number of nodes produced by a single root when $\ell$ steps remain:
\begin{equation}
G(\ell;\rho)
\;=\;
\sum_{d=0}^{\ell} \rho^d.
\label{eq:gw-size}
\end{equation}
Let $T$ be the length (in steps) of the trace, and define the expected total number of error nodes
\begin{equation}
\mathbb{E}\big[|\mathcal{V}(S)|\big]
\;\le\;
\sum_{t=1}^{T}
\Pr(R_t=1)\, G(T-t;\rho).
\label{eq:expected-errors}
\end{equation}
This expression isolates the \textbf{root dominance} mechanism: decreasing $\Pr(R_t=1)$ (fewer roots) or decreasing $\rho$ (slower reproduction) suppresses \emph{the entire} downstream error forest.

\subsubsection{From FoE size to answer error: a general bound.}
The extracted answer $\hat{Y}(S)$ is wrong if at least one \emph{decision-critical} error influences the final decision.
Associate each error node $v\in\mathcal{V}(S)$ with an event $C_v$ indicating that $v$ becomes decision-critical (directly or through descendants) for the final answer.
Assume a bounded per-node criticality probability:
\begin{equation}
\Pr(C_v=1 \mid v \in \mathcal{V}(S)) \;\le\; \kappa,
\label{eq:criticality}
\end{equation}
for some task/model-dependent constant $\kappa$.

By Boole's inequality (union bound), the error probability is bounded by the total critical risk across all nodes in $\mathcal{V}(S)$. Conditioning on the random error set and taking the expectation, we have:
\begin{equation}
\small
\begin{aligned}
\Pr\!\big(\hat{Y}(S)\neq Y\big) 
&= \Pr\!\Big(\bigcup_{v\in\mathcal{V}(S)} C_v\Big) \\
&\le \mathbb{E}\bigg[\sum_{v\in\mathcal{V}(S)} \Pr(C_v=1 \mid v)\bigg].
\end{aligned}
\label{eq:error-bound-step1}
\end{equation}
Recalling the bounded per-node criticality $\Pr(C_v=1) \le \kappa$ from \eqref{eq:criticality}, this simplifies to:
\begin{equation}
\Pr\!\big(\hat{Y}(S)\neq Y\big) \le \kappa \,\mathbb{E}\big[|\mathcal{V}(S)|\big].
\label{eq:error-bound-by-foe}
\end{equation}

Let $\mathcal{M}(S) := \mathbb{E}[|\mathcal{V}(S)|]$ denote the expected size of the error forest. By substituting the branching process result \eqref{eq:expected-errors} into \eqref{eq:error-bound-by-foe}, we obtain the explicit FoE-to-error bound:
\begin{equation}
\small
\Pr(\hat{Y}(S) \neq Y) \le \Phi(T, \mathbf{P}_R, \rho) := \kappa \cdot \mathcal{M}(S),
\label{eq:phi-def-simple}
\end{equation}
where the expected forest size $\mathcal{M}(S)$ is expanded as:
\begin{equation}
\mathcal{M}(S) = \sum_{t=1}^{T} \Pr(R_t=1) G(T-t; \rho).
\label{eq:m-expand}
\end{equation}
Here, $\mathbf{P}_R := \{\Pr(R_t=1)\}_{t=1}^T$ denotes the root-triggering profile.

\subsubsection{Bounds for \first\ and \subs\ (why \pheno).}
Let $S_{\first}$ denote the \first solution trace, and $S_{\subs}$ denote a representative subsequent solution trace.
Define the FoE-based upper bounds for the first and subsequent traces as:
\begin{equation}
\begin{aligned}
\Phi_{\first} &:= \Phi(T_{\first},\mathbf{P}_R^{\first},\rho_{\first}), \\
\Phi_{\subs} &:= \Phi(T_{\subs},\mathbf{P}_R^{\subs},\rho_{\subs}),
\end{aligned}
\label{eq:phi-first-subs}
\end{equation}
where $\mathbf{P}_R^{\first}$ and $\mathbf{P}_R^{\subs}$ denote the respective root-triggering profiles $\{\Pr(R_t^{( \cdot )}=1)\}_{t=1}^{T_{( \cdot )}}$.

The FoE findings imply the following structural dominance relations (holding in expectation, and often pointwise along the trace):
\begin{equation}
\begin{aligned}
\Pr(R^{\first}_t=1) \;&\le\; \Pr(R^{\subs}_t=1),\\
\rho_{\first} \;&\le\; \rho_{\subs},\\
T_{\first} \;&\le\; T_{\subs}.
\end{aligned}
\label{eq:domination-first}
\end{equation}

Since $G(\ell;\rho)=\sum_{d=0}^{\ell}\rho^d$ is nondecreasing in both $\ell$ and $\rho$, the bound function $\Phi(T,\mathbf{P}_R,\rho)$ is monotone in (i) the horizon $T$, (ii) the componentwise root-trigger profile $\mathbf{P}_R$, and (iii) the reproduction rate $\rho$.
Therefore, \eqref{eq:domination-first} implies
\begin{equation}
\Phi_{\first} \;\le\; \Phi_{\subs}.
\label{eq:phi-order}
\end{equation}

Consequently, the error probabilities satisfy the following \emph{non-asymptotic} upper bounds:
\begin{equation}
\begin{aligned}
\Pr\!\big(\hat{Y}(S_{\first})\neq Y\big) &\le \Phi_{\first} \le \Phi_{\subs}, \\
\Pr\!\big(\hat{Y}(S_{\subs})\neq Y\big) &\le \Phi_{\subs}.
\end{aligned}
\label{eq:bounds-chain}
\end{equation}

These relations indicate that \subs\ operates in a regime of higher \emph{informational stress} (higher $\mathbf{P}_R$) with a weaker \emph{error-pruning} capability (higher $\rho$), compounded by a longer accumulation horizon ($T$), which yields a strictly looser FoE-based risk upper bound.

\subsubsection{A bound for ``\subs biases \first''.}
We formalize the event that subsequent reasoning segments override a correct first solution (“\subs biases \first”) as the event that the first solution is correct, but the final decision after continued exploration is incorrect.
Let $\hat{Y}_{\text{final}}$ denote the extracted answer after the model continues reasoning beyond the first solution.
Define the misguidance event
\begin{equation}
\mathcal{B}
\;:=\;
\big\{\hat{Y}(S_{\first})=Y \ \wedge\  \hat{Y}_{\text{final}}\neq Y\big\}.
\label{eq:bias-event}
\end{equation}

\paragraph{Information contamination (artifact reuse across segments).}
The continuation is generated under the \emph{same} context as $S_{\first}$, so its error dynamics are conditioned on the full history $\mathcal{H}_{\first}$ of the first trace.
In particular, attention-based artifact reuse can ``contaminate'' the continuation: previously introduced (possibly non-decision-critical) artifacts can be revisited and amplified, effectively increasing root-triggering propensity and/or reproduction.

Let $\mathcal{T}_{\text{ext}}:=\{T_{\first}+1,\dots,T_{\text{final}}\}$ denote the time indices of the continuation segment.
Conditioning on $\mathcal{H}_{\first}$, we can upper-bound the final-error probability by reusing the FoE bound on the continuation:

To simplify, let $\mathcal{M}_{\text{ext}}(\mathcal{H}_{\first})$ denote the expected size of the error forest generated during the continuation, conditioned on the first trace's history $\mathcal{H}_{\first}$:
\begin{equation}
\begin{aligned}
\mathcal{M}_{\text{ext}}
&= \sum_{t\in \mathcal{T}_{\text{ext}}}
\Pr(R^{\text{ext}}_t=1 \mid \mathcal{H}_{\first}) \\
&\quad \cdot G(T_{\text{final}}-t;\rho_{\text{ext}}).
\end{aligned}
\label{eq:m-ext-def}
\end{equation}
Conditioning on $\mathcal{H}_{\first}$, the error probability of the final decision can be upper-bounded by:
\begin{equation}
\small
\Pr(\hat{Y}_{\text{final}}\neq Y \mid \mathcal{H}_{\first}) \;\le\; \kappa \cdot \mathcal{M}_{\text{ext}}(\mathcal{H}_{\first}).
\label{eq:conditional-final-error-simple}
\end{equation}

Using a pessimistic domination that captures degraded reflection in later segments, we bound the conditional error probability by its \subs-regime parameters. Let $\Phi_{\text{ext}} := \kappa \cdot \mathcal{M}_{\text{ext}}^{\subs}$ denote the error bound for the continuation segment, where:
\begin{equation}
\small
\Phi_{\text{ext}} = \kappa \sum_{t\in \mathcal{T}_{\text{ext}}}\Pr(R^{\subs}_t=1)\,G(T_{\text{final}}-t;\rho_{\subs}).
\label{eq:phi-ext-expand}
\end{equation}
Then, the probability that the final decision is incorrect given a correct first solution is bounded by:
\begin{equation}
\Pr(\hat{Y}_{\text{final}}\neq Y \mid \hat{Y}(S_{\first})=Y) \;\le\; \Phi_{\text{ext}}.
\label{eq:conditional-bias-bound-simple}
\end{equation}

Therefore,
The probability of the misguidance event $\mathcal{B}$ then simplifies to the product of the first solution's success probability and the continuation's error bound:
\begin{equation}
\Pr(\mathcal{B}) \;\le\; \Pr(\hat{Y}(S_{\first})=Y) \cdot \Phi_{\text{ext}}.
\label{eq:bias-bound-final}
\end{equation}

\subsection{Takeaway.}
Equations \eqref{eq:phi-order} and \eqref{eq:bounds-chain} formalize a FoE-based risk characterization: when root errors are more likely under elevated entropy--variance and amplified via higher reproduction over a longer horizon, the resulting error upper bound is tighter for \first\ than for \subs. 
In other words, \subs\ operates under higher \emph{informational stress} and weaker \emph{error-pruning}, yielding a looser non-asymptotic risk bound.

Equation \eqref{eq:bias-bound-final} further shows that generating \subs\ is not merely ``extra compute'': it opens an additional failure channel whereby continued exploration overrides a correct \first\ solution, primarily through artifact reuse that introduces new root errors. 

Together, these results support the theoretical statement that extending reasoning beyond the first solution is not guaranteed to improve reliability and can strictly worsen it---hence, \emph{FoE makes the first solution the best}.

\section{Few-Shot Prompt of Parent-Children Score(PCS) Judging}
\label{sec:few-shot}
Prompts are as shown as below.
\begin{PromptFrame}{\textsc{Few-Shot Prompt of PCS Judging}}
\begin{lstlisting}[style=PromptStyle]
# PCS (Parent-Children Score) - Few-shot Prompt for FoE Causal Scoring

## Role
You are a **forensic causal annotator** for a *Forest of Errors (FoE)* reasoning trace.

## Objective
Given an earlier error node **e_i** (candidate parent) and a later error node **e_j** (child), plus the full reasoning prefix up to and including **e_j**, output a **PCS score** measuring whether **e_i directly induces e_j**.

You are scoring **direct causal induction** (error propagation), not topical similarity.

---

## Inputs
You will receive:

### CONTEXT_UP_TO_CHILD
The full reasoning prefix up to and including the child node **e_j**.

### CANDIDATE_PARENT_NODE (earlier)
The text of error node **e_i** (chronologically earlier than e_j).

### CHILD_NODE (later)
The text of error node **e_j**.

---

## Output (STRICT)
Output **exactly one line** containing **one number**:
- range: **1.0 to 5.0** (inclusive)
- format: **exactly one decimal place**
- regex: `^(?:[1-4]\.[0-9]|5\.0)$`

**Do not output anything else**: no explanation, no labels, no JSON, no punctuation, no extra whitespace.

---

## Core definitions (do not reinterpret)

### Wrong artifact
A "wrong artifact" introduced by **e_i** can be any incorrect item that later gets reused, including:
- wrong numeric value / denominator / count / derived quantity,
- wrong formula choice (e.g., using circumference for area),
- wrong theorem or rule application (e.g., inclusion-exclusion used with a wrong intersection term),
- wrong constraint interpretation or assumption ("disjoint", "independent", "without replacement", etc.),
- wrong definition / variable binding ("Let T=...", wrong meaning of a symbol),
- wrong intermediate statement (equation, inequality, recurrence, invariant, case split rule, etc.),
- wrong algorithmic step / update rule.

### Direct parent (direct inducer)
**e_i is a direct parent of e_j** iff BOTH hold:
1) **Dependency**: e_j's wrongness depends on at least one wrong artifact introduced by e_i; AND  
2) **Directness**: within the provided prefix, there is **no closer, more-specific error step** that better explains the particular artifact reuse in e_j.

---

## High-score propagation patterns 
High PCS scores are appropriate for **any strong artifact flow**, including:
- **numeric propagation** (wrong value reused downstream),
- **formula propagation** (wrong formula chosen, then used downstream),
- **theorem/rule propagation** (wrong theorem usage yields a derived equation/constraint, then reused),
- **assumption propagation** ("independent", "disjoint", "monotone", etc.),
- **definition/binding propagation** (a symbol is bound wrong and reused).

---

## Threshold safety rule (keep it conservative)
Because downstream will treat **PCS >= 4.0** as "connect an edge", be conservative.

You may output **PCS >= 4.0** only when the context provides **concrete evidence** that:
- the child reuses the parent's wrong artifact (same value/symbol/assumption/derived equation), AND
- there is no nearer step in the prefix that more directly explains the reused artifact.

If these are not satisfied, output **PCS <= 3.9**.

Tie-break (critical):
- If you are torn between **3.9 and 4.0**, output **3.9**.

---

# Anchor meanings (1.0 / 2.0 / 3.0 / 4.0 / 5.0) - detailed

## 5.0 - Certain direct parent (near-certain direct induction)
Use 5.0 only if:
- the artifact flow is **explicit** and **central** (the child's error is overwhelmingly a consequence of reusing the parent's artifact), AND
- fixing e_i would **almost certainly** remove or materially change e_j, AND
- there is no plausible competing parent.

Typical 5.0 cases:
- child directly computes from a wrong number produced by parent,
- child directly computes from a wrong formula/theorem result produced by parent,
- child is essentially the "application step" of the parent's wrong conclusion.

Disqualifier:
- if the child has a substantial independent error that would remain even if e_i were corrected, prefer 4.0+.

## 4.0 - Likely direct parent (threshold)
Use 4.0 when:
- there is concrete artifact reuse evidence, AND
- e_i is the most plausible direct inducer, BUT
- there is non-trivial uncertainty (e.g., child also contains an additional independent mistake, or the dependency is not maximally explicit).

Typical 4.0 cases:
- child reuses the artifact, but also introduces another separate error,
- dependency is clear but not "maximally explicit",
- mild competing-cause ambiguity exists, but e_i is still the best direct cause.

## 3.0 - Meaningfully related, but probably not direct
Use 3.0 when there is **meaningful relatedness**, yet direct parenthood is not established, such as:
- e_i is a plausible **ancestor** but a closer mediator exists,
- the dependency seems plausible but lacks concrete signals in the prefix,
- the child's wrongness would likely persist even if e_i were corrected.

3.0 means: "There is a real relationship, but not a direct parent edge."

## 2.0 - Almost unrelated; only weak, surface relatedness
Use 2.0 when:
- the nodes are in the same general topic/segment or share variable names, BUT
- there is **no credible artifact flow**, and any causal story would be speculative.

2.0 is **near-unrelated** but not fully: there is slight overlap (same symbols/topic/proximity), yet no dependency.

## 1.0 - Unrelated
Use 1.0 when:
- there is no meaningful link at all,
- different subgoals/branches/domains,
- no shared entities or any shared tokens are clearly from the original problem statement rather than reuse.

---

# One-decimal scoring rule
Output **exactly one number** in **[1.0, 5.0]** with **exactly one decimal place**.  
Choose the score based on the anchor meanings and the strength of direct artifact flow.  
No other text.

---

# Few-shot examples (anchor points)
IMPORTANT: Each example's output is **only the score**, one line.

## Anchor = 5.0 (3 examples)

### Example 5.0-A (simple numeric propagation)
**CONTEXT_UP_TO_CHILD**
[01] There are 6 chapters with 18 pages each.  
[02] Total pages = 6 * 18 = 96.  
[03] Printing costs $0.10 per page, so total cost = 96 * 0.10 = $9.60.

**CANDIDATE_PARENT_NODE**
[02] Total pages = 6 * 18 = 96.

**CHILD_NODE**
[03] Printing costs $0.10 per page, so total cost = 96 * 0.10 = $9.60.

**OUTPUT**
5.0

### Example 5.0-B (simple formula propagation)
**CONTEXT_UP_TO_CHILD**
[01] Radius r = 4. We need the area of the circle.  
[02] Area = 2 * pi * r = 8 * pi.  
[03] Therefore area ~= 8 * pi ~= 25.13.

**CANDIDATE_PARENT_NODE**
[02] Area = 2 * pi * r = 8 * pi.

**CHILD_NODE**
[03] Therefore area ~= 8 * pi ~= 25.13.

**OUTPUT**
5.0

### Example 5.0-C (complex propagation: misjudgment -> wrong theorem use -> derived conclusion)
**CONTEXT_UP_TO_CHILD**
[01] Let A be numbers in {1..100} divisible by 2, and B be numbers divisible by 5.  
[02] |A| = 50 and |B| = 20.  
[03] Since 2 and 5 are coprime, assume |A \cap B| = 0.  
[04] By inclusion-exclusion, |A \cup B| = |A| + |B| - |A \cap B| = 50 + 20 - 0 = 70.

**CANDIDATE_PARENT_NODE**
[03] Since 2 and 5 are coprime, assume |A \cap B| = 0.

**CHILD_NODE**
[04] By inclusion-exclusion, |A \cup B| = |A| + |B| - |A \cap B| = 50 + 20 - 0 = 70.

**OUTPUT**
5.0

---

## Anchor = 4.0 (3 examples)

### Example 4.0-A (direct dependency, but child also has an extra mistake)
**CONTEXT_UP_TO_CHILD**
[01] A box has 9 blue marbles and 5 green marbles.  
[02] Total marbles = 9 + 5 = 12.  
[03] Probability(green) = 5/12 ~= 0.45.

**CANDIDATE_PARENT_NODE**
[02] Total marbles = 9 + 5 = 12.

**CHILD_NODE**
[03] Probability(green) = 5/12 ~= 0.45.

**OUTPUT**
4.0

### Example 4.0-B (complex propagation: wrong theorem form reused; child has extra arithmetic error)
**CONTEXT_UP_TO_CHILD**
[01] We want |A \cup B \cup C|. Given |A|=30, |B|=25, |C|=20, |A \cap B|=10, |A \cap C|=5, |B \cap C|=4, and |A \cap B \cap C|=3.  
[02] Use inclusion-exclusion: |A \cup B \cup C| = |A|+|B|+|C| - |A \cap B| - |A \cap C| - |B \cap C|.  
[03] Plug in: 30+25+20 = 70, so |A \cup B \cup C| = 70 - 10 - 5 - 4 = 51.

**CANDIDATE_PARENT_NODE**
[02] Use inclusion-exclusion: |A \cup B \cup C| = |A|+|B|+|C| - |A \cap B| - |A \cap C| - |B \cap C|.

**CHILD_NODE**
[03] Plug in: 30+25+20 = 70, so |A \cup B \cup C| = 70 - 10 - 5 - 4 = 51.

**OUTPUT**
4.0

### Example 4.0-C (simple formula propagation, plus child introduces an extra arithmetic mistake)
**CONTEXT_UP_TO_CHILD**
[01] Radius r = 4. We need the area of the circle.  
[02] Area = 2 * pi * r = 8 * pi.  
[03] Therefore area ~= 8 * 3.14 = 23.12.

**CANDIDATE_PARENT_NODE**
[02] Area = 2 * pi * r = 8 * pi.

**CHILD_NODE**
[03] Therefore area ~= 8 * 3.14 = 23.12.

**OUTPUT**
4.0

---

## Anchor = 3.0 (3 examples)

### Example 3.0-A (ancestor: a nearer mediator is more direct)
**CONTEXT_UP_TO_CHILD**
[01] A box has 9 blue marbles and 5 green marbles.  
[02] Total marbles = 9 + 5 = 12.  
[03] Let T = 12 be the total number of marbles.  
[04] Probability(green) = 5/T = 5/12.

**CANDIDATE_PARENT_NODE**
[02] Total marbles = 9 + 5 = 12.

**CHILD_NODE**
[04] Probability(green) = 5/T = 5/12.

**OUTPUT**
3.0

### Example 3.0-B (related, but a closer step is the direct inducer)
**CONTEXT_UP_TO_CHILD**
[01] Assume the two draws are independent.  
[02] Therefore P(two reds) = P(first red) * P(second red).  
[03] Take P(first red)=3/5 and P(second red)=3/5, so P(two reds)=(3/5)*(3/5).

**CANDIDATE_PARENT_NODE**
[01] Assume the two draws are independent.

**CHILD_NODE**
[03] Take P(first red)=3/5 and P(second red)=3/5, so P(two reds)=(3/5)*(3/5).

**OUTPUT**
3.0

### Example 3.0-C (enabling/ancestor; child mostly driven by another nearer wrong artifact)
**CONTEXT_UP_TO_CHILD**
[01] The function passes through points (0,1) and (1,3).  
[02] So it must be linear.  
[03] Slope m = (3-1)/(1-0) = 1.  
[04] Therefore f(x) = 1 + 1*x = x + 1.

**CANDIDATE_PARENT_NODE**
[02] So it must be linear.

**CHILD_NODE**
[04] Therefore f(x) = 1 + 1*x = x + 1.

**OUTPUT**
3.0

---

## Anchor = 2.0 (3 examples)

### Example 2.0-A (same symbol/topic, but child does not use the parent's artifact)
**CONTEXT_UP_TO_CHILD**
[01] The problem states n = 8.  
[02] Assume n = 10 for convenience.  
[03] Using n = 8, compute 8! = 30240.

**CANDIDATE_PARENT_NODE**
[02] Assume n = 10 for convenience.

**CHILD_NODE**
[03] Using n = 8, compute 8! = 30240.

**OUTPUT**
2.0

### Example 2.0-B (same general domain; errors are independent)
**CONTEXT_UP_TO_CHILD**
[01] For a fair die, P(roll <= 2) = 2/6 = 1/2.  
[02] The expected value of a fair die is 4.

**CANDIDATE_PARENT_NODE**
[01] For a fair die, P(roll <= 2) = 2/6 = 1/2.

**CHILD_NODE**
[02] The expected value of a fair die is 4.

**OUTPUT**
2.0

### Example 2.0-C (same variable name, but child overwrites/ignores parent's value)
**CONTEXT_UP_TO_CHILD**
[01] Solve x + 1 = 3 => x = 1.  
[02] In part (b), set x = 5.  
[03] Using x = 5, compute y = 2x = 12.

**CANDIDATE_PARENT_NODE**
[01] Solve x + 1 = 3 => x = 1.

**CHILD_NODE**
[03] Using x = 5, compute y = 2x = 12.

**OUTPUT**
2.0

---

## Anchor = 1.0 (2 examples)

### Example 1.0-A (unrelated subgoals)
**CONTEXT_UP_TO_CHILD**
[01] Simplify 18/24 by dividing by 6 to get 3/5.  
[02] Different step: derivative of x^2 is 2.

**CANDIDATE_PARENT_NODE**
[01] Simplify 18/24 by dividing by 6 to get 3/5.

**CHILD_NODE**
[02] Different step: derivative of x^2 is 2.

**OUTPUT**
1.0

### Example 1.0-B (different domains, no shared artifacts)
**CONTEXT_UP_TO_CHILD**
[01] Triangle area = (1/2)bh = (1/2)*10*3 = 60.  
[02] Probability of heads in a fair coin is 1/3.

**CANDIDATE_PARENT_NODE**
[01] Triangle area = (1/2)bh = (1/2)*10*3 = 60.

**CHILD_NODE**
[02] Probability of heads in a fair coin is 1/3.

**OUTPUT**
1.0

---

# Now score the real case

## CONTEXT_UP_TO_CHILD
{{CONTEXT_UP_TO_CHILD}}

## CANDIDATE_PARENT_NODE (earlier)
{{CANDIDATE_PARENT_NODE}}

## CHILD_NODE (later)
{{CHILD_NODE}}

# Output: one line, one number, exactly one decimal (1.0-5.0). No other text.
\end{lstlisting}
\end{PromptFrame}
\twocolumn

\end{document}